\pdfoutput=1
\documentclass[11pt,a4paper]{article}
\usepackage{arabtex}
\usepackage{utf8}
\usepackage[]{emnlp2022}

\usepackage{times}
\usepackage{latexsym}

\usepackage[utf8]{inputenc}

\usepackage{microtype}

\usepackage{cmap}
\usepackage[T5]{fontenc}
\usepackage{subcaption}
\usepackage{graphicx}
\usepackage{rotating}
\usepackage{booktabs}
\usepackage[normalem]{ulem}
\usepackage{multirow}
\usepackage{enumitem}
\usepackage{amsmath}
\usepackage{xspace}
\usepackage{footnote}
\makesavenoteenv{tabular}
\makesavenoteenv{table}
\usepackage{makecell}
\usepackage{colortbl}
\usepackage{placeins}

\usepackage{arydshln}
\usepackage{CJKutf8}

\newcommand{\mD}{\mathcal{D}}
\newcommand{\mE}{\mathcal{E}}

\newcommand{\mL}{\mathcal{L}}
\newcommand{\mM}{\mathcal{M}}

\newcommand{\mP}{\mathcal{P}}

\newcommand{\mT}{\mathcal{T}}

\newcommand{\mV}{\mathcal{V}}
\newcommand{\mY}{\mathcal{Y}}

\newcommand{\rom}[1]{%
	\textup{\uppercase\expandafter{\romannumeral#1}}%
}

\newcommand{\highest}[1]{\textbf{#1}}% == highest score

\newcommand{\eat}[1]{\ignorespaces}

\def\|#1|{\mathid{#1}}
\newcommand{\mathid}[1]{\ensuremath{\mathit{#1}}}
\def\<#1>{\codeid{#1}}
\protected\def\codeid#1{\ifmmode{\mbox{\smaller\ttfamily{#1}}}\else{\smaller\ttfamily
		#1}\fi}

\newcommand{\lang}[1]{\textit{#1}}

\definecolor{light-gray}{rgb}{.902, .902, .902}

\newcommand{\mask}{\texttt{[Mask]}}
\newcommand{\sep}{{\small\texttt{[Sep]}}}
\newcommand{\xtest}{\Tilde{x}}

\newcommand{\model}[2]{#1$_\text{#2}$}
\newcommand{\Ours}{XGLM\xspace}
\newcommand{\Oursns}{XGLM}
\newcommand{\xlmgen}{GPT-3$_\text{6.7B}$ {\it repl.}\xspace}

\newcommand{\cchxl}{CC100-XL\xspace}

\newcommand{\hide}[1]{}

\newcommand{\entext}[1]{\textcolor{blue}{#1}}
\newcommand{\vitext}[1]{\textcolor{brown}{#1}}
\newcommand{\zhtext}[1]{\begin{CJK*}{UTF8}{gbsn}#1\end{CJK*}}

\DeclareMathOperator*{\argmax}{arg\,max}

\newcommand{\comment}[1]{{\bf \textcolor{red}{ {\sf #1}}}}

\definecolor{maroon}{rgb}{0.5, 0.0, 0.0}
\newcommand{\vic}[1]{\textcolor{maroon}{\small{\bf [Victoria: #1]}}}

\newcommand{\xian}[1]{\textcolor{olive}{\small{\bf [Xian: #1]}}}

\title{Few-shot Learning with Multilingual Generative Language Models}
\hide{ 
\author{First Author \\
  Affiliation / Address line 1 \\
  Affiliation / Address line 2 \\
  Affiliation / Address line 3 \\
  \texttt{email@domain} \\\And
  Second Author \\
  Affiliation / Address line 1 \\
  Affiliation / Address line 2 \\
  Affiliation / Address line 3 \\
  \texttt{email@domain} \\}
 }
 
\author{
{\bf Xi Victoria Lin}\thanks{\ \ Equal contribution. Correspondence to: $\langle$victorialin@meta.com, xianl@meta.com$\rangle$.}\footnotemark[1]{\normalfont ,}
{\bf Todor Mihaylov}{\normalfont ,}
{\bf Mikel Artetxe}{\normalfont ,}
{\bf Tianlu Wang}{\normalfont ,}
{\bf Shuohui Chen},
{\bf Daniel Simig}{\normalfont ,} \\
{\bf Myle Ott}{\normalfont ,}
{\bf Naman Goyal},
{\bf Shruti Bhosale}{\normalfont ,}
{\bf Jingfei Du}, 
{\bf Ramakanth Pasunuru}, 
{\bf Sam Shleifer}, \\
{\bf Punit Singh Koura}, 
{\bf Vishrav Chaudhary},
{\bf Brian O'Horo}, 
{\bf Jeff Wang},\\ 
{\bf Luke Zettlemoyer}, 
{\bf Zornitsa Kozareva}, 
{\bf Mona Diab}, 
{\bf Veselin Stoyanov},
{\bf Xian Li}\footnotemark[1]{\normalfont } \\
Meta AI \\
}
 
\begin{document}
\maketitle

\begin{abstract}
% \vic{shorten abstract based on shortened content}
Large-scale generative language models such as GPT-3 are competitive few-shot learners. % few-shot learners that can perform a wide range of language tasks without fine-tuning. 
While these models are known to be able to jointly represent multiple languages, their training data is dominated by English, potentially limiting their cross-lingual generalization. 
%We train the first multilingual task-agnostic language model and 
In this work, we train multilingual generative language models on a % balanced 
corpus covering a diverse set of languages, and study their few- and zero-shot learning capabilities in a wide range of tasks. Our largest model with 7.5 billion parameters sets new state of the art in few-shot learning in % 25
more than 20 representative languages, outperforming GPT-3 of comparable size in multilingual commonsense reasoning (with +7.4\% absolute accuracy improvement in 0-shot settings and +9.4\% in 4-shot settings) and natural language inference (+5.4\% in each of 0-shot and 4-shot settings). %, as well as medium- and low-resource machine translation.
On the FLORES-101 machine translation benchmark, our model outperforms GPT-3 counterparts on 171 out of 182 % translation 
directions with 32 training examples, while surpassing the official supervised baseline in 45 directions.
% , outperforming GPT-3 by 6.7 accuracy points in 0-shot, and 8 points in 4-shot on average. % and finetuning with existing multilingual models. 
% We analyze cross-lingual transfer of the model in in-context learning setting, as well as its behavior in safety-critical tasks such as hate speech detection and gender bias
% It also establishes a new state of the art on few-shot machine translation across a large number of language pairs on the FLORES-101 benchmark.
We conduct an in-depth analysis of different multilingual prompting approaches, showing in particular that strong in-context few-shot learning performance across languages can be achieved via cross-lingual transfer through both templates and demonstration examples.% where the model succeeds and fails, showing in particular that it enables cross-lingual in-context learning on some tasks, while there is still room for improvement on language interference mitigation and adaptation to tasks that do not have a natural cloze format.
\footnote{Our checkpoints, code and new dataset (XStoryCloze): \url{https://github.com/facebookresearch/fairseq/tree/main/examples/xglm}.}% cannot be naturally formulated as language completion.  
%and demonstrate its few-shot generalization on a diverse set of languages beyond English. We observe strong crosslingual capabilities of these models and competitive performance across tasks such as commonsense reasoning, natural language inference and paraphrasing. %We show that multilingual language models' few-shot performance improves with scale. Our largest model with 7.5 billion parameters yields new state-of-the-art performance in few-shot learning settings, outperforming fine-tuning with existing multilingual models. Finally, we analyze the impact on responsible AI such as hate speech detection and gender bias, and discuss its broader impact to the society. We hope this work inspire more future work to build truly inclusive general-purpose language models.    
% Finally, we evaluate our models in social value tasks such as hate speech detection in 5 languages and find it has limitations similar to comparably sized GPT-3 models.
\end{abstract}

\section{Introduction}
% The paradigm in natural language processing (NLP) is shifting from fine-tuning (e.g., BERT, T5)~\cite{devlin2018bert,raffel2020exploring} to pretraining a general-purpose model to perform various tasks with a few demonstration examples (e.g. GPT-3)~\cite{brown2020gpt3}.\vic{The previous sentence is too strong. The prompting paradigm is in an exploratory stage, and the main paradigm is still fine-tuning.}\xian{That's why we used ``shifting" instead of ``has shifted" to imply there's a transition.} 
%The few-shot learning behavior emerged from large-scale generative language models such as GPT-3~\cite{brown2020gpt3} is triggering a potential paradigm shift in natural language processing, from fine-tuning on different tasks (e.g., BERT, T5)~\cite{devlin2018bert,raffel2020exploring} to letting a pre-trained language model perform various tasks by conditioning on task instructions or few demonstration examples, known as  ``prompts" \cite{liu2021pre}. 
% General-purpose models % are more scalable and applicable 
Large autoregressive language models such as GPT-3 can be adapted, via few- and zero-shot learning, to a  wide range of tasks with % competitive performance
significantly less cost than full fine-tuning~\cite{brown2020gpt3,bommasani2021opportunities}. % , making them foundation models\cite{bommasani2021opportunities}. 
 % such foundation 
These models % along with the corresponding key enablers such as model and data % size 
% scaling 
have been % only 
primarily developed for English. Although the training data of GPT-3  contains a small percentage of non-English text (7\%) allowing it to achieve some promising cross-lingual generalization, the model is almost exclusively deployed for use cases in English. Multilingual masked and sequence-to-sequence % PLMs 
language models have been studied, including mBERT, XLM-R, mT5, and mBART~\cite{devlin2018bert,conneau-etal-2020-unsupervised,Xue2020MT5:Transformer,fedus2021switch,DBLP:journals/corr/abs-2105-00572,liu2020multilingual}. These models are typically fine-tuned on large amount of labeled data in downstream tasks. Despite notable recent work at smaller scales~\cite{DBLP:journals/corr/abs-2109-03630} and for domain-specific tasks~\cite{DBLP:journals/corr/abs-2109-07684}, the multilingual few-shot learning capabilities of language models are less well understood.

In this paper, we % study large-scale zero- and few-shot learning with multilingual generative LMs. We 
train four multilingual generative language models (up to 7.5 billion parameters), \Ours's, and present % the first 
a comprehensive study of multilingual zero- and in-context few-shot learning. We train the models using a large-scale corpus of 500B tokens that comprises 30 diverse languages, % , curated from Common Crawl\footnote{\url{https://commoncrawl.org/}}. 
up-sampling the less-resourced languages to render a more balanced language representation. 
We evaluate the models on multiple multilingual natural language understanding (NLU) tasks, machine translation and a subset of English tasks demonstrated in~\citet{brown2020gpt3}. 
% We evaluate the models on five multilingual tasks across domains (commonsense reasoning, natural language inference, and paraphrasing). 

%We find that simply using English prompts yields competitive zero- and few-shot learning performance with non-English examples, which is a strong indication of the cross-lingual capability of such models. 
We found \Ours demonstrate strong cross-lingual capability where using English prompts together with  non-English examples yields competitive zero- and few-shot learning performance.  Our largest model (\model{\Oursns}{7.5B}) % in particular 
achieves strong zero- and few-shot learning performance on language completion and inference tasks (e.g. XStoryCloze: 65.4\% 0-shot, 66.5\% 4-shot; XNLI: 46.3\% 0-shot, 47.3\% 4-shot). It also establishes a new state-of-the-art on few-shot machine translation across a large number of language pairs in the FLORES-101 benchmark~\cite{goyal2021flores}, significantly outperforming the GPT-3 model of comparable size (6.7 billion parameters). 
% We also present a detailed comparison between \model{\Oursns}{7.5B} and a GPT-3 model of comparable size (6.7B parameters). 
% \model{\Oursns}{7.5B} outperforms GPT-3$_\text{6.7B}$ on a diverse set of languages, especially the less-resourced ones, suggesting the effectiveness of cross-lingual pre-training with a more balanced language distribution. % can lead to inclusive multilingual few-shot learners. 
On the other hand, multilingual pre-training causes performance drop on English. On 8 English NLU tasks, \model{\Ours}{7.5B} underperforms \model{GPT-3}{6.7B} by 10.9\% on average in zero-shot learning. \model{GPT-3}{6.7B} also surpasses \model{\Ours}{7.5B} in machine translation on several high-resource language pairs, including WMT-14 \lang{en}$\leftrightarrow$\lang{fr}, WMT-16 \lang{en}$\leftrightarrow$\lang{de} and WMT-19 \lang{en}$\leftrightarrow$\lang{zh}.
% On the other hand, GPT-3$_\text{6.7B}$ performs strongly on English and related high-resource languages (e.g. French and German), surpassing \model{\Oursns}{7.5B} on most English tasks and some language pairs in machine translation (e.g. WMT-14 \lang{en}$\leftrightarrow$\lang{fr}, WMT-16 \lang{en}$\leftrightarrow$\lang{de}, WMT-19 \lang{en}$\leftrightarrow$\lang{zh})~\cite{bojar-EtAl:2014:W14-33,bojar-etal-2016-findings,barrault-etal-2019-findings}. 

We conduct an in-depth analysis of different multilingual prompting approaches and examine cross-lingual transfer through template and demonstration examples respectively. We show that non-English templates sometimes yield unexpected low zero- and few-shot learning accuracy even if they are crafted by native speakers (\S\ref{sec:comparing_multilingual_prompts}). Both using the English template (\S\ref{sec:cross-lingual-prompting}) and adding demonstration examples (\S\ref{sec:cross-lingual-few-shot-learning-exp}) provide effective remedy. However, using demonstration examples from another language often cannot further improve the zero-shot learning performance when a strong prompting language (e.g. Engilsh) is used, which indicates room for improvement in cross-lingual pre-training and in-context transfer approaches. % Our results show that conditioning on demonstration examples from other languages almost always improves zero-shot performance for target languages.    %We also find it is possible to conduct cross-lingual in-context learning in some cases, where the model performance significantly improves over zero-shot learning by conditioning on examples in languages different from the target language. 
\begin{figure*}[t]
    \centering
    \includegraphics[width=0.9\linewidth]{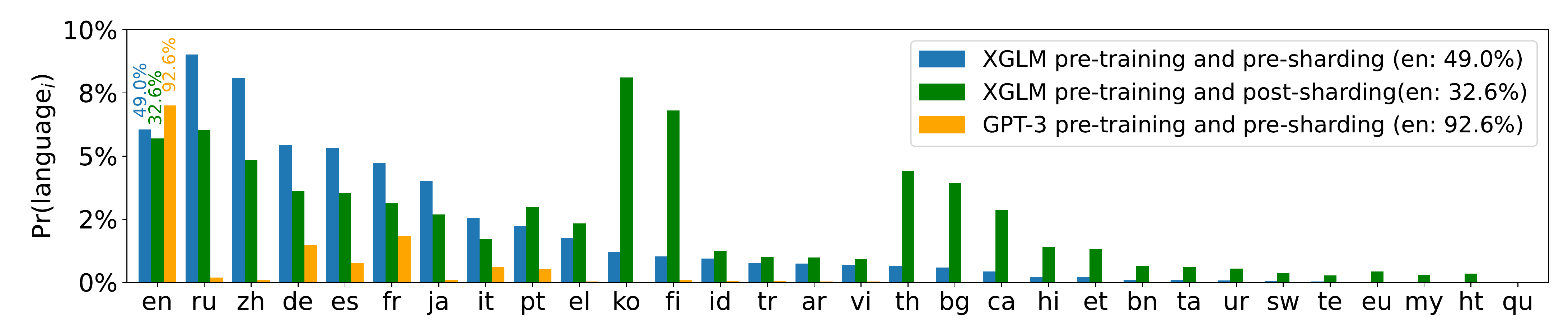}
    \caption{The \% of each language $l$ $(l=1, 2, ..., 30)$ in \Ours's pre-training data pre-upsampling (blue), post-upsampling (green), and its corresponding \% in GPT-3's training data (orange). We truncate the y-axis at 10\% to better visualize the tail distribution.} % Top: number of tokens in \Ours's training data pre-upsampling. Bottom: the probability of a language $l$ $(l=1, 2, ..., 30)$ is selected for pretraining XLM-G or GPT-3 models (blue and red bars); for XLM-G, various up-sampling ratios are applied at the languages level after sharding the data for training (green bars) (the y-axis is truncated at 10\%; and the height of bars for English is not proportional to their real probability). \vic{(4) For bottom figure, clarify the red bars represent only a subset of languages in the GPT-3 corpus (and the ratio values were computed based on all languages).}\xian{Can we make the bars for En the real height in the second plot?}} %\footnote{\url{https://github.com/openai/gpt-3/tree/master/dataset_statistics}} GPT-3 is trained on a superset of the languages in \Ours's pre-training data, and we only show the overlapped ones. 
    \label{fig:pretraining_tokens}
\end{figure*}

\section{Models and Pre-training Data}
\label{sec:model_and_pre_training_data}

% We first present the data, architecture and learning objective used for training \Ours. 

\subsection{Pre-training Data}
\label{sec:pretraining_data}

\paragraph{Language selection and pre-processing.} We extend the pipeline used for mining the {CC100} %\footnote{\url{http://data.statmt.org/cc-100/}} 
corpus~\citep{conneau-etal-2020-unsupervised, wenzek-etal-2020-ccnet} to generate CC100-XL, a significantly larger multilingual dataset covering 68 Common Crawl (CC) snapshots (from \href{http://commoncrawl.org/2013/11/new-crawl-data-available/}{Summer 2013} to \href{https://commoncrawl.org/2020/04/march-april-2020-crawl-archive-now-available/}{March/April 2020}) and 134 languages. % (Table~\ref{tab:cc100_data_table}). 
Our pretraining data include % a typologically diverse set of 30 % representative 
30 languages covering 16 language families.
%%%%%%%%% Last minute cut %%%%%%%%
% We group the languages into four resource tiers: \emph{high},  \emph{medium}, \emph{low} and \emph{extremely-low}, based on the amount of data they have after filtering\footnote{Let $X_l$ be the number of tokens of a language $l$ in the pre-training data, we define the resource tier according to the following criteria \emph{high}: $X_l\geq48\text{B}$, \emph{medium}: $5\text{B}\leq X_l<48\text{B}$, \emph{low}: $0.1\text{B}\leq X_l<5\text{B}$, \emph{extemely-low}: $X_l<0.1\text{B}$.}. 
%Multilingual training data is known to have a heavy-tailed distribution skewed towards high resource languages. To mitigate the potential bias from low resource being drawn in high resource languages\vic{What does this sentence mean?}, 
The natural data distribution is skewed with the number of English tokens being 6 times that of the second largest language. Following previous work on multilingual pre-training \cite{conneau-etal-2020-unsupervised, liu2020multilingual}, we up-sampled the medium and low resource languages to create a more balanced language distribution (Appendix~\ref{sec:data_sources}).\footnote{We inadvertently over-sampled some of the less resourced languages which is reflected in the statistics of \lang{ko}, \lang{fi}, \lang{th}, \lang{bg}, \lang{ca}, \lang{hi}, \lang{et} languages, as shown in Figure~\ref{fig:pretraining_tokens}. We did not ablate the effect of this mistake due to the extreme computational cost. Studying optimal language balancing is an important area for future work.} 
% The sampling probabilities across languages are summarized in Figure \ref{fig:pretraining_tokens}.  % We shard the data into X shards.  As a result, the distribution of pretraining data per language is shown in Figure \ref{fig:pretraining_tokens}.
Figure~\ref{fig:pretraining_tokens} shows the language distribution of our pre-training data before (blue) and after (green) up-sampling. 
% We compare it to the training language distribution of GPT-3~\cite{brown2020gpt3}. % GPT-3 is significantly more English dominant with 92.6\% of GPT-3 training data is in English, while for \Ours the ratio is only 32.6\% after up-sampling for other languages. On average, the \Ours training data has significant coverage for all languages (except for Quechua). % compared to only a few non English representation  (\lang{fr}, \lang{de}, \lang{es}, \lang{it}, \lang{pt}, \lang{ru}, \lang{zh}, \lang{ja}) in the GPT-3 data.  

% \begin{figure}[t!]
%     \centering
%     \includegraphics[width=\linewidth]{figures/pretraining_ratios_30.pdf}
%     \caption{Ratio of tokens (words) by language in XLM-G vs GPT-3 pre\-training data.}
%     \label{fig:pretraining_ratios}
% \end{figure}

\paragraph{Joint sub-word vocabulary.} We process all languages with a joint vocabulary of size 250k created through % Byte Pair Encoding (BPE)~\cite{DBLP:journals/corr/SennrichHB15} 
unigram language modeling~\cite{DBLP:conf/acl/Kudo18},
%\footnote{Our preliminary experiments show that comparing to Byte Pair Encoding (BPE), unigram-LM vocabulary achieves better language representation as measured by fertility~\cite{DBLP:conf/acl/RustPVRG20} and unknown token ratio. It is also much faster to train over the same amount of data.}
using the SentencePiece library~\cite{Kudo2018SentencePiece:Processing}. We train the unigram-LM model using 10 million sentences randomly sampled from the filtered data, according to the multinomial distribution defined in~\citet{lample2019cross} with $\alpha=0.3$. 

\subsection{Models and Training}
% \comment{Xian Myle to fill in}
% \paragraph{Scaling up model size.} 
We train decoder-only causal language models with the Transformer architecture similar to GPT-3 \cite{brown2020gpt3}. % We 
This allows us to study the effect of scaling up model size along both width and depth dimensions. As a result, we compare four models with 564M, 1.7B, 2.9B and 7.5B parameters, respectively. The architecture details are summarized in Table \ref{tab:models}. Our models match that of GPT-3 models\footnote{For \Ours 2.9B we used the optimal depth-to-width parameter allocation for GPT-3 architectures based on rank bottleneck analysis~\cite{DBLP:conf/nips/LevineWSBS20}. This allocation is expected to have improved training efficiency. However, it did not converge for \Ours 7.5B in our experiments, and we fell back to the original GPT-3 setup.} except with the additional embedding parameters from a larger vocabulary. All models are trained for up to 500B tokens, with context length of 2048 tokens. Further training details are described in Appendix \ref{sec:language_model_pre_training_details}.
% \footnote{\punit{Further details are listed in the Model card at \url{}}.} % \comment{TODO: Add ppl scaling with data and model size.}

\begin{table}[ht]
\begin{center}
% \begin{small}
\addtolength{\tabcolsep}{-2.5pt}
\resizebox{.34\textwidth}{!}{
\begin{tabular}{crrrrrrrrrr}
\toprule
& \multicolumn{3}{c}{GPT-3} && \multicolumn{3}{c}{\Ours } &&  \\
\cmidrule{2-4} \cmidrule{6-8} %\cmidrule{14-18}
& \multicolumn{1}{c}{\emph{size}} & % \multicolumn{1}{c}{\emph{cost}} & 
\multicolumn{1}{c}{$l$} & \multicolumn{1}{c}{$h$} & % \multicolumn{1}{c}{$e$} &
& \multicolumn{1}{c}{\emph{size}} & %\multicolumn{1}{c}{\emph{cost}} & 
\multicolumn{1}{c}{$l$} & \multicolumn{1}{c}{$h$} & % \multicolumn{1}{c}{$e$} &
% & \multicolumn{1}{c}{\emph{size}} & \multicolumn{1}{c}{\emph{cost}} & \multicolumn{1}{c}{$l$} & \multicolumn{1}{c}{$h$} & \multicolumn{1}{c}{$e$} &
\\
\midrule
& 125M & 12 & 768 & 
& \multicolumn{3}{c}{---} &
% & 15B & 0.43 & 12 & 768 & 512 &
\\
& 355M & 24 & 1024 & 
& 564M & 24 & 1024 & 
% & 52B & 1.30 & 24 & 1024 & 512 &
\\
& 760M & 24 & 1536 & 
& \multicolumn{3}{c}{---} &
% & \multicolumn{5}{c}{---} &
\\
& 1.3B & 24 & 2048 & 
& 1.7B & 24 & 2048 & 
% & 207B & 4.53 & 24 & 2048 & 512 &
\\
& 2.7B & 32 & 2560 & 
& 2.9B & 48 & 2048 & 
% & \multicolumn{5}{c}{---} &
\\
& 6.7B & 32 & 4096 & 
& 7.5B & 32 & 4096 & 
% & 1.1T & 22.27 & 32 & 4096 & 512 &
\\

\bottomrule
\end{tabular}
}
% \end{small}
\end{center}
\caption{\textbf{Model details}. \emph{size}: number of parameters, $l$: layers, $h$: hidden dimension. Models within the same row % are roughly comparable
have comparable sizes.}
\label{tab:models}
\end{table}
% \paragraph{Training.} Our multilingual models are trained for 500B tokens with a sequence length of 2048 tokens. 

\section{Multilingual In-context Learning}
\label{sec:multilingual_in_context_learning}

\hide{
We perform zero-shot and in-context few-shot learning using XGLM by extending the prompting framework introduced by~\citet{brown2020gpt3} to the multilingual and cross-lingual setting\footnote{Appendix~\ref{sec:task_formulation} provides the formal problem formulation}. % using multilingual natural % 
}
We % extend the evaluation of % GPT-2~\cite{radford2019language} and GPT-3~\cite{brown2020gpt3} to more languages by measuring 
measure the performance of our multilingual language models on % a variety of 
downstream tasks in different languages given the tasks and few-shot demonstrations specified via % text input in multiple languages. % the target language of interest. 
prompts without further parameter updates (Appendix \ref{sec:task_formulation}).
% \begin{table*}[t]
%     \centering
%     \resizebox{.92\textwidth}{!}{%
%     \begin{tabular}{c|c|l|c}
%         Task Category & Dataset & Template & Candidate Verbalizer  \\
%         \midrule
%         \multirow{3}{*}{Reasoning} & \multirow{2}{*}{XCOPA} & (1): \texttt{\{Sentence 1\}} because \mask & \multirow{3}{*}{Identity} \\
%          & & (2): \texttt{\{Sentence 1\}} so \mask & \\
%          & StoryCloze & \texttt{\{Context\}} \mask & \\
%         \midrule
%         NLI & XNLI & \texttt{\{Sentence 1\}}, right? \mask, \texttt{\{Sentence 2\}} & \makecell{\textit{Entailment}: Yes | \textit{Neural}: Also | \textit{Contradiction}: No}\\
%         \midrule
%         QA & ARC & Question: \texttt{\{Question\}} Answer: \mask & Identity \\
%         \midrule
%         Translation & WMT & \texttt{\{Source sentence\}} = \mask & Identity \\
%         \bottomrule
%     \end{tabular}
%     }
%     \caption{English prompts for commensense reasoning, natural language inference (NLI), QA and translation tasks.}
%     \label{tab:en_task_prompts}
% \end{table*}

\begin{table*}[t!]
    \centering
    \resizebox{\textwidth}{!}{%
    \begin{tabular}{c|c|l|c}
        \toprule
        Task Category & Dataset & Template & Candidate Verbalizer  \\
        \midrule
        \multirow{4}{*}{Reasoning} & \multirow{2}{*}{XCOPA} & \textit{cause}: \texttt{\{Sentence 1\}} because \mask & \multirow{4}{*}{Identity} \\
         & & \textit{effect}: \texttt{\{Sentence 1\}} so \mask & \\
         & XStoryCloze & \texttt{\{Context\}} \mask & \\
         & XWinograd & \texttt{\{Context\}} (\textit{with '\_' replaced by} \mask) & \\
        \midrule
        NLI & XNLI & \texttt{\{Sentence 1\}}, right? \mask, \texttt{\{Sentence 2\}} & \makecell{\textit{Entailment}: Yes | \textit{Neural}: Also | \textit{Contradiction}: No}\\
        \midrule
        Paraphrase & PAWS-X & \texttt{\{Sentence 1\}}, right? \mask, \texttt{\{Sentence 2\}} & \makecell{\textit{True}: Yes | \textit{False}: No}\\
        \midrule
        Translation & WMT, FLORES-101 & \texttt{\{Source sentence\}} = \mask & Identity \\
        \bottomrule
    \end{tabular}
    }
    \caption{Handcrafted (English) prompts for multilingual natural language understanding and translation tasks.}
    \label{tab:all_multi_task_prompts}
\end{table*}

{
\setcode{utf8}
\begin{table*}[ht]
    \centering
    \resizebox{\textwidth}{!}{%
    \setlength{\tabcolsep}{4pt}
    \begin{tabular}{c|c|l|ccc}
        \toprule
         \multirow{2}{*}{Task} & \multirow{2}{*}{Lang} & \multirow{2}{*}{Template} & \multicolumn{3}{c}{Candidate Verbalizer} \\
         & & & Entailment & Contradiction & Neutral \\
         \midrule
         \multirow{3}{*}{XNLI} & en & \texttt{\{Sentence 1\}}, right? \mask, \texttt{\{Sentence 2\}} & Yes & No & Also\\
         & zh & \texttt{\{Sentence 1\}}\begin{CJK*}{UTF8}{gbsn}\end{CJK*}\mask\begin{CJK*}{UTF8}{gbsn}，\end{CJK*}\texttt{\{Sentence 2\}} &  \zhtext{由此可知，} & \begin{CJK*}{UTF8}{gbsn}所以，不可能\end{CJK*} & \begin{CJK*}{UTF8}{gbsn}同时，\end{CJK*}\\
         & es & \texttt{\{Sentence 1\}}, ¿verdad? \mask, \texttt{\{Sentence 2\}} & Sí & No & Además\\
         % zh & \texttt{\{Sentence 1\}}\begin{CJK*}{UTF8}{gbsn}，对吗？\end{CJK*}\mask\begin{CJK*}{UTF8}{gbsn}，\end{CJK*}\texttt{\{Sentence 2\}} &  \begin{CJK*}{UTF8}{gbsn}是的\end{CJK*} & \begin{CJK*}{UTF8}{gbsn}不\end{CJK*} & \begin{CJK*}{UTF8}{gbsn}而且\end{CJK*}\\
         % ur & \texttt{\{Sentence2\}} \RL{،}\mask \RL{، ٹھیک ہے؟ }\texttt{\{Sentence1\}} & \RL{جی ہاں} & \RL{نہیں}  & \RL{بھی}\\
         \midrule
         \multirow{2}{*}{XCOPA} & en & \textit{cause}: \texttt{\{Sentence 1\}} because \mask | \textit{effect}: \texttt{\{Sentence 1\}} so \mask & \multicolumn{3}{c}{\multirow{2}{*}{Identity}}\\
         & zh & \textit{cause}: \zhtext{因为}\mask\zhtext{，所以}\texttt{\{Sentence 1\}} | \textit{effect}: \zhtext{因为}\texttt{\{Sentence 1\}}\zhtext{，所以}\mask & & & \\
         \bottomrule
    \end{tabular}
    }
    \caption{Handcrafted multilingual prompts. English (\lang{en}), Chinese (\lang{zh}) and Spanish (\lang{es}) for XNLI; English (\lang{en}) and Chinese (\lang{zh}) for XCOPA.}
    \label{tab:multilingual_prompts}
\end{table*}
}

\subsection{Multilingual and Cross-lingual Prompting}
\label{sec:multilingual_prompts}
% In a multilingual setting, we can prompt $\mM$ with languages other than English. 
Previous work on English in-context learning has shown that % language model 
performance heavily depends on the prompt construction, and it is challenging to find the optimal prompt for a given language model~\cite{DBLP:conf/acl/GaoFC20,DBLP:journals/corr/abs-2105-11447}.
% To perform few-shot learning in multiple languages, a critical step is to obtain the prompts for the non-English examples. % Searching for the optimal prompts 
% Prompt engineering is known to be challenging for English language models. 
This problem is further complicated in the multilingual setting, where we need to find the optimal prompts for examples in different languages. 

In this work, we consider three approaches for obtaining the prompts for non-English tasks.

\paragraph{Handcrafting prompts.} The first approach is to ask native speakers of the target language to handcraft the prompts. Prompts created this way are expected to have the most natural surface form. % \footnote{As we show in \S\ref{sec:comparing_multilingual_prompts}, native prompts, while being intuitively better, still underperform the English templates for most languages on existing benchmarks.}. 
However, language expertise is expensive and we further consider two alternatives.

\paragraph{Translating from English prompts.} We assume high-quality prompts of a task can be easily sourced in English% are available in English, thanks to recent work on massive-scale English prompt sourcing
\cite{sanh2021multitask,mishra2021cross}. % \footnote{we now have access to English prompts for a large number of NLP tasks. However, manually authored English prompts are not necessarily optimal and constructing multilingual prompts from them may lead to sub-optimal performance. We leave better multilingual prompt-engineering and prompt-free few-shot learning approaches to future work~\cite{min2021metaicl}.}. 
% We characterize prompts as \emph{verbal} and \emph{non-verbal}. 
Non-verbal prompts do not contain words in any particular language % in its constant portion 
(e.g. the StoryCloze  and WMT prompts shown in Table~\ref{tab:all_multi_task_prompts}), % and are hence language-universal. % \mona{I don't understand what this means? verbal vs non verbal? The examples are not indicative. Do you mean generic underspecified as in universal prompts (always in English) vs. localized language (verbalized per language)?} 
while verbal prompts have different realizations in different languages (Table~\ref{tab:multilingual_prompts}). % For example, Table~\ref{tab:multilingual_prompts} shows our XNLI prompt realized in three languages: English, Chinese and Urdu. 
If the prompt is non-verbal, we simply apply it to the other languages. If the prompt is verbal, we translate it into the other languages using automatic translation APIs. % While automatic translation is imperfect, especially in the lower-resource regime, it is much cheaper compared to professional translation or writing the prompts from scratch. % \vic{Discuss limitation of hand-crafted prompts.} 

\paragraph{Cross-lingual prompting.} 
% While automatic prompt translation provides a relatively cheap way to obtain prompts in many more languages, the accuracy of the state-of-the-art machine translation systems significantly drops for medium- and low-resource languages. On the other hand, 
% Large language models have demonstrated competitive cross-lingual capabilities, even with only a small portion of non-English text in their pre-training data~\cite{conneau-etal-2020-unsupervised,DBLP:journals/corr/abs-1911-01464,brown2020gpt3,DBLP:journals/corr/abs-2109-07684}. We hence 
We consider the third approach which directly applies the prompts in English (or another high-resource language) to non-English examples. We expect this approach to be competitive, as a result of the cross-lingual capability of the model after being trained on a diverse set of languages.

\subsection{Learning from Cross-lingual Demonstrations}
\label{sec:cross-lingual-few-shot-learning}
The cross-lingual nature of multilingual language models further enable the possibility of learning from a different language in context without parameter updates. To do so we simply append examples from another language as the demonstration examples in the language model context. Such capability enables cheap transfer from high-resource languages to the low-resource target languages. % As we show in \S\ref{sec:cross-lingual-few-shot-learning-exp}, with the vanilla prompting approach, only a small number of languages can benefit from in-context transfer. 

\section{Experiments and Results}
% We evaluate our models using both in-domain perplexity (\S~\ref{sec:model_scaling}) and zero-shot and few-shot learning performances on the downstream tasks. 
\subsection{Tasks}
\label{sec:tasks}

We evaluate the zero-shot and in-context few-shot learning capabilities~\cite{brown2020gpt3} of XGLM on a spectrum of downstream tasks (Table~\ref{tab:tasks}). 
\begin{table*}[t!]
\centering
\resizebox{.9\textwidth}{!}{%
\begin{tabular}{l l r r r r r}
\toprule
Task Category & Task & $|$Train$|$ & $|$Dev$|$ & $|$Test$|$ & Non-En Sets & $|$Lang.$|$ \\ \midrule
% \multicolumn{8}{c}{Single sentence classification} \\ \midrule
% \multirow{2}{*}{\rotatebox[origin=b]{90}{\parbox{1.0}{Single seq. clas.}}} & \multirow{2}{*}{CliC} & \multirow{2}{*}{30,521} & \multirow{2}{*}{4,181} & \multirow{2}{*}{1,692--8,621} & \multirow{2}{*}{3} & \multirow{2}{*}{intent} & \multirow{2}{*}{ac.} & \multirow{2}{*}{assistant} \\
% \\ \midrule
% & MLDoc & &  &  & 8 & topic & ac. & news \\ \midrule
% \multicolumn{8}{c}{Sentence pair classification} \\ \midrule
% \multirow{2}{*}{\rotatebox[origin=b]{90}{\parbox{1.0}{Sent. pair clas.}}} 
\multirow{3}{*}{\makecell[l]{Reasoning}} & XStoryCloze {\small $\spadesuit$} & 361 & -- & 1,511 & translations & 11 \\
& XCOPA {\small $\spadesuit$}~\cite{DBLP:conf/emnlp/PontiGMLVK20} & 33,410+400 & 100 & 500 & translations & 11 \\ % \midrule
& XWinograd~\cite{DBLP:conf/acl/TikhonovR21} & -- & -- & 2,325$^\dagger$ & translations & 6 \\
\hdashline
NLI & XNLI {\small $\spadesuit$}~\cite{DBLP:journals/corr/abs-1809-05053} & -- & 2,490 & 5,010 & translations & 15 \\ % \midrule
\hdashline
Paraphrase & PAWS-X~\cite{DBLP:journals/corr/abs-1908-11828} & -- & 2,000 & 2,000 & translations & 7 \\
\bottomrule
\end{tabular}
}
\caption{Multilingual tasks used in our few-shot learning evaluation. All tasks use accuracy as the evaluation metrics. {$^\dagger$: In XWinograd, each language has different number of test examples: \lang{en}: 2,325, \lang{jp}: 959, \lang{ru}: 315, \lang{pt}: 263. $^\ddagger$: We use the COPA release in SuperGLUE~\cite{wang2019superglue}. $\spadesuit$: Held-out tasks.}}
%\footnote{We include COPA when reporting the results of XCOPA for the rest of the paper. This way all our multilingual tasks include English as an evaluation language.}
% , the $|$Train$|$, $|$Dev$|$, $|$Test$|$ statistics were sums over all languages
\label{tab:tasks}
\end{table*}

\paragraph{Multilingual tasks.} We select four multilingual tasks spanning commonsense reasoning (XCOPA), anaphora resolution (XWinograd), natural language inference (XNLI) and paraphrasing (PAWS-X). We also created a new dataset, XStoryCloze, by professionally translating the validation split\footnote{ We further split the translated data into train and test (20\% vs. 80\%, respectively) for each language, keeping the parallel sentence mapping in both splits.} of the English StoryCloze dataset (Spring 2016 version) to 10 other typologically diverse languages (\lang{ru}, \lang{zh} {Simplified}, \lang{es} {Latin American}, \lang{ar}, \lang{hi}, \lang{id}, \lang{te}, \lang{sw}, \lang{eu}, \lang{my})\footnote{For all of our multilingual NLU datasets, the non-English sections of the data are (professionally) translated from the English section. Despite being the dominant approach adopted by the community~\cite{DBLP:conf/emnlp/RuderCBSFF00GNJ21}, it was previously shown to introduce data artifacts that inflate the measured cross-lingual transfer of models~\cite{DBLP:conf/emnlp/ArtetxeLA20}. % However, this , given most tasks are first available in English. 
We leave collecting native multilingual datasets that include  non-English data as future work, and strongly encourage the community to also adopt this practice.}. %, with the split consistent across all languages. 
In addition, we evaluate our models on machine translation (\S\ref{sec:translation}) and multilingual social value tasks (Appendix~\ref{sec:hate-speech-detection}). %, namely Hate Speech Detection and Occupation Identification.

\paragraph{English tasks.} We also evaluate our models on English commonsense reasoning and QA, a subset of benchmark tasks used by~\citet{brown2020gpt3}, % (Table~\ref{tab:tasks}) 
and compare the performance to state-of-the-art English-centric few-shot learning models. The tasks are detailed in Table~\ref{tab:en_tasks}.

% \vic{Shorten model selection}
\subsection{Setup}
\label{sec:model_selection}
\paragraph{Scoring function and calibration.}
% ~\citet{brown2020gpt3} demonstrate that the choice of language model scoring functions can significantly impact the performance of in-context learning, and the optimal scoring function is task specific. 
% In addition, other works~\cite{DBLP:conf/icml/ZhaoWFK021,J1WhitePaper} have shown that contextual calibration and normalization techniques can significantly improve the prediction accuracy for certain tasks.  
We follow the guidelines suggested by~\citet{DBLP:journals/corr/abs-2105-11447} and adopt a \emph{cross-task generalization} setting~\cite{DBLP:conf/iclr/TriantafillouZD20} to select our scoring function. % and to check if additional calibration should be applied. 
We reserve three held-out tasks (XNLI, XCOPA and XStoryCloze) to perform the selection based on their development set performance, and directly apply the selected settings to the rest of the tasks. % (XWinograd and PAWS-X). 
In the end, we use the averaged per-token log-probabilities ignoring the common prefix of different candidates as the scoring function for all multilingual tasks with no additional calibration or normalization. Appendix~\ref{sec:scoring_function} details the selection.

\paragraph{Few-shot learning evaluation.} We focus on benchmarking the 0- and 4-shot learning performance of the models on all tasks. For cross-lingual demonstration (\S\ref{sec:cross-lingual-few-shot-learning-exp}), scaling law (\S\ref{sec:model_scaling}) and translation (\S\ref{sec:translation}) we also reported 1-shot and 32-shot performance. We report the average results across 5 runs, randomly sampling a different set of few-shot examples each time. 
% For tasks with a training set, we sample the few-shot examples from the training set; for tasks with no training set, we sample from the dev set and report results on the test set; for development-set experiments on XNLI and XCOPA, we sample few-shot examples from the test set, since these two tasks do not have the training sets for all languages. 
Without further specification, we use few-shot examples in the same language as the target example.
Appendix~\ref{sec:eval_prototcol} details our complete evaluation protocol.

\subsection{Comparing Prompting Approaches}
\label{sec:comparing_multilingual_prompts}
We first compare different multilingual prompting approaches proposed in \S\ref{sec:multilingual_prompts} using \model{\Oursns}{7.5B} on XNLI and XCOPA\footnote{The original XCOPA release~\cite{ponti-etal-2020-xcopa} does not contain the English section. We added the English release from SuperGLUE~\cite{wang2019superglue} to facilitate cross-lingual experiments.}. Native speakers among the authors handcrafted\footnote{The native speakers were instructed to create a prompt that convert the task into a natural cloze-style question in their native language with no further restrictions.} the prompts for the following tasks: XNLI (\lang{en}, \lang{zh}, \lang{es} and \lang{hi}) and XCOPA (\lang{en}, \lang{zh}), as shown in Table~\ref{tab:multilingual_prompts}. We compare the performance of these human-written prompts to English prompts, machine-translated (MT) prompts and human-translated (HT) prompts. 

Table~\ref{tab:xnli_prompt_comparison} and~\ref{tab:xcopa_prompt_comparison} show the performance of different prompting approaches\footnote{ Appendix~\ref{sec:comparing_multilingua_prompting_appendix} provides the comparison between English prompts and the \texttt{MT} and \texttt{HT} prompts on the complete dev sets of XNLI and XCOPA.}. English templates perform the best on average across languages for both tasks except for the 4-shot setting of XCOPA, where it slightly underperforms the machine translated templates. % We find this result surprising given the cross-lingual nature of the template. 
On the XNLI task, the English template significantly improves the performance of Chinese (\lang{zh}) and Hindi (\lang{hi}) over their native templates and translated templates. Similar trends are observed for Thai (\lang{th}) and Swahili (\lang{sw}) on XCOPA\footnote{The strong performance of English templates may be partially contributed to the fact that the non-English evaluation data on XNLI and XCOPA are obtained from translation. Testing how well the English templates perform on native non-English test sets is an interesting future work.}. For both tasks there exist languages where the native templates strongly outperforms the English templates (Spanish (\lang{es}) for XNLI and Chinese for XCOPA), indicating significant room for future work on language-specific prompt engineering. 

\begin{table}[]
    \centering
    \setlength{\tabcolsep}{4pt}
    \resizebox{.49\textwidth}{!}{
    \begin{tabular}{lrrrrr}
    \toprule
    Temp. & \multicolumn{1}{l}{en} & \multicolumn{1}{l}{zh} & \multicolumn{1}{l}{es} & \multicolumn{1}{l}{hi} &  Avg \\
    \midrule
    En (\texttt{HW}) & \highest{50.8/50.6} & \highest{48.5/47.7} & 37.5/44.4 & \highest{44.0/45.5} & \highest{45.2/47.0} \\
    Zh (\texttt{HW}) & 33.5/35.5 & 33.5/36.4 & 34.5/34.8 & 36.0/34.0 & 34.4/35.1 \\
    Es (\texttt{HW}) & 39.2/49.9 & 44.8/45.3 & \highest{46.2/48.2} & 41.5/43.5 & 42.9/46.7 \\
    Hi (\texttt{HW}) & 45.0/43.5 & 39.5/41.0 & 34.2/40.5 & 36.2/40.5 & 38.8/41.4 \\
    \hdashline
    Multi. (\texttt{HW}) & 50.8/50.6 & 33.5/36.4 & \highest{46.2/48.2} & 36.2/40.5 & 41.7/43.9 \\
    Multi. (\texttt{MT}) & 50.8/50.6 & 35.8/39.5 & 36.5/45.0 & 41.0/39.9 & 41.0/43.8 \\
    Multi. (\texttt{HT}) & 50.8/50.6 & 38.5/41.2 & 46.0/48.1 & 37.5/38.9 & 43.1/44.7 \\
    \bottomrule
    \end{tabular}
    }
    \caption{0/4-shot performance of \model{\Oursns}{7.5B}, evaluated on the first 400 examples of XNLI (development set in \lang{en}, \lang{zh}, \lang{es} and \lang{hi}) using different prompting approaches. Top: all inputs are instantiated with templates in the language specified in column 1. Bottom: all inputs are instantiated with templates in the same language as themselves. \texttt{HW}: human-written. \texttt{MT}: machine-translated. \texttt{HT}: human-translated.}
    \label{tab:xnli_prompt_comparison}
\end{table}

\begin{table}[]
    \centering
    \setlength{\tabcolsep}{4pt}
    \resizebox{.49\textwidth}{!}{
    \begin{tabular}{lrrrrr}
    \toprule
    Temp. & \multicolumn{1}{l}{en} & \multicolumn{1}{l}{zh} & \multicolumn{1}{l}{th} & \multicolumn{1}{l}{sw} &  Avg \\
    \midrule
    En (\texttt{HW}) & \highest{69.0/73.2} & 63.0/66.8 & \highest{53.0/57.4} & \highest{54.0/58.2} & \highest{59.8}/63.9 \\
    Zh (\texttt{HW}) & 63.0/71.0 & \highest{69.0/67.6} & 50.0/57.8 & 47.0/54.2 & 57.2/62.6 \\
    \hdashline
    Multi. (\texttt{HW}) & 69.0/73.8 & \highest{69.0/67.6} & -- & -- & -- \\
    Multi. (\texttt{MT}) & 69.0/73.8 & 62.0/68.4 & 48.0/56.6 & 51.0/60.2 & 57.5/\highest{64.8} \\
    \bottomrule
    \end{tabular}
    }
    \caption{0/4-shot performance of \model{\Oursns}{7.5B}, evaluated on XCOPA (development set in \lang{en}, \lang{zh}, \lang{th} and \lang{sw}). }
    \label{tab:xcopa_prompt_comparison}
\end{table}

\subsection{Cross-lingual Transfer through Templates}
\label{sec:cross-lingual-prompting}

\begin{table*}[t!]
\centering
\scalebox{0.68}{
\begin{tabular}{l|ccccc|ccc|c|ccc}
\toprule
  &   \multicolumn{9}{c|}{high}           &   \multicolumn{2}{c|}{medium} & low \\
 %\cmidrule{2-10}
 &   \multicolumn{8}{c}{en}  &       ru &       tr &    ar &   hi \\
 \midrule
 & \multicolumn{5}{c|}{medium} & \multicolumn{3}{c|}{low} &   medium &      \multicolumn{3}{c}{low} \\
 
 prompt & bg &    el &   th &    tr &    vi &   hi &    sw &   ur &       bg &       ur &    sw &   ur \\
 \midrule
 Same-lang & \bf 2.55 & \bf 0.98 & \bf 2.16 &  \bf 1.27 & \bf 2.23 & \bf 2.51 & -0.69 & \bf 1.21 &    -2.49 &    -0.38 & -1.64 & \bf 3.31 \\
 \midrule
Source-lang  &  -4.59 & -2.44 & \bf 7.87 & -4.97 & -1.08 & \bf 2.01 & -1.15 & \bf 7.42 &    -1.43 &     \bf  6.67 & -5.86 & \bf 2.31 \\

\bottomrule
\end{tabular}
}
\caption{Learning from cross-lingual demonstrations on XNLI, evaluated on the test set. The results are the absolute improvement over the zero-shot performance for the evaluated language using human-translated prompts. The first language group refers to the source language and the second one refers to the target language. \textit{Same-lang} refers to a setting there the template is in the example language and \textit{source-lang} refers to a setting where the template is only in the source language. 
}
     \label{tab:crosslingual_xnli}
\end{table*}

\begin{table*}[t!]
\centering
\scalebox{0.72}{
\begin{tabular}{l|p{0.5\textwidth}|p{0.62\textwidth}}
\toprule
 & Source prompt (instantiated) & Target prompt (instantiated) \\
 \midrule
 \emph{Same-lang} & \entext{The best thing that may be said of Podhoretz and Decter is that their biological clocks can't have many more minutes left on them\underline{, right? Yes,} Decter is old.} & \vitext{Vâng, tôi thậm chí không nghĩ về điều đó, nhưng tôi đã rất thất vọng, và, tôi lại nói chuyện với anh ta lần nữa\underline{, đúng không? Đúng,} tôi đã không nói chuyện với anh ta nữa.}\\
\midrule
\emph{Source-lang} & \entext{The best thing that may be said of Podhoretz and Decter is that their biological clocks can't have many more minutes left on them\underline{, right? Yes,} Decter is old.} & \vitext{Vâng, tôi thậm chí không nghĩ về điều đó, nhưng tôi đã rất thất vọng, và, tôi lại nói chuyện với anh ta lần nữa}\underline{\entext{, right? Yes,}} \vitext{tôi đã không nói chuyện với anh ta nữa.}\\
\bottomrule
\end{tabular}
}
\caption{XNLI example prompts for cross-lingual transfer from \textit{\entext{English (en)}} to \textit{\vitext{Vietnamese (vi)}}, with the \textit{same-language} and \textit{source-language} settings. The \underline{underlined} text shows the verbalized part of the prompt.} 
\label{tab:cross_lingual_code_switching_example}
\end{table*}

We further examine if the ability of % acting as a universal prompting language 
universal prompting is English specific, and in addition, what characterize a language pair for which cross-lingual prompting can work. To this end, we apply each of the human-written non-English templates to the rest of the languages. As shown in Table~\ref{tab:xnli_prompt_comparison} and~\ref{tab:xcopa_prompt_comparison}, using the Spanish prompt yields competitive 0- and 4-shot performance across all languages, with the 4-shot average performance being comparable to that of the English template. The Hindi template also achieves significantly above random performance on the XNLI tasks for most languages (especially \lang{en}). The Chinese template, however, achieves close-to-random performance for all languages on XNLI, as well as close-to-random for Thai (0-shot) and Swahili (0-shot) on XCOPA. We hypothesize that the common sub-tokens and the amount of code-switching text in the pre-training data play a significant role in enabling cross-lingual prompting. And in general, high-resource languages with large amounts of pre-training data and vocabulary overlap with other languages act as better universal prompting languages. We leave a more systematic verification of this hypothesis to future work.

\subsection{Cross-lingual Transfer through Demonstration Examples}
\label{sec:cross-lingual-few-shot-learning-exp}
We examine the capabilities of learning from cross-lingual demonstration examples (\S\ref{sec:cross-lingual-few-shot-learning}) of \model{\Oursns}{7.5B} on XNLI. % We condition the test examples in the target language on the instantiated prompts of 32 few-shot examples from a given source language. 
%\ves{I don't understand the next sentence} 
%For XNLI we use crafted templates for forming the in-context learning. 
% XNLI requires framing the prompts using a template with text fields and candidate verbalizers (Table \ref{tab:multilingual_prompts}). 
% This allows us to also test the impact of code-switching between the language of the template and verbalizer, and the language of the examples as described at the end of Section \ref{sec:multilingual_prompts}. In particular, we 
% We hence 
We examine two settings for each train-eval language pair: \textit{same-language-prompting}, where the prompt templates and the example are in the same language, and \textit{source-languauge-prompting} where the prompt templates for both the demo and test examples are in the source language. We use the human-translated prompts for \textit{same-language-prompting}.
% \mona{Not understanding the previous paragraph. I suggest changing source language template to cross lingual template.} 
%\vic{Should we mention these two settings at the beginning of section 6.3? Then we can say we evaluated XNLI in two settings, XCOPA in XX setting, for XStoryCloze both settings are the same? In the current flow, it is unclear which setting was used for XCOPA.}

% Similar to XCOPA, 
Table \ref{tab:crosslingual_xnli} shows results on a subset of language pairs of XNLI, where we evaluate transfer through demonstration examples from % high and low resource settings 
in-context demonstration examples from high-resource languages to lower-resourced ones, and between languages that are typologically similar. We report the difference between the 32-shot learning results and the 0-shot learning results. The non-English templates in this experiment are obtained via human-translation. While they typically underperform the in-language few-shot setting (Figure~\ref{fig:crosslingual-xnli-over-ht}), most cross-lingual few-shot settings significantly improve over the 0-shot setting for the target language. Bulgarian is an exception as it does not benefit from Russian despite being in the same language family. %, with the performance drop being smaller with \textit{source-language-prompting}. 
Another language that does not work well in the cross-lingual settings is Swahili (\textit{low} resource), for which we examined transfer from English (\textit{high} resource) and Arabic (\textit{medium} resource). In contrast, Thai (\textit{medium}) and Urdu (\textit{low} resource) significantly benefit from cross-lingual demonstrations% And the source-language setting works even better with English (to Thai and Urdu) and Turkish (to Urdu) as the source languages % when very different medium to high resource level languages like 
\footnote{Both Thai and Urdu obtained close-to-random zero-shot learning performances using the translated templates, which might make them easier to be further improved. Besides, there is inherent code switching in these languages (English presence in Thai and Urdu both lexical and morphological). Turkish and Arabic also have influence on Urdu. We hypothesize that these factors also positively impacted the cross-lingual in-context learning performance.}. 

We also observed % that demonstration examples in a different language generally performs significantly worse than the examples from the target language. Besides, 
{the benefit of cross-lingual transfer from demonstration examples is generally canceled if a better prompt (e.g. the English prompt) is used for the target language}. We report the crosslingual demonstration experiments between all pairs of languages for XNLI, XCOPA and XStoryCloze and provide more discussion in Appendix~\ref{sec:cross-lingual-in-context-learning-full}. % In addition, transfer from demonstration examples in another language in general significantly underperforms transfer from those in the same language (Table~\ref{tab:xnli_template_analysis}).

\begin{table*}[t]
    \centering
    \scalebox{0.64}{
        \begin{tabular}{ll|lrrrrr|lrrlll|lll|l}
        \toprule
             & & \multicolumn{6}{c|}{high} & \multicolumn{6}{c|}{medium} & \multicolumn{3}{c}{low} & Avg.\\
             model & \#\ shot &   en &   de &   es &   fr &   ru &   zh & ar &       bg &   el &   th &   tr &   vi &   hi &   sw &   ur &  \\

        \midrule
        \model{GPT-3}{6.7B} & 0 & 
            \bf 55.4 & 36.8 & 37.0 & \bf 51.2 & 44.8 & 42.6 &
            38.5 & 42.9 & 38.8 & 38.4 & 40.6 & 41.3 & 
            36.5 & 34.6 & 34.5 &
            40.9 \\
             & 4   & 
            53.0 & \bf 46.4 & \bf 48.5 & 48.3 & 44.3 & 45.8 &
            38.2 & 41.7 & 42.1 & 36.8 & 38.7 & 42.3 &
            34.3 & 33.7 & 34.5&
            41.9 \\
        \midrule
        % \model{\Oursns}{6.7B} \textit{EN-only}  & 0 &
        %     54.6 &	39.8 &	38.8 &	39.2 &	34.2 &	35.6 &
        %     36.1 &	34.6 &	33.6 &	33.5 &	34.3 &	34.2 &
        %     33.6 &	34.6 &	33.9 &
        %     35.7 \\
        %     & 4 & 
        %     \bf 54.1 &	40.9 &	42.3 &	42.5 &	35.3 &	34.7 &
        %     34.3 &	34.7 &	33.5 &	32.9 &	34.5 &	37.0 &
        %     33.1 &	34.2 &	33.2 &
        %     37.1 \\
        % \midrule
        \model{\Oursns}{7.5B} & 0   & 
            55.3 & \bf 42.3 & \bf 39.1 & 50.8 & \bf 48.4 & \bf 44.8 &  
            \bf 48.1 & \bf  49.1 & \bf 46.4 & \bf 46.8 & \bf 45.5 & \bf 47.6 &
            \bf 43.4 & \bf 45.5 & \bf 41.9 &
            \bf 46.3 \\
             & 4   &  
            52.6 & 45.6 & 45.8 & \bf 49.4 & \bf 48.6 & \bf 48.8 &
            \bf 46.4 & \bf 48.9 & \bf 48.7 & \bf 46.6 & \bf 45.4 & \bf 48.5 & 
            \bf 46.8 & \bf 44.5 & \bf 43.4 &
            \bf 47.3 \\
             
        \midrule
        \midrule
        Translate + \xlmgen & 0 &
            54.6 & 53.7 &	54.5 &	53.9 &	52.0 &	52.6 &
            52.0 & 53.4 & 53.5 & 50.6 &	53.3 &	52.6 &
            \textit 50.7 & 51.3 &	48.7 &
            52.5 \\
                            & 4 &
            54.1 & 52.4 &	49.2 &	50.3 &	53.2 &	51.1 &
            50.5 & 53.7 &	53.0 & 48.2 &	51.8 &	52.8 &
            49.8 & 50.2 &	47.2&
            51.2 \\
        \bottomrule
        \end{tabular}
    }
    % \vic{Add translate-test baseline average copying over 6.7B En En performance.}
    % [Daniel] I already added EN entry to the avg. Copied over the EN numbers to make it less confusing
    \caption{Comparison of different models on XNLI.}
    \label{tab:comp_xnli}
\end{table*}

\begin{table*}[t]
    \centering
    \setlength{\tabcolsep}{2pt}
    \scalebox{0.68}{
        \begin{tabular}{lc|rrrr|rr|rrr|rr|c||r|rrrrr|rrr|rr|c}
        \toprule
         &  & \multicolumn{12}{c}{XStoryCloze} & \multicolumn{12}{c}{XCOPA} \\
         \cmidrule(lr){3-14}\cmidrule(lr){15-26}
            %  & {} & \multicolumn{11}{l}{accuracy::mean} \\
             & \# & \multicolumn{4}{c}{high} & \multicolumn{2}{c}{medium} & \multicolumn{3}{c}{low} & \multicolumn{2}{c}{ex-low} & Avg. &
             \multicolumn{1}{c}{high} & \multicolumn{5}{c}{medium} & \multicolumn{3}{c}{low} & \multicolumn{2}{c}{ex-low} & Avg. \\
        %\midrule
           model & shot &     en &   es &   ru &   zh &   ar &      id &   hi &     sw &   te &     eu &        my &    &      zh &   id &       it &   th &   tr &   vi &   et &   sw &   ta &            ht &   qu &  \\
       
        \midrule
       
        \model{GPT-3}{6.7B} & 0  &   73.4 & 62.4 & 56.9 & 55.8 & 48.4 &  56.6    & 50.1 & 49.4 & 52.8 & 51.2 &          49.5 & 55.1 &  55.0 & 60.2 & \bf 61.6 & 53.6 & 53.4 & 52.8 & 50.8 & 52.2 & 55.0 & 51.8 & \bf 50.0 & 54.2\\
             
             & 4  &          74.4& 62.2 & 56.4 & 54.7 & 47.7 &  55.4    & 49.6 & 49.3 & 52.8 & 51.1 &          49.5 & 54.8 &  57.8 & 60.8 & 64.5  & 54.2 & 52.9 & 54.8 & 51.8 & 52.0 & 54.9 &  51.5 & \bf 49.7 & 55.0 \\
        \midrule  
        
        % \model{\Oursns}{6.7B} \textit{EN-only}  & 0&	\bf 81.2 &	62.4 &	53.7 &	54.0 &	53.1 &	54.5 &	52.8 &	53.9 &	54.5 &	52.3 &	51.3 & 56.7 \\ 
        %                              & 4&	\bf 82.6 &	61.9 &	53.6 &	54.2 &	53.4 &	53.5 &	53.2 &	53.6 &	54.5 &	51.4 &	52.4 & 56.8 \\
        
        % \midrule
        
        \model{\Oursns}{7.5B} & 0  &           75.0 & \bf 68.1 & \bf 71.0 & \bf 66.6 &   \bf 58.3 & \bf 70.1 &  \bf 60.9 &   \bf 65.0 & \bf 61.7 &   \bf 62.3 &       \bf 60.7  & \bf 65.4 & \bf 62.4 &   \bf 66.6  &  60.8  & \bf 56.8 & \bf 56.8 & \bf 61.4 & \bf 61.6 & \bf 57.6 & \bf 56.2 &  \bf 57.0 & 47.4 & \bf  58.6 \\
            
             & 4  &            75.9 & \bf 69.2 & \bf 72.4 & \bf 67.7 &  \bf 59.8 &  \bf 70.8 &  \bf 62.5 &   \bf 65.2 & \bf 63.4 &     \bf 63.8 &     \bf 61.2 & \bf 66.5 & \bf 67.2  & \bf 68.9 & \bf 69.2 & \bf 62.0 & \bf 58.5 & \bf 65.6 & \bf 65.9 & \bf 62.9 & \bf 56.3 &          \bf 58.9 & 47.1 & \bf 62.0\\
             
        \midrule
        \midrule
        Translate & 0 & 81.2 & 75.6 &	75.4 &	72.9 &	71.5 &	71.2 &	70.5 &	70.0 &	66.9 &	70.5 &	72.7 & 72.6 &	75.0 &	73.2 &	76.0  & 53.8 &	72.4 &	72.2 &	72.4 &	63.8 &	67.2 &	65.0 & -  & 67.4$^\dagger$ \\
                + \xlmgen            & 4 & 82.6 & 75.0 &	75.3 &	73.1 &	71.8 &	72.0 &	71.6 &	71.0 &	68.4 &	72.2 &	72.0 & 73.2 &	78.5 & 	75.8 &	80.6  & 57.7 &	73.7 &	76.0 &	73.6 &	67.2 &	69.9 &	67.0 & - & 70.0$^\dagger$ \\
             
        \bottomrule
        \end{tabular}
    }
    \caption{Comparison of different models on XStoryCloze and XCOPA. $^\dagger$Google Translation API is not available for \lang{qu}. For the averaged translate-test results we % used the results without translation ("\Ours 6.7B EN-only" row) 
    directly used the \xlmgen model for \lang{qu} entry.} % \xian{double check: our 6.7B En model outperforms GPT-3 by large margin on En. I've rerun the curie eval on 0-shot and the result stays the same.} }
    
    % Log /checkpoint/xianl/few_shot/multilingual_eval/openai_curie_storycloze_val2016_split_20_80_train_val2016_split_20_80_eval_lang/task.storycloze_tmp.storycloze_train.val2016_split_20_80_train.en_val.None.en_eval.val2016_split_20_80_eval.en_calib.None_fs0_results.json.}
    
    %[Daniel] /checkpoint/danielsimig/multilingual/en_6.7B/storycloze_6.7B_gpt3_setting_fs.0_t5_smpl.All_current/task.storycloze_tmp.storycloze_train.val2016_split_20_80_train.en_val.None.en_eval.val2016_split_20_80_eval.en_calib.None_fs0_results.json
    % max_tokens seems to be 2048 in both cases.
    
     \label{tab:comp_xstorycloze}
\end{table*}

% \footnotetext{}

% \subsection{Zero-shot and In-language Few-shot Learning}
% \subsection{Comparison to State-of-the-Art Baselines}
\subsection{Performance on Multi-lingual Tasks}
% \subsection{Comparison to English Centric Models}
\label{sec:zero_shot_and_in_language_few_shot_learning}
%%% Outline
% \comment{Report final results on test set}
% We report our zero-shot results in Table \ref{tab:zeroshot}, and visualize how the different model sizes and languages in Figure 
% \ref{fig:zeroshot}. \comment{Victoria to fill in}
Using English as the universal prompting language, we characterize the zero- and few-shot in-context learning capabilities of \model{\Oursns}{7.5B} on XNLI, XCOPA and XStoryCloze and compare them to English centric language models of comparable size. %We also benchmark its few-shot learning performance on the English tasks. 

\paragraph{Comparison to GPT-3.}
% Done. \vic{Shall we move it to after ``Performance by resource level''? We also compared to GPT-3 on the En tasks.}
% \vic{For Table 5, 6, 7, turn into condensed bar/line charts and move full tables to appendix?}
We compare % our multilingual 
\model{\Oursns}{7.5B} to % the state-of-the-art GPT-3 model by
\model{GPT-3}{6.7B} on high, medium, low and extremely low resources languages\footnote{We use GPT-3 Curie: \url{https://blog.eleuther.ai/gpt3-model-sizes/}}. %: XNLI, XStoryCloze and XCOPA. 
The results are summarized in Table \ref{tab:comp_xnli} and \ref{tab:comp_xstorycloze}. 
% Across all tasks, we observe that the multilingual model outperforms the English-centric model on medium, low and extremely low resource languages. 
On all three tasks, \model{\Oursns}{7.5B} outperforms \model{GPT-3}{6.7B} by a large margin according to the average performance across languages, especially on medium, low and extremely low resource languages.  % Especially, it outperforms \model{GPT-3}{6.7B} on medium, low and extremely low resource languages, indicating that % more representative language models can be trained through balancing the pre-training language distribution can lead to more representative language models. 
On XNLI, \model{GPT-3}{6.7B} performs well on English and similar languages, surpassing \model{\Oursns}{7.5B} on \lang{en}, \lang{de} (4-shot), \lang{es} (4-shot), \lang{fr} (0-shot). A possible explanation is that these languages have significant presence in the GPT-3 training data (\lang{fr}: 1.8\%, \lang{de}: 1.5\%, \lang{es}: 0.8\% as shown in Figure~\ref{fig:pretraining_tokens}) and can benefit more from the % similarity 
lexical cognates from English. % \mona{can we quote some numbers or percentages for latin based languages. ALso there is the cognates (more shared vocab) for latin based languages}
\paragraph{Comparison to Translate-test Baseline.}
% Following the previous section, we evaluate the translation based approach on XNLI, XStoryCloze and XCOPA, the key results are summarized 
We also create a translate-test baseline, where we translate the non-English examples of the multilingual tasks to English using the Google Cloud Translation API\footnote{\url{https://cloud.google.com/translate}} and use \xlmgen, an in-house replication of \model{GPT-3}{6.7B}, to perform inference. We found the translate-test is a strong baseline of multilingual zero- and few-shot learning as is shown in Table \ref{tab:comp_xnli} and \ref{tab:comp_xstorycloze}. 
% Overall, we find that first translating the tasks to English consistently \footnote{With one exception: XCOPA Thai} outperforms any current approach that runs inference on multilingual data directly - let that be GPT-3 or \Ours. 
%Our translate-test baseline performs remarkably well in the zero- and few-shot learning settings. 
Across all three tasks, it significantly narrows the performance gap between English and other languages, especially on XNLI% by first converting the input to English and then using an English-centric language model for prediction. 
%On XNLI, it brings the average performance across languages close to the English performance. % A more comprehensive set of results can be found in Table \ref{tab:translate_test_full} in the Appendix.
\footnote{The performance of translate-test baselines might be inflated given MT systems are often trained on backtranslations which makes it good at translating translationese \citep{edunov2019evaluation}, which commonly exist in non-English evaluation data. Besides, the translation-test approach relies on high-quality machine translation (MT) systems trained on large amounts of parallel data.}.
%Despite its strong performance, a translate-test approach is not always erred. When the input text is long, executing the translation API may cause significant overhead. The comparison is also % not completely fair to \model{\Ours}{7.5B} and other self-supervised language models, 
%set on unequal ground. % First, 
% A commercial translation API is often developed using a large amount of parallel data and manual heuristics while the language models are self-supervised and language-agnostic. % Moreover, a commercial translation API may use separate models for different language pairs, while the multilingual language models distribute capacity across languages. 
% Future work can potentially integrate parallel data and language-specific modules into the pre-training and fine-tuning of multilingual language models. % to improve its performance. 
% In addition, the non-English data of all three evaluation tasks is professionally translated from English, which can make translating it back easier. % \footnote{In future work, it would be interesting to test the performance of the translate-test baseline on natively constructed non-English data.}

\subsection{Performance on English Tasks}
\label{subsec:en_tasks}
% We also benchmark the performance of \model{\Oursns}{7.5B} on English tasks (Table~\ref{tab:en_tasks}). While \model{\Oursns}{7.5B} demonstrated competitive performance on all tasks benchmarked, its performance is significantly lower comparing to \model{GPT-3}{6.7B}. % , likely as a result of the curse of mulitlinguality~\cite{Lample2019Cross-lingualPretraining}.
% Appendix~\ref{sec:en_tasks_performance} shows a detailed comparison between \model{\Oursns}{7.5B} and \model{GPT-3}{6.7B} on the English tasks.

\begin{figure*}[ht!]
    \centering
    \includegraphics[width=\linewidth]{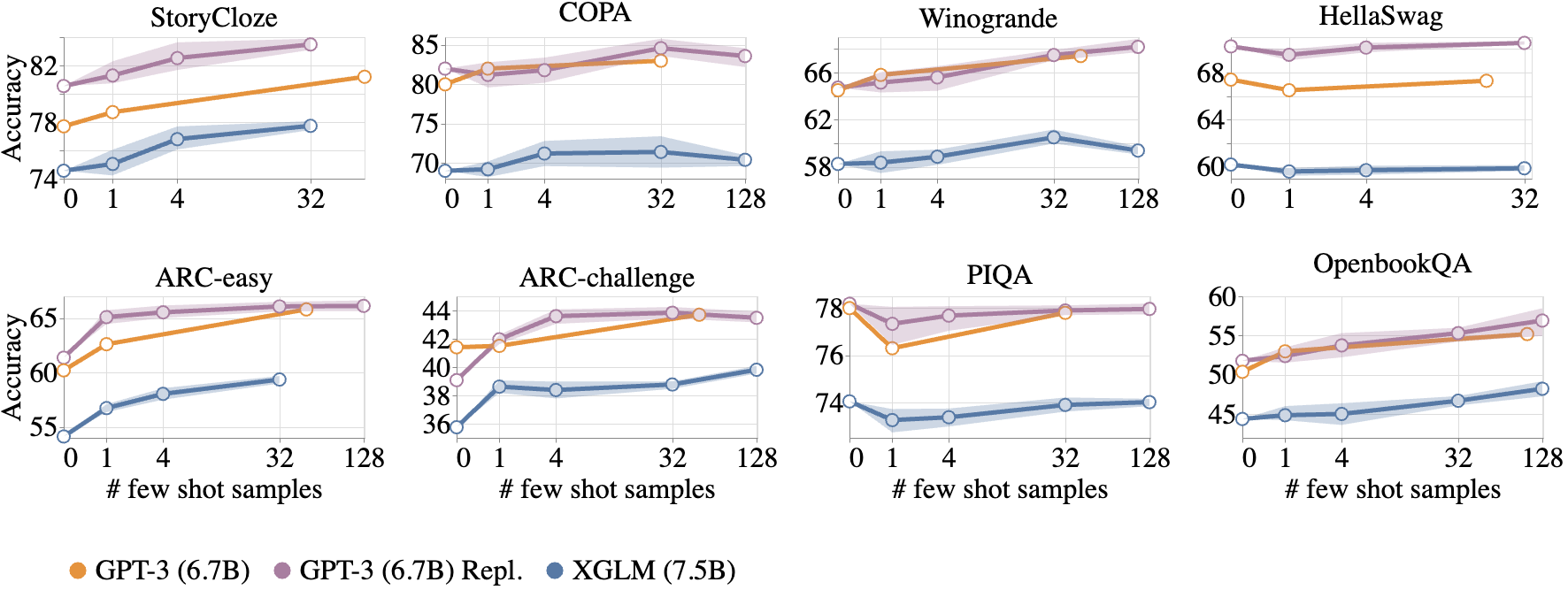}
    \caption{Performance on English tasks. For \model{\Oursns}{7.5B} and \xlmgen, we plot the confidence interval from 5 different runs corresponding to different training sets when $k>0$. For \model{GPT-3}{6.7B} we use the performance reported by~\citet{brown2020gpt3}.}% Performance of  with English dominant language models trained with 2x pretraining data on English tasks.} % \vic{Place GPT-3 results in the right shots. Add missing GPT-3 results for the rest of the tasks.}}
    \label{fig:en_few_shot_with_error_band}
\end{figure*}

We also benchmark the performance of \model{\Oursns}{7.5B} on English tasks. Figure \ref{fig:en_few_shot_with_error_band} shows the comparison between \model{\Oursns}{7.5B}, \model{GPT-3}{6.7B} and \xlmgen on a subset of English tasks used by~\citet{brown2020gpt3}. Our replication of \model{GPT-3}{6.7B}, \xlmgen, performs better than or close to \model{GPT-3}{6.7B} on all tasks. % for all $k$'s. 
While \model{\Oursns}{7.5B} performs competitively on all tasks, there remains a considerable performance gap comparing to \model{GPT-3}{6.7B} and \xlmgen. 
% \ves{I don't understand this next paragraph. Are we saying the improvements going from k to 2k is the same?}
On most tasks \model{\Oursns}{7.5B} and \xlmgen % increase and decrease by about the same magnitude 
show similar performance trend as $k$ changes. For example, both models show a performance dip at 1-shot on HellaSwag and PIQA, and 128-shot on COPA.
% being a multilingual language model does not significantly impact the model's in-context learning ability in English. In Figure~\ref{fig:en_few_shot_with_error_band}, except for Winogrande (128-shot) and ARC-challenge (32-shot), we observe consistent performance changes between \model{\Oursns}{7.5B} and the English centric models as $k$ changes across tasks. %  shows very few cases where conditioning on demonstration examples are effective for the English centric models but not \model{\Oursns}{7.5B}.

% Both English-centric models were pretrained on 300B tokens with 99\% English, while multilingual model was trained with 500B tokens with 150B English.  
There are multiple reasons why \model{\Oursns}{7.5B} underperforms English centric models on the English tasks. First, only 32.6\% of \model{\Oursns}{7.5B}'s 500B-token training data is English while both English-centric models are trained on close to 300B English tokens. 
Second, the model capacity of \model{\Oursns}{7.5B} is shared by 30 languages, and the ``curse of multilinguality'' can degrade the performance across all languages~\cite{conneau-etal-2020-unsupervised}. % Additional experiments which control these factors would shed more lights on the impact of each factor on the observed gap. 
% \ves{It seems that the first and third reasons are somewhat repetitive unless we are saying that if we train on multilingual data three times longer that will result in the same English performance}\vic{We can also have bilingual language model trained on 150B En data and 150B Fr data, in that case the English performance maybe better than \Ours. This is the difference between the third and first reasons.}
Further scaling up the model capacity and training data can potentially close this gap. %, which we leave as  future work. % Training with larger amount of tokens, higher quality English data and larger model capacity to mitigate the ``curse of multilinguality'' are all 
\footnote{The differences between the training corpora of the three models may have also contributed to the performance difference. While both English centric models incorporate high-quality English monolingual corpora such as BookCorpus~\cite{zhu2015bookcorpus} in their training data (\model{GPT-3}{6.7B} also upsamples such high-quality data), \model{\Oursns}{7.5B} is trained solely on data extracted from Common Crawl. % The data quality and domain shift may cause \model{\Oursns}{7.5B} to perform worse on the English tasks.
However, we do not expect this to be the main impact factor. \citet{scao2022what} conducted a similar experiment showing that a multilingual model (1.3B parameters) pre-trained over 13 languages also significantly underperforms an English model trained from the same data source in terms of zero-shot generalization.} % \footnote{We provide quantitative evidence of the differences in data in Appendix~\ref{sec:quantitative_data_comparison}.} 

% Does multilingual pretraining outperform En-only pretraining? \comment{Explain differences in training data, i.e. one third of En model data.}

\subsection{Performance on Machine Translation}
\label{sec:translation}

We report machine translation results on popular WMT pairs in Table \ref{tab:wmt}, and a subset of FLORES-101 \citep{goyal2021flores} in Table \ref{tab:flores101}. We use greedy decoding for both GPT-3 and our own model, and use the same 32 examples for few-shot learning in each case.

% Table \ref{tab:flores101} reports machine translation results on FLORES-101 \citep{goyal2021flores} for all combinations of 14 representative languages.
\begin{table*}[ht!]
\centering
\begin{small}
\addtolength{\tabcolsep}{-2.5pt}
\scalebox{0.9}{
\begin{tabular}{rrlcclcclcccccclcc}
\toprule
&&& \multicolumn{2}{c}{WMT-14} && \multicolumn{2}{c}{WMT-16} && \multicolumn{6}{c}{WMT-19} && \multicolumn{2}{c}{Avg.} \\
\cmidrule{4-5} \cmidrule{7-8} \cmidrule{10-15} \cmidrule{17-18}
&&& fr-en & en-fr && de-en & en-de && fi-en & en-fi & ru-en & en-ru & zh-en & en-zh && xx-en & en-xx \\
\midrule
\multirow{3}{*}{\shortstack{GPT-3 \\ (API)}}
& Ada && 22.4 & 13.0 && 19.9 & 10.3 && 4.5 & 2.7 & 8.9 & 1.0 & 4.5 & 3.5 && 12.0 & 6.1 \\
& Babbage && 29.8 & 22.4 && 30.5 & 16.9 && 12.3 & 5.4 & 20.8 & 4.1 & 12.3 & 9.1 && 21.1 & 11.6 \\
& Curie && \textbf{35.3} & \textbf{28.7} && \textbf{36.1} & \textbf{23.7} && 18.4 & 9.9 & 28.6 & 9.8 & \textbf{17.6} & \textbf{17.4} && \textbf{27.2} & 17.9 \\
\midrule
\multicolumn{2}{c}{\model{\Oursns}{7.5B}} && 33.2 & 28.5 && 34.6 & 23.5 && \textbf{20.2} & \textbf{15.5} & \textbf{29.3} & \textbf{18.7} & 16.7 & \textbf{17.4} && 26.8 & \textbf{20.7} \\
\bottomrule
\end{tabular}
}
\end{small}
\caption{Machine translation results on WMT (detokenized BLEU). We use 32 examples from the previous edition for few-shot learning. BLEU scores computed using SacreBLEU with default settings \citep{post-2018-call}.}
\label{tab:wmt}
\end{table*}

\begin{table*}[ht!]
\centering
\begin{small}
\addtolength{\tabcolsep}{-2.5pt}
\scalebox{0.9}{
\begin{tabular}{lrcccccccccccccc|c}
\toprule
&& en & de & fr & ca & fi & ru & bg & zh & ko & ar & sw & hi & my & ta & Avg. \\
\midrule
\multirow{3}{*}{avg out of xx}
& Supervised & 24.0 & 21.0 & 20.4 & 19.1 & 17.5 & 18.6 & 20.2 & 15.5 & 14.9 & 16.1 & 16.6 & 16.2 & 7.2 & 4.8 & 16.6 \\
& \model{GPT-3}{6.7B} & 9.9 & 9.1 & 9.4 & 9.3 & 6.4 & 7.0 & 5.5 & 4.9 & 2.4 & 2.9 & 1.7 & 0.5 & 0.2 & 0.3 & 5.0 \\
& \model{\Oursns}{7.5B} & 21.1 & 16.5 & 17.1 & 13.6 & 13.4 & 13.2 & 13.9 & 9.1 & 6.5 & 10.4 & 12.1 & 9.8 & 6.9 & 7.1 & 12.2 \\
\midrule
\multirow{3}{*}{avg into xx}
& Supervised & 26.0 & 20.2 & 26.7 & 20.0 & 16.7 & 18.5 & 24.5 & 14.1 & 13.5 & 11.8 & 16.3 & 19.3 & 2.1 & 2.5 & 16.6 \\
& \model{GPT-3}{6.7B} & 18.9 & 9.9 & 14.2 & 9.3 & 4.2 & 4.8 & 2.7 & 4.0 & 0.6 & 0.5 & 0.2 & 0.3 & 0.1 & 0.1 & 5.0 \\
& \model{\Oursns}{7.5B} & 28.5 & {14.9} & {20.6} & {14.4} & {10.9} & {12.4} & {18.5} & {10.9} & {5.9} & {6.1} & {8.5} & {9.7} & {{5.8}} & {{3.5}} & {12.2} \\
\bottomrule
\end{tabular}
}
\end{small}
\caption{Machine translation results on FLORES-101 devtest (spBLEU). \model{GPT-3}{6.7B} and \model{\Oursns}{7.5B} use 32 examples from the dev set for few-shot learning. Supervised results correspond to the M2M-124 615M model from \citet{goyal2021flores}. spBLEU computed using the % official 
implementation from \citet{goyal2021flores}. Full results in Appendix \ref{app:mt}. % \vic{Keep a graph and move full table to appendix.}
}
\label{tab:flores101}
\end{table*}

GPT-3 yields strong results on a few languages that are best represented in its training data, narrowly surpassing our model on WMT French-English, German-English, and Chinese-English, as well as a few pairs the  FLORES-101 set. GPT-3 is particularly strong when English is the target language, presumably due to its strong English language modeling capability. However, it does poorly on the broader set of less-resourced languages. For instance, GPT-3 fails completely when translating into Korean, Arabic, Swahili, Hindi, Burmese and Tamil in FLORES-101, with a spBLEU score of 1.2 in the best case.

% Our model obtains solid results across the board, surpassing GPT-3 in 171 out of 182 language pairs on FLORES-101. GPT-3 does well on a few languages that are best represented in its training data, narrowly surpassing our model on the remaining 11 language pairs on FLORES-101. GPT-3 is particularly strong when English is the target language, presumably due to its strong English language modeling capability. However, it does poorly on the broader set of less-resourced languages. 

% We report machine translation results on popular WMT pairs in Table \ref{tab:wmt}, and a subset of FLORES-101 \citep{goyal2021flores} in Table \ref{tab:flores101}. We use greedy decoding for both GPT-3 and our own model, and use the same 32 examples for few-shot learning in each case.

% GPT-3 obtains strong results on a few languages that are best represented in its training data, narrowly surpassing our model on WMT French-English, German-English, and Chinese-English. GPT-3 is particularly strong when English is the target language, presumably due to its strong English language modeling capability. However, it does poorly on the broader set of less-resourced languages, in particular when English is the source language. For instance, GPT-3 fails completely when translating into Korean, Arabic, Swahili, Hindi, Burmese and Tamil in FLORES-101, with a spBLEU score of 1.2 in the best case.

In contrast, our model obtains solid results across the board. In addition to surpassing GPT-3 in 171 out of 182 language pairs in the FLORES-101 set, our model is also competitive with the official supervised baseline for this dataset, even surpassing it in 45 language pairs. This suggests that large-scale multilingual language models have a great potential for building machine translation systems for low-resource languages, even if little or no parallel data is available. % \mona{This is a really cool result that warrants more highlighting. Maybe compare to the current multilingual model that was used by NLLB which relies on parallel corpora and show the competitiveness as this can open the door for NOT relying strictly on parallel corpora.}

\subsection{Scaling up Model Size}
\label{sec:model_scaling}

\begin{figure}[t!]
    \centering
    \includegraphics[width=.95\linewidth]{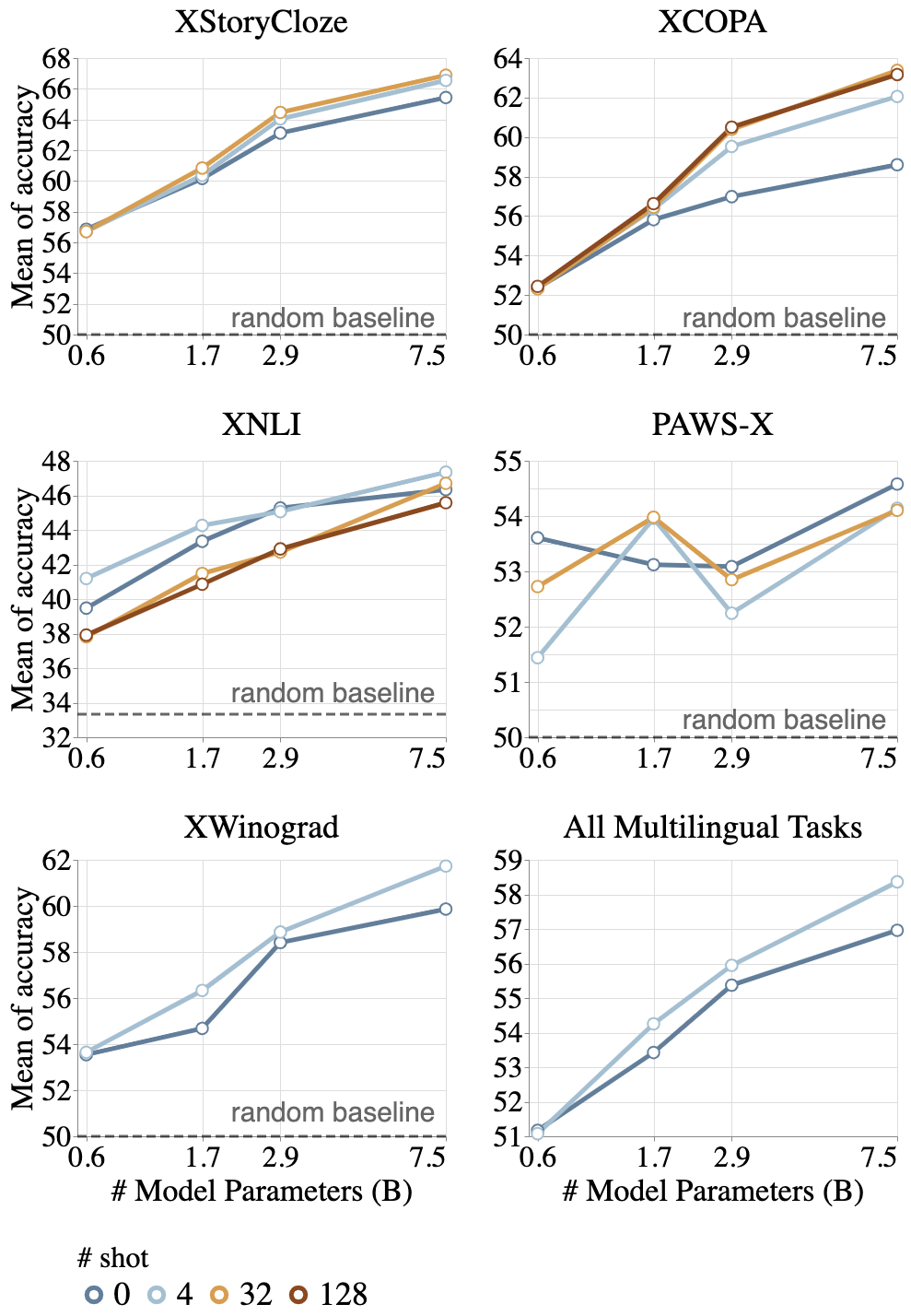}
    \caption{Zero-shot and in-language few-shot learning performance as a function of model size. The last plot shows the average performance over all five tasks in 0- and 4-shot learning.}
    \label{fig:multi_few_shot_scaling}
\end{figure}
% \paragraph{Few-shot learning on downstream tasks.} We found that improvement in language modeling perplexity as scaling model sizes also correspond to improved performance in few-shot learning on downstream tasks. As is shown in Figure \ref{fig:multi_few_shot_scaling}, in-context learning with 0, 4, 32 and 128 examples improved accuracy on all five multilingual tasks. 
Finally, we study the impact of scaling up the model parameter size on its 0- and few-shot learning capabilities. 
Figure \ref{fig:multi_few_shot_scaling} shows the % in-context learning 
performance ($k=0, 4, 32, 128$) of the four \Ours models (564M, 1.7B, 2.9B, 7.5B) on the five multilingual tasks. The $y$-axis represents the average accuracy across languages for each task. 
On commonsense reasoning tasks (XStoryCloze, XCOPA, XWinograd), the performance of all models % monotonically 
increases as $k$ increases from 0 to 32. % (on XCOPA, going from 32- to 128-shot results in small performance drop). 
The performance gain from % conditioning on 
demonstration examples % ($k>0$) 
also gets larger as the model size increases, indicating bigger models can better leverage the in-context examples. On XNLI, the performance of all models increases as $k$ increases from 0 to 4, but decreases for $k$ at 32 and above. %, indicating that larger $k$'s do not always yield better results for some tasks\footnote{\citet{brown2020gpt3} has reported the same observation.}. 
% When conditioning on 
With the same number of demonstration examples, larger models do not always benefit more.
PAWS-X is a task where in-context learning struggles -- the performance of all models oscillates near random (50\%) as $k$ changes. A possible reason is the adversarial nature of PAWS-X, where the paraphrase and non-paraphrase pairs by design have high lexical overlap.
%%%%%%%%%%%%%% EMNLP submission %%%%%%%%%%%
\hide{
In all tasks except for PAWS-X\footnote{PAWS-X is adversarially created. It contains non-paraphrase sentence pairs with high lexical overlap, which has been shown challenging to state-of-the-art paraphrase models~\cite{}}, % 4 out of 5 tasks (XStoryCloze, XCOPA, XNLI, XWinograd), 
the performance % monotonically 
improves as the model scales for all $k$'s, indicating larger models can better leverage the in-context examples. 
}
We expect scaling to be an effective recipe for building stronger multilingual language models, given the current trend. %  We further examine each model's behavior with respect to different $k$'s. 

\section{Related Work} \label{section:related_work}

\paragraph{Language model prompting.} \citet{brown2020gpt3} first demonstrated in-context few-shot learning using the GPT-3 model. This method %, commonly referred to as "in-context" learning, 
removes the need for task-specific updates to the model parameters: the few-shot examples that one would normally use for fine-tuning are provided at inference time to the same model for each task. 
% In-context learning capabilities of language models only emerge at scale. %, when a model has a sufficiently large number of parameters. 
% \citet{Wei2021FINETUNEDLEARNERS} shows that such pre-trained language models can also be further fine-tuned to work with prompts better in general. 
On several high-resource Latin language pairs, GPT-3 achieves machine translation performance that is close to or better than state-of-the-art supervised models, given only a handful of demonstration examples.\footnote{Study shows that language contamination in pre-training data can effectively boost the cross-lingual capability of English-centric language models~\cite{DBLP:journals/corr/abs-2204-08110}. With a heavier tail of deliberately introduced multilingual data, PALM-540B~\cite{DBLP:journals/corr/abs-2204-02311} later achieves even stronger few-shot machine translation performance.}
% Similarly to previous changes in paradigms, in-context learning also 
Such change in the learning paradigm raises new questions about multilinguality, which % aspect of in-context learning 
has not been studied as extensively. % as it had been the case for previous families of multilingual language models at smaller scales. % models described in Section \ref{subsection:multilingual_models}. 
\citet{DBLP:journals/corr/abs-2109-07684} evaluates the in-context few-shot learning abilities of several GPT-2, \model{GPT}{NEO} and T5 on three additional languages (\lang{de}, \lang{es}, \lang{fr}) using multiple NLU tasks, considering monolingual prompts as well as cross-lingual prompts, demonstrating the multilingual in-context learning skills of the English GPT and T5 models. ~\citet{DBLP:journals/corr/abs-2109-03630} evaluated different fine-tuning and prompt-tuning~\cite{DBLP:journals/corr/abs-2103-10385} approaches on XLM-R and demonstrates the effectiveness of prompting in few-shot crosslingual transfer and in-language training of a multilingual masked language model. 
% We demonstrate the competitive cross-lingual zero-shot and in-context few-shot learning capabilities of multilingual language models trained with language rebalancing and at a much larger scale.

\paragraph{Multilingual pre-training.} \label{subsection:multilingual_models}
Early % pre-trained word representations were replicated on other languages \cite{Mikolov2013ExploitingTranslation} simply by replacing the English corpus used for training to corpora from 
multilingual pre-training work train word embeddings over multilingual corpora~\cite{Mikolov2013ExploitingTranslation}. The multilingual versions of contextualized embedding models such as BERT~\cite{devlin2018bert}, RoBERTa~\cite{liu2019roberta}, BART~\cite{Lewis2019BART:Comprehension} and T5~\cite{raffel2020exploring} were also developed: mBERT~\cite{devlin2018bert}, XLM-R \cite{conneau-etal-2020-unsupervised}, mBART \cite{liu2020multilingual}, and mT5~\cite{Xue2020MT5:Transformer}.  % Unlike earlier models, these
Such models were trained on a single, multilingual text corpus such as mC4 \cite{Xue2020MT5:Transformer} or CC25 \cite{liu2020multilingual}. 

% This brought the benefit of having a single model able to work with various languages, but also meant that 
% To this end, 
Several approaches have been developed to facilitate cross-lingual transfer, including %, a number of new considerations had to be taken into account. % As the set of words a model had to recognize increased, 
sub-word tokenizers which enabled efficient, shared vocabulary learning across languages \cite{Kudo2018SentencePiece:Processing}, joint training for efficient knowledge transfer across languages \cite{Pires2019HowBERT,Jiang2020X-FACTR:Models,DBLP:conf/eacl/KassnerDS21}, etc. A notable concurrent work is BLOOM~\footnote{\url{https://bigscience.huggingface.co/blog/bloom}}, which scales multilingual pre-training to 46 languages and 175 billion parameters.
\section{Conclusion}
We introduce four multilingual generative language models (XGLMs) at different scales, and study their in-context few- and zero-shot learning capabilities. 
% Our models are trained on a large-scale multilingual corpus covering 30 diverse languages. % with a relatively balanced language distribution. 
We show that the few-shot learning capability of XGLM steadily improves as it scales. Our largest model (7.5B parameters) sets a new state of the art for few-shot learning in more than 20 languages (including mid- and low-resource languages) on commonsense reasoning, NLI and machine translation tasks. An in-depth analysis shows the models are highly cross-lingual, which leads to strong few-shot learning performance in non-English languages. % We present detailed analysis and discussion of the model strengths and limitations. We also conduct a preliminary empirical investigation of XGLM models in social value tasks such as hate speech detection and gender bias. Similar to comparable sized GPT-3 models, XGLM exhibits performance challenges on such tasks.

% \section*{Acknowledgements}
% We thank Halil Akin for his contribution setting up the initial multilingual data processing pipeline, and Shiyue Zhang and Jonas Pfeiffer for their input on multilingual vocabulary construction.
\section*{Limitations}
% \comment{TODO: condense the message and shorten the section} \xian{I found the GPT-3 paper has a very comprehensive section for limitations, which is not a bad thing if we do not have page limit. }

Although the multilingual language model is an important step towards building inclusive general-purpose foundation models, our current models have the following limitations.

% training data
\paragraph{Training Data.} Our models are trained on a static multilingual corpus extracted from CommonCrawl, with English text comprising 32.6\% of the total number of tokens corresponding to 163B tokens. The English data portion of the corpus corresponds to roughly 54\% only of GPT-3's training data. We applied several data filtering strategies as proxies for data quality assurance (see a comprehensive list in the Data Card in Appendix \ref{sec:appendix_datacard}), such as removing duplicated documents and paragraphs by URLs, filtering out paragraphs with high ratio of digits and punctuation, removing paragraphs with profanity, filtering by max number of URLs and minimum length, etc. Such filtering may potentially result in bias of the remaining data used in pretraining, which would need further analysis to understand. Furthermore, the raw data were taken from static CommonCrawl snapshots, which may not include entities and events beyond the time span of the snapshots (till March 2020), such as COVID-19, etc. As such we also note the potential difference in genres between CommonCrawl and the genres used in GPT-3 comprising in addition to CommonCrawl, corpora such as BookCorpus and Wikipedia. 

Moreover, GPT-3  is trained on 118 languages %,  to conduct controlled experiments seen and unseen languages. In contrast, GPT-3 models' training data contains 118 languages 
despite the fact that 93\% of the data is  English.\footnote{\url{https://github.com/openai/gpt-3/blob/master/dataset_statistics/languages_by_word_count.csv}} In contrast our models are trained on 30 languages after rigorous language identification and filtering.

\paragraph{Performance on English tasks.} As is shown in Section \ref{subsec:en_tasks} and Figure \ref{fig:en_few_shot_with_error_band}, our model underperforms English-centric models on eight tasks ranging from commonsense reasoning to QA. There are several factors which could be contributing to this gap, such as 
\begin{itemize}
    \item Difference in training data quality (XGLM is trained on filtered CommonCrawl data only, while the English-centric models are trained on data including both CommonCrawl as well as high-quality corpora such as BookCorpus and Wikipedia) and quantity (as is described in the previous paragraph, the multilingual model was trained on 54\% of the English data used in English-centric models);
    \item Curse of multilinguality. Previous work in multilingual training has shown that increasing the number of languages in model with shared parameters hurts performance on all training languages, e.g. English \cite{conneau-etal-2020-unsupervised}.
\end{itemize}
Additional experiments  controlling for these factors would shed more light on the observed gap. 

\paragraph{Model architecture and training objective.} In this work, we only experimented with  causal language models with a decoder-only architecture, which had previously demonstrated promising few-shot learning capabilities \cite{brown2020gpt3}. However, such architecture and pretraining objective do not leverage bidirectional context such as those used by masked language models (MLM), or sequence-to-sequence architectures with denoising autoencoder pretraining objectives. % In this study, we conduct a preliminary comparison with a multilingual MLM (XLM-R) exploring few-shot performance in Section \ref{sec:compare-to-small-lm}. A more comprehensive study would warrant a comparison with other architectures such as mT5, mBART \cite{liu2020multilingual}.% is left for future work.  
% performance on RAI tasks and safety implications

\paragraph{Model evaluation via in-context learning.} We compare our language models to the baselines primarily in the in-context learning paradigm, using the same prompts for all language models in the comparison unless explicitly specified. % However, language models is sensitive to the prompt construct. 
Despite minimal effort engineering the prompts for any model, it is possible that the prompts work better with some models than the others, which introduces bias to the evaluation. However, we expect this factor to have small impact and the relative strengths of the models can be reliably measured given the volume of tasks they were evaluated on.

\paragraph{Evaluation on social value tasks for more languages.} %Understanding foundation language model's property for responsible AI and its societal impact is of great importance. In this work, 
We evaluate and analyze the models' performance on hate speech detection and gender bias for professional occupations. These studies are limited by the available evaluation datasets. We are limited in our study as we only investigate this problem space for six languages (English, French, Spanish, Italian, Portuguese, and Polish) where a majority of them (5) pertain to the Romance language family. It would be pertinent to investigate the impact of multilingual models on social value tasks across a wider and more diversified set of languages before drawing solid conclusions. %To fully understand multilingual language models' society impact on various cultures and regions, evaluation on more languages corresponding to linguistic and geo-location diversity would be needed Similarly
Moreover, we contend that studies on other tasks such as stereotype~\cite{nangia2020crows, nadeem2020stereoset}, ethics~\cite{hendrycks2020aligning} would provide a more comprehensive view of model behavior for social value tasks. 
% training efficiency

% training dynamics among languages
\section*{Ethical Considerations}
Devising multilingual pre-trained language models can serve as a powerful tool in the NLP arsenal for multiple reasons. 

\paragraph{Energy and maintenance efficiency.} From an engineering perspective, XGLM pertains to a family of models that represent single unified models catering to many languages which have wide application across many applications. Such a unified single model saves on carbon footprint as well as energy consumption (comparing to the alternative: separate models for different languages) leading to more energy efficiency. A single model, despite having the risk of being a single point of failure, has the powerful incentive of being easier to maintain, access, distribute, and track. 

\paragraph{Diversity and inclusion.} Models such as XGLM represent a paradigm shift from the Anglo-centric view of the world of NLP to being able to cater to all languages on an equal footing. Paying attention to the design of such models is critical to ensure equitability and inclusion, exemplified here by attempting to balance language representation. The further power of XGLM specifically is its ability to perform comparably to Anglo-centric models in zero to few shot settings. Possessing powerful multilingual models that can perform well in such settings especially for medium to extremely low resource languages helps alleviate the burden of creating supervised data for such languages especially for economically challenged languages (medium to low digital presence typically goes hand in hand with economic disparities). Moreover, having such models catering to scarcer languages spurs scientific research in such languages leading to more diversified NLP, and more diversified science in the broader sense.

\paragraph{Social values.} We further investigate the impact of our models on social valued problems such as hate speech detection and bias (Appendix~\S\ref{sec:safety_and_bias_analysis}). Despite inconclusive results overall (bordering on negative), we note that for the relatively scarcer data setting (Polish) the multilingual models outperform the Anglo-centric models indicating that XGLM will be performant for less resourced languages. This is especially significant for social value tasks where obtaining training data is quite problematic due to the inherent expense of obtaining high quality annotated data.

\paragraph{Transparency and Accountability.} 
In the spirit of transparency and accountability for large-scale language modeling we include detailed model card and data card with the model and paper release. %\xian{Do we have model card? cc Punit}  

\bibliography{anthology,custom}
\bibliographystyle{acl_natbib}

\clearpage

\appendix
\onecolumn

\setcounter{table}{0}
\renewcommand{\thetable}{A\arabic{table}}
\setcounter{figure}{0}
\renewcommand{\thefigure}{A\arabic{figure}}

% \section{Languages}
% \label{sec:appendix_lang}
% \vic{Report language information in the data card section}
\section{Pretraining Details}
\label{sec:language_model_pre_training_details}
\paragraph{\Ours.} All models are trained with the Fairseq library \cite{ott2019fairseq}. We use Adam optimizer with $\beta_1=0.9$, $\beta_2=0.98$, $\epsilon=1e-8$. We adjust the learning rate based on model size, e.g. $1.5e-3$ for the 564M and 1.7B model, $7.5e-4$ for the 2.9B model, and $1.2e-4$ for the 7.5B models. Learning rates were adjusted with a 2000 warm-up updates followed by a polynomial decay schedule. All models are trained with data parallel and an effective batch size of 4M tokens. The XGLM 7.5B model was trained on 256 A100 GPUs for about 3 weeks, at a speed of 311.6k words per second\footnote{On 256 A100 GPUs, the inference speed can reach 1.47 million words per second. Besides, inference can be done with significantly less resources. For example, using 8 v100 GPUs, it took ~6 hrs to evaluate XGLM 7.5B on XStoryCloze.}. 

\paragraph{\xlmgen.} We replicate the \model{GPT-3}{6.7B} architecture and optimization hyperparameters to the best of our knowledge for training this model. % when training the \Oursen model. 
The most significant difference between this model and \model{GPT-3}{6.7B} is in the training data. The training data used by \xlmgen is a combination of six English-language datasets, % including the five datasets used to pretrain RoBERTa~\citep{liu2019roberta} and the English subset of CC100, 
totaling 453GB and 112B tokens (which we up-sampled to 300B tokens):
\begin{itemize}[leftmargin=*]
\item \textit{BookCorpus}~\citep{zhu2015bookcorpus}, a dataset consisting of more than 10K unpublished books (4GB);
\item \textit{English Wikipedia}, excluding lists, tables and headers (12GB);
\item \textit{CC-News}~\citep{nagel2016ccnews}, a dataset containing 63 millions English news articles crawled between September 2016 and February 2019 (76GB);
\item \textit{OpenWebText}~\citep{gokaslan2019openwebtext}, an open source recreation of the WebText dataset used to train GPT-2 (38GB);
\item \textit{CC-Stories}~\citep{trinh2018simple}, a dataset containing a subset of CommonCrawl data filtered to match the story-like style of Winograd schemas (31GB);
\item \textit{English CC100}~\citep{wenzek-etal-2020-ccnet}, a dataset extracted from CommonCrawl snapshots between January 2018 and December 2018, filtered to match the style of Wikipedia (292GB).
\end{itemize}
The data are encoded using the same Byte-Pair Encoding (BPE) as GPT-2~\citep{radford2019language} and RoBERTa~\citep{liu2019roberta} with a vocabulary of 50K subword units.

\subsection{Validation Perplexity} % We evaluate models of different sizes on multilingual language modeling. 
\begin{figure}[ht]
    \centering
    \includegraphics[width=0.36\linewidth]{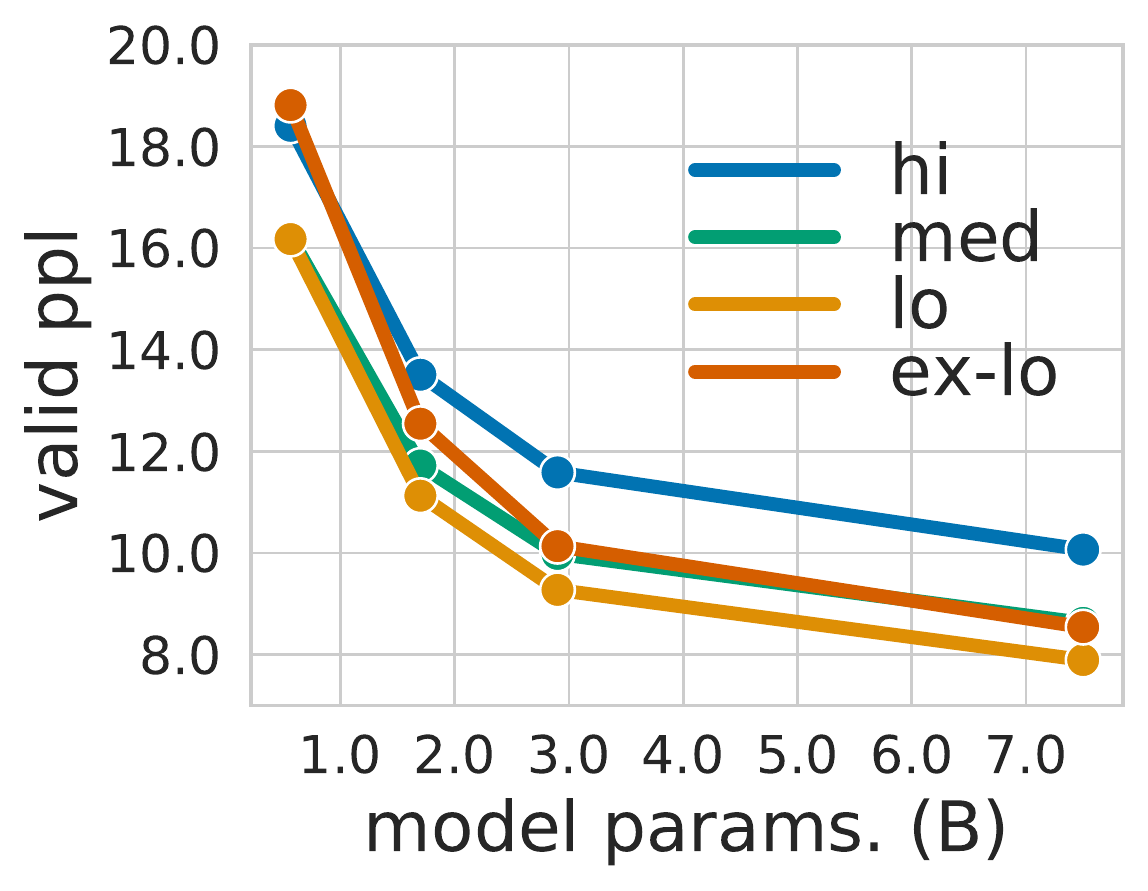}
    \caption{\Ours perplexity on CC100\_XL validation set as a function of model size.}
    \label{fig:multi_ppl_scaling}
\end{figure}

We use in-domain validation perplexity to validate the convergence status of the models. Figure \ref{fig:multi_ppl_scaling} shows the average perplexity of the four models evaluated using a validation dataset sampled from CC100-XL. 
The validation data contains 30k sentences for each language that do not overlap with the pre-training data.
% We can see the 
We group the results by resource level.
%Scaling up model size consistently improves the language modeling performance for % languages in 
%all resource levels.

\section{Multilingual In-context Learning Formulation}
\label{sec:task_formulation}

We extend the in-context learning framework proposed by~\citet{brown2020gpt3} to the multilingual setting.
Let $\mM$ be a causal % multilingual 
language model and $\mD$ be a task. $\mD=(\mP, \mE)$ consists of a task description $\mP$ and a few demonstration examples in one or more languages $$\mE=\bigcup_{l=1}^{|\mL|}\mE^l.$$ % \mE^l=\{(x^l_i, y_i)\}^{k^l}_{i=1}.$$ $\mE_l=\{(x^l_i, y^l_i)\}^k_{i=1}$. %, and a test set $\mT=\{(\Tilde{x}_j, \Tilde{y}_j\}^N_{j=1}$. 
We consider the setting where the task description comes in the form of a prompt $\mP=(\mT, v)$. $\mT$ is a cloze-style template that converts an example input $x$ into a string $\mT(x)$ that contains a \mask symbol.\footnote{We relaxed the prompt format of GPT-3 by allowing the \mask symbol to appear anywhere in $\mT(x)$ instead of just in the end. Having this additional flexibility leads to better performance on some tasks. This is inspired by the masked language modeling prompts constructed by recent work~\cite{DBLP:conf/eacl/SchickS21,DBLP:journals/corr/abs-2108-13161}.} For classification and multiple-choice problems, $v: \mY\rightarrow \mV^*$ is a verbalizer that maps each candidate label or choice $y\in\mY$ into a string $v(y)$. % that contains
Both $\mT(x)$ and $v(y)$ can be tokenized into a sequence of one or more tokens in the % multilingual 
language model vocabulary $\mV$. An instantiated prompt $\mP(x, y)$ is obtained by substituting the \mask symbol in $\mT(x)$ with $v(y)$. Table~\ref{tab:all_multi_task_prompts} shows the prompts used by all tasks in our main experiments.

% This formulation covers % the prediction objectives of 
% most NLP tasks, % (classification, multiple choices and generation), 
% depending on the different constructs of $v$. 

\paragraph{Zero-shot learning.} Given a test example $\xtest^t$ in any target language $t$, the zero-shot prediction is $\hat{y}$ which maximizes a language model based scoring function (\S\ref{sec:scoring_function}). % the language modeling probability: % when\mask in $\mT(\xtest^t)$ is replaced by $v(y)$:
\begin{equation}
\label{eq:zero_shot_objective}
    % \hat{y} = \argmax_y{p(\mask=v(y) | \mT(\xtest^t))}.
    \hat{y} = \argmax_y\sigma{(\mM, \mP(\xtest^t, y))}.
\end{equation}

This general formulation can cover most NLP tasks. For classification problems, $v$ is a mapping from classes to strings; for multiple-choice problems, $v$ is an identity function that maps each candidate choice to itself. For text generation problems, $v$ is identity and we decode free-form text from \texttt{[Mask]}, which in this case is positioned at the end of $\mT(x)$. 

\paragraph{Few-shot learning.} Suppose we have $k$ demonstration examples available in a source language: $$\mE^s=\{(x_i^s, y_i)\}^k_{i=1}.$$ 
In this case, we % make the model prediction condition on both % the task description $\mP$ and $\mE^s$. We 
concatenate the instantiated prompts of the demonstration examples $\{\mP(x^s_i, y_i)\}^k_{i=1}$ and make it the prefix of the input string used in the zero-shot learning setting to form the objective:
\begin{equation}
% \begin{multline}
\label{eq:few_shot_objective}
    \hat{y} = \argmax_y\sigma(\mP(x^s_1, y_1)\ \sep\hdots\ \mP(x^s_k, y_k)\ \sep\ \mP(\xtest^t, y)),
% \end{multline}
\end{equation}
where $\sep$ is a separator symbol chosen empirically.
\begin{itemize}
    \item When $s=t$, we have the \emph{in-language} few-shot learning setup. % \label{item:1}
    \item When $s\neq t$, we have the \emph{cross-lingual} few-shot learning setup. % \label{item:2}
\end{itemize}

\section{Evaluation Details}
\label{sec:evaluation_details}
\subsection{English Evaluation Tasks}
\label{sec:en_tasks}
Table~\ref{tab:en_tasks} shows all the English tasks used in our evaluation.
\begin{table*}[t!]
\centering
\resizebox{.72\textwidth}{!}{%
\begin{tabular}{l l r r r r r}
    \toprule
    % Knowledge & mLAMA~\cite{DBLP:conf/eacl/KassnerDS21} & -- & 2,000 & 2,000 & translations & 53 & Accuracy \\\midrule
    \multirow{4}{*}{\makecell[l]{Reasoning}} 
    & StoryCloze~\cite{DBLP:journals/corr/MostafazadehCHP16} & -- & 1,871 & 1,871 & N/A & 1 \\
    & COPA$^\ddagger$~\cite{DBLP:conf/semeval/GordonKR12} & 400 & 100 & 499 & N/A & 1 \\
    & WinoGrande~\cite{DBLP:conf/aaai/SakaguchiBBC20} & 40,398 & 1,267 & 1,767 & N/A & 1 \\ % \midrule
    & HellaSwag~\cite{DBLP:conf/acl/ZellersHBFC19} & 39,905 & 10,042 & 10,003 & N/A & 1 \\
    \hdashline
    \multirow{4}{*}{QA} & ARC-easy~\cite{DBLP:journals/corr/abs-1803-05457} & 2,251 & 570 & 2,376 & N/A & 1 \\ 
    & ARC-cha.~\cite{DBLP:journals/corr/abs-1803-05457} & 1,119 & 299 & 1,172 & N/A & 1 \\ 
    & PIQA~\cite{DBLP:conf/aaai/BiskZLGC20} & 16,113 & 1,838 & 3,084 & N/A & 1 \\
    & OpenbookQA~\cite{DBLP:journals/corr/abs-1809-02789} & 4,957 & 500 & 500 & N/A & 1 \\
\bottomrule
\end{tabular}
}
\caption{English tasks used in our few-shot learning evaluation. All tasks use accuracy as the evaluation metrics.}
%\footnote{We include COPA when reporting the results of XCOPA for the rest of the paper. This way all our multilingual tasks include English as an evaluation language.}
% , the $|$Train$|$, $|$Dev$|$, $|$Test$|$ statistics were sums over all languages
\label{tab:en_tasks}
\end{table*}

\subsection{Scoring Functions}
\label{sec:scoring_function}
% \comment{Todor, Mikel: fill in different scoring functions we've implemented.}
% Besides the standard language modeling objective which computes the joint probability of the input sequence, we considered 
We considered the following functions for scoring an instantiated prompt using a language model:
\begin{enumerate}
    \item sum of per-token log probabilities;   \label{item:1}
    \item average of per-token log probabilities;   \label{item:2}
    \item average of per-token log probabilities, ignoring the common prefix of different candidates.   \label{item:3}
\end{enumerate}
We also considered the calibration approach proposed by~\citet{DBLP:conf/icml/ZhaoWFK021} and character normalization proposed by~\citet{J1WhitePaper}.

% In the end, we use the average of per-token log probabilities ignoring the common prefix of different candidates as the scoring function for all multilingual tasks, without additional calibration and normalization. This is selected based on the development set performance of StoryCloze and XNLI. 

% \paragraph{Summary.} 
In the end, we use the average of per-token log-probabilities ignoring the common prefix of different candidates as the scoring function for all multilingual tasks. This is selected based on the development set performance of StoryCloze and XNLI. 
\hide{
We use English prompts for all languages % when reporting our main results (\S\ref{sec:main_results}) 
% in in-language few-shot learning 
by default, selected based on the development set performance on XNLI and XCOPA (\S\ref{sec:multilingual_prompt_construction}). 
}
% The prompts used for all of our multilingual tasks are detailed in Appendix~\ref{sec:all_task_english_prompts}.
% We use the same learning options as GPT-3~\cite{brown2020gpt3} when evaluating on 

For English tasks, we use the same modeling choices as~\citet{brown2020gpt3}. 
% The exceptions are ARC-easy and ARC-challenge, for which we use a different template (Table~\ref{tab:en_task_prompts}) and no unconditional probability normalization, as described next.
Specifically, we use the task prompts as detailed in Appendix G of~\citet{brown2020gpt3}, and a single newline as the separator for few-shot learning. For WinoGrande, we take the log-likelihood of the common suffix of the different candidates as the scoring function. % \comment{Mikel: did we inherit the suffix scoring function from GPT-3 or did we come up with it ourselves?} \mikel{We do inherit this from GPT-3 (which at the same time inherited it from some prior work)}
For ARC-easy, ARC-challenge and OpenBookQA, we normalize by the unconditional probability of each candidate by taking $\frac{p(\mathtt{completion}|\mathtt{context})}{p(\mathtt{completion}|\mathtt{answer\_context})}$, where we use the string \textit{``Answer: ''} as answer\_context. For all the other tasks, we take the average of per-token log-probabilities, ignoring the common prefix of the different candidates.

\subsection{Evaluation Protocol}
\label{sec:eval_prototcol}
% \comment{Victoria, Todor to fill in}
% Emphasize how to do true few-shot evaluation. 
% We evaluate our models using both in-domain perplexity (\S~\ref{sec:model_scaling}) and zero-shot and few-shot learning performances on the downstream tasks. 

% \paragraph{In-domain perplexity.} We report the perplexity of our language models on validation documents sampled from the same distribution as the pre-training data.
% We sample documents from the same distribution as the pre-training data, and report the average perplexities of languages in four resource tiers\footnote{We grouped the languages into four resource levels based on their token count $X$ in our pre-training dataset. High: $X\geq 48\text{B}$, Medium: $5\text{B}\leq X<48\text{B}$, Low: $0.1\text{B}\leq X<5\text{B}$, Extremely-low: $X< 0.1\text{B}$.}.
All few-shot learning results are obtained with the \emph{in-language} setting (both the training and test examples are in the same language) unless otherwise specified.
We report results on the test set for all multilingual tasks (including the held-out tasks). % described in \S~\ref{sec:tasks}. 
For English tasks, we report results on the test set for ARC-easy, ARC-challenge, OpenBookQA and StoryCloze, and on the development set for the rest, following~\citet{brown2020gpt3}. For few-shot learning, we report the average results across 5 runs, randomly sampling a different set of few-shot examples each time. For tasks with a training set, we sample the few-shot examples from the training set; for tasks with no training set, we sample from the dev set and report evaluation results on the test set; for dev-set examples on XNLI and XCOPA, we sample few-shot examples from the test set, since these two tasks do not have the training sets for all languages. While~\citet{brown2020gpt3} tuned the few-shot value $k$ as a hyperparameter on the dev set, we pre-selected a few $k$ values % commonly used in literature 
(0, 1, 4, 32, 128) and report the corresponding results. % We also consider a \emph{max} setting, where $k$ is set to be the maximun number of examples that fit the maximum context length ($n_{\text{ctx}}=2048$) of the language model. In this setting, 

\subsubsection{Example Truncation}
\label{sec:num_few_shot_examples_max_fit}
Following~\citet{brown2020gpt3}, we truncate the input such that they fit the maximum context length of \Ours ($n_{\text{ctx}}=2048$) and preserve only the complete demonstration examples after truncation. For each task, we report results up to the $k$'s corresponding to the maximum fit.\footnote{XWinograd has only a test split, and we sampled few-shot examples directly from it, following the practice used by \citet{brown2020gpt3} for evaluating GPT-3 on Winograd. As a result we only report 0-, 1- and 4-shot results for XWinograd to minimize inflating the few-shot performance by training and testing on the same examples.} 
Table~\ref{tab:task_max_fit} shows the average number of demonstration examples that fit the maximum context length of \Ours ($n_{\text{ctx}}=2048$) for each task in our experiments. % Bounded context length is a limitation of in-context learning, as most tasks allow significantly less than 100 demonstration examples. % \footnote{We leave in-context learning from unlimited context length as a future work.}. 
% In addition, comparing the example counts for different languages of the same tasks (XStoryCloze, XCOPA, XNLI and PAWS-X\footnote{For XWinograd, the examples across different languages are not translations hence their lengths are not comparable.}) 
% More interestingly, Table~\ref{tab:task_max_fit} shows that 

\paragraph{Representation Bias.}We observe that the language model tend to fit more examples in a high-resource language in context compared to those in a low-resource language.\footnote{XStoryCloze, XCOPA, XNLI and PAWS-X all contain parallel examples, which allows us to compare the maximum fit of the same set of examples across different languages.} %, and Latin languages usually fit more examples compared to non-Latin languages. 
English, as the highest resourced language (Table~\ref{tab:cc100_data_table}), always fit the most examples. % One reason that contributes to this result is 
This reflects the unequal representation of different languages in our joint multilingual BPE vocabulary (\S\ref{sec:pretraining_data}). With this vocabulary induction scheme~\cite{DBLP:journals/corr/SennrichHB15}, the underrepresented languages tend to have smaller sub-word units and higher fertility (defined as number of subwords per linguistic word), making it more challenging to learn word- and higher-level semantics for such languages. Other factors can also impact the tokenization granularity. For example, sharing sub-strings with other high resource languages can boost the granularity of a language; and some languages have smaller tokenization granularity as a result of their alphabet system (e.g. Chinese has an average sub-word length of 1.4, indicating the dominance of single-character tokens, despite being the third largest language in our pre-training data according to disk size). 

\begin{table}[t!]
    \centering
    \setlength{\tabcolsep}{4.6pt}
    \scalebox{0.74}{
    \begin{tabular}{lccccccccccccccc}
         % Task & \multicolumn{15}{c}{Avg. \# Examples} \\
         % \midrule
         \multirow{2}{*}{XStoryCloze} & en &   zh &   ru &   es &   id &   ar &   hi &   sw &   te &   eu &   my \\
         \cmidrule(lr){2-12}
         & 32.0 & 31.6 & 30.6 & 31.5 & 32.0 & 27.6 & 24.7 & 29.5 & 25.2 & 25.6 & 18.8 \\
         \multirow{2}{*}{XCOPA} & en & zh &   it &    id &    th &   vi &    tr &    et &   ta &   sw &   ht &   qu \\
         \cmidrule(lr){2-13}
         & 100.0 & 100.0 & 98.0 & 100.0 & 100.0 & 99.4 & 100.0 & 100.0 & 75.2 & 97.9 & 95.4 & 84.9 \\
         \multirow{2}{*}{XWinograd} & en &  ru &   ja  &   pt \\
         \cmidrule(lr){2-5}
         & 93.7 & 63.7 & 62.4 & 83.1 \\
         \multirow{2}{*}{XNLI} & en &   zh &   ru &   es &   de &   fr &   el &   th &   vi &   tr &   ar &   bg &   hi &   ur &   sw \\
         \cmidrule(lr){2-16}
         & 48.3 & 47.6 & 43.4 & 44.7 & 43.1 & 39.3 & 37.8 & 44.8 & 39.8 & 46.8 & 42.3 & 41.4 & 37.4 & 38.4 & 42.9 \\
         \multirow{2}{*}{PAWS-X} & en &   zh &   es &   ja &   de &   fr &   ko \\
         \cmidrule(lr){2-8}
         & 34.5 & 27.2 & 31.3 & 23.1 & 32.0 & 29.1 & 28.0 \\
         % \midrule
         & en \\
         \cmidrule(lr){2-2}
         COPA & 124.3 \\
         Winogrande & 84.6 \\
         HellaSwag & 21.4 \\
         ARC-easy & 62.9 \\
         ARC-challenge & 53.2 \\
         PIQA & 59.6 \\
         OpenbookQA & 106.4 \\
         % \bottomrule
    \end{tabular}
    }
    \caption{Average \# of few-shot examples that fit the maximum context length of \Ours ($n_{\text{ctx}}=2048$) in our few-shot evaluation benchmark. The languages are sorted according to the amount of pre-training data (high to low).}
    \label{tab:task_max_fit}
\end{table}

\section{Additional Results}
% \section{Analysis}
% \label{sec:analysis}

\subsection{Comparing Multilingual Prompting Approaches on XNLI and XCOPA}
\label{sec:comparing_multilingua_prompting_appendix}

We compare % \emph{same-language} prompting and \emph{high-resource} prompting (\S\ref{sec:multilingual_prompts}) for \model{\Oursns}{7.5B} 
the performance of English prompts and \texttt{MT} and \texttt{HT} prompts on two of our held-out tasks, XNLI and XCOPA, using their development sets.
% We compare under zero-shot and in-language few-shot learning settings. 
For % \emph{same-language} prompting, 
\texttt{MT} prompts, we translate the English prompts into the target languages using the Google Cloud Translation API. % for all verbal prompts (XCOPA, XNLI, PAWS-X). 
We use the exact prompts as shown in Table~\ref{tab:all_multi_task_prompts} as the input of the translation API and manually recover the placeholders in the API output based on brackets markers (e.g. ``\texttt{\{Sentence 1\}} because \mask'' is translated to ``\texttt{\{Sentence 1\}}\zhtext{因为}\mask''). When the candidate set is closed, we replace \mask with each verbalized label and translate them separately. For example, ``\texttt{\{Sentence 1\}}, right? Yes, \texttt{\{Sentence 2\}}'' is translated to ``\texttt{\{Sentence 1\}}\zhtext{，对吗？是的，}\texttt{\{Sentence 2\}}''.
% For \emph{high-resource} prompting, we % directly apply the English prompts to the examples in other languages. 
% use English prompts for all languages. % To eliminate the noise from machine translation errors, 
On XNLI, we also compared to prompts manually translated from English to eliminate the impact of translation noise on the comparison.\footnote{We ask native speakers to translate the English template into \lang{zh}, \lang{es}, \lang{fr}, \lang{el}, \lang{hi}, \lang{vi}, \lang{ar} and \lang{bg}. For the rest of the languages, one of the authors verified and corrected the machine translated templates using bilingual dictionaries.}

\begin{table*}[ht]
\centering
\scalebox{0.72}{
% \begin{tabular}{lrrrrrrrrrrrrrrrrr}
% \toprule
%  \textit{XNLI} &              en &   ar &   bg &   de &   el &  es &   fr &   hi &   ru &   sw &   th &   tr &   ur &  vi &  zh & % Avg & Std\\
% \cline{2-18}
% En temp. &           54.5 & \highest{47.5} & \highest{50.0} & 42.4 & \highest{47.1} & 38.2 & \highest{50.7} & \highest{43.4} & 47.2 & \highest{46.2} & \highest{46.1} & 44.6 & \highest{42.7} & \highest{47.5} & \highest{45.0} & \highest{46.2} & \highest{3.9} \\
% MT temp.  &          54.5 & 46.5 & 49.3 & 37.6 & 45.4 & 37.5 & 47.8 & 38.8 & 40.7 & 44.4 & 33.3 & 37.1 & 33.5 & 35.4 & 34.6 & 41.1 & 6.5\\
% HT temp.  &          54.5 & 46.0 & 49.3 & \highest{49.5} & \highest{47.1} & \highest{50.0} & 50.4 & 37.5 & \highest{47.4} & 44.4 & 34.9 & \highest{46.3} & 33.7 & 45.2 & 34.1 & 44.7 & 6.6 \\
% \midrule
%  \textit{XCOPA} &       en &   et &   ht &   id &   it &   qu &   ru &   sw &   ta &   th &   tr &   vi &   zh & & & Avg & Std\\
% \cline{2-18}
% En temp. &           69.0 & \highest{63.0} & 57.0 & 54.0 & 59.0 & \highest{70.0} & \highest{64.0} & \highest{54.0} & \highest{59.0} & \highest{53.0} & \highest{65.0} & 63.0 & \highest{63.0} & & & \highest{61.0} & \highest{5.5} \\
% MT temp. &           69.0 & 60.0 & \highest{60.0} & 51.0 & \highest{69.0} & 69.0 & 59.0 & 51.0 & \highest{59.0} & 48.0 & 64.0 & \highest{64.0} & 62.0 & & & 60.4 & 7.0\\
% No temp. &           69.0 & 61.0 & 55.0 & \highest{55.0} & 56.0 & 69.0 & 59.0 & 52.0 & 54.0 & 51.0 & 63.0 & 55.0 & 62.0 & & & 58.5 & 5.9 \\
% \bottomrule
% \end{tabular}
\begin{tabular}{ll|rrrrrr|rrrrrr|rrr|r}
\toprule
   & & \multicolumn{6}{l}{hi} & \multicolumn{6}{l}{medium} & \multicolumn{3}{l}{low} & \\
 \#\ shot & prompt &       en &   zh &   ru &   es &   de &   fr &    el &   th &   vi &   tr &   ar &   bg &   hi &   ur &   sw  & Avg.\\
\midrule
0  & En &           54.5 & \highest{45.0} & 47.2 & 38.2 & 42.4 & \highest{50.7} &  \highest{47.1} & \highest{46.1} & \highest{47.5} & 44.6 & \highest{47.5} & \highest{50.0} & \highest{43.4} & \highest{42.7} & \highest{46.2} & \highest{46.2}\\
   & MT &           54.5 & 34.6 & 40.7 & 37.5 & 37.6 & 47.8 &  45.4 & 33.3 & 35.4 & 37.1 & 46.5 & 49.3 & 38.8 & 33.5 & 44.4 & 41.1\\
   & HT &           54.5 & 34.1 & \highest{47.4} & \highest{50.0} & \highest{49.5} & 50.4 &  \highest{47.1} & 34.9 & 45.2 & \highest{46.3} & 46.0 & 49.3 & 37.5 & 33.7 & 44.4 & 44.7\\
\midrule
4  & En &           51.8 & \highest{48.0} & 48.2 & 45.1 & 45.2 & 49.2 &  \highest{48.4} & \highest{46.3} & \highest{48.2} & 45.1 & \highest{46.4} & \highest{49.1} & \highest{46.6} & \highest{43.5} & \highest{44.9} & \highest{47.1}\\
   & MT &           51.8 & 39.8 & 45.1 & 45.8 & 43.6 & \highest{49.4} &  44.3 & 33.7 & 41.9 & 35.0 & 45.5 & 48.6 & 39.9 & 35.4 & 43.5 & 42.9\\
   & HT &           51.8 & 39.8 & \highest{50.0} & \highest{49.9} & \highest{49.8} & 45.7 &  44.0 & 37.3 & 41.9 & \highest{47.0} & 45.7 & 48.6 & 40.2 & 35.0 & 43.5 & 44.7\\
\midrule
32 & En &           53.4 & \highest{50.4} & 40.0 & 46.4 & 46.2 & 46.7 &  47.3 & \highest{46.8} & \highest{48.6} & 44.3 & 43.3 & 42.8 & \highest{45.6} & \highest{45.6} & \highest{46.5} & \highest{46.3} \\
   & MT &           53.4 & 40.7 & 38.0 & 47.0 & 43.1 & \highest{49.1} &  48.5 & 35.0 & 44.5 & 35.2 & 41.9 & \highest{51.9} & 37.2 & 34.8 & 43.8 & 42.9\\
   & HT &           53.4 & 43.1 & \highest{51.8} & \highest{51.7} & \highest{49.6} & 46.2 &  \highest{48.9} & 38.0 & 47.0 & \highest{47.5} & \highest{43.9} & \highest{51.9} & 40.9 & 34.6 & 43.8 & 46.2\\
\bottomrule
\end{tabular}
}
\caption{Comparison between English prompts, MT (machine-translated) prompts and HT (human-translated) prompts for 0, 4 and 32-shot learning on XNLI dev set using \model{\Oursns}{7.5B}.} % \texttt{En}: English. \texttt{MT}: machine translation. \texttt{HT}: Human verified translation.}
\label{tab:xnli_template_analysis}
\end{table*}

\begin{table*}[ht]
    \centering
    \scalebox{0.77}{
    \begin{tabular}{ll|rr|rrrrr|rrr|rr|r}
    \toprule
       & & \multicolumn{2}{l}{hi} & \multicolumn{5}{l}{medium} & \multicolumn{3}{l}{low} & \multicolumn{2}{l}{ex-low} \\
    \#\ shot & prompt &             zh &   ru &    it &   id &   th &   vi &   tr &   et &   ta &   sw &    ht &   qu & Avg.\\
    \midrule
    0  & En &           \highest{63.0} & \highest{64.0} &  59.0 & \highest{54.0} & \highest{53.0} & 63.0 & \highest{65.0} & \highest{63.0} & \highest{59.0} & \highest{54.0} &  57.0 & \highest{70.0} & \highest{60.3} \\
       & MT &           62.0 & 59.0 &  \highest{69.0} & 51.0 & 48.0 & \highest{64.0} & 64.0 & 60.0 & \highest{59.0} & 51.0 &  \highest{60.0} & 69.0 & 59.7 \\
    %    & Non-verbal &           62.0 & 59.0 &  56.0 & \highest{55.0} & 51.0 & 55.0 & 63.0 & 61.0 & 54.0 & 52.0 &  55.0 & 69.0 & 57.7 \\
    \midrule
    4  & En &           66.8 & \highest{73.8} &  66.2 & \highest{54.6} & \highest{57.4} & \highest{68.8} & 69.8 & \highest{69.2} & 60.6 & 58.2 &  \highest{62.6} & \highest{62.2} & \highest{64.2} \\
       & MT &           \highest{68.4} & 65.4 &  \highest{68.8} & 54.0 & 56.6 & 67.2 & 67.4 & 66.6 & \highest{62.2} & \highest{60.2} &  60.8 & 61.8 & 63.3 \\
    %    & Non-verbal &           65.0 & 65.4 &  64.8 & 55.2 & \highest{58.0} & 61.2 & 68.0 & 65.0 & 60.4 & 54.0 &  55.2 & 61.8 & 61.2 \\
    % \midrule
    % 32 & En &           74.6 & 80.8 &  77.4 & 67.4 & 72.0 & 78.0 & 77.8 & 76.2 & 71.2 & 70.4 &  75.0 & 72.2 \\
    %    & MT &           77.0 & 75.0 &  77.8 & 68.4 & 72.6 & 79.4 & 79.8 & 76.2 & 72.0 & 70.8 &  73.6 & 71.0 \\
    %    & Non-verbal &           73.4 & 75.0 &  74.0 & 67.6 & 71.4 & 73.8 & 77.0 & 73.8 & 69.4 & 67.6 &  70.6 & 71.0 \\
    \bottomrule
    \end{tabular}
    }
    \caption{Comparison bewtween English prompts and MT (machine-translated) prompts for 0-shot and 4-shot learning on XCOPA dev set using \model{\Ours}{7.5B}.} % \texttt{En}: English. \texttt{MT}: machine translation.}
    \label{tab:xcopa_template_analysis}
\end{table*}

As shown in Table~\ref{tab:xnli_template_analysis} and Table~\ref{tab:xcopa_template_analysis}, the in-context learning performance % of \model{\Ours}{7.5B} 
is sensitive to the prompting choices across all languages.
% , especially when viewed through the lens of multiple languages. % (e.g. \lang{zh} and \lang{ur}). 
For both XNLI and XCOPA, using the English prompts on average yield significantly better performance than using the machine-translated prompts. % in all three settings. 
For XNLI, human translated (HT) prompts significantly improve over machine translated (MT) prompts for most languages. Surprisingly, the performance of human translated prompts lags % significantly 
behind that of the English prompts in the 0-shot and 4-shot settings. 

Further examination of the per-language performance reveals that the relative strengths of different prompting approaches vary across languages. For \lang{es} and \lang{de}, HT prompts offer large gains compared to the MT prompts and the English prompts. However, for \lang{zh} and \lang{ur}, using translated prompts (either HT or MT) significantly hurts the performance. For \lang{zh}, \lang{fr}, \lang{vi}, \lang{ar} and \lang{hi}, using native-speaker translated prompts still yields significantly lower performance compared to using the English prompts in at least one setting, suggesting that translation error is not the sole cause of the performance drop.

\subsection{Full Results on Learning from Cross-lingual Demonstrations}
\label{sec:cross-lingual-in-context-learning-full}
We evaluated \model{\Ours}{7.5B} on XNLI in the learning from cross-lingual demonstration setting, using both the \emph{same-language-prompting} and \emph{English-prompting} setups. In \emph{same-language-prompting}, the prompt fields and the examples are always in the same language. And in \emph{English-prompting}, English prompts are used for all examples. All few-shot performances in this section are obtained using the $k$-shot \emph{per label} setting as described in \S\ref{sec:few-shot-training-set-distribution}.

\begin{figure*}[ht!]
    \centering
    \begin{subfigure}[b]{0.49\textwidth}
        \centering
        \includegraphics[width=\linewidth]{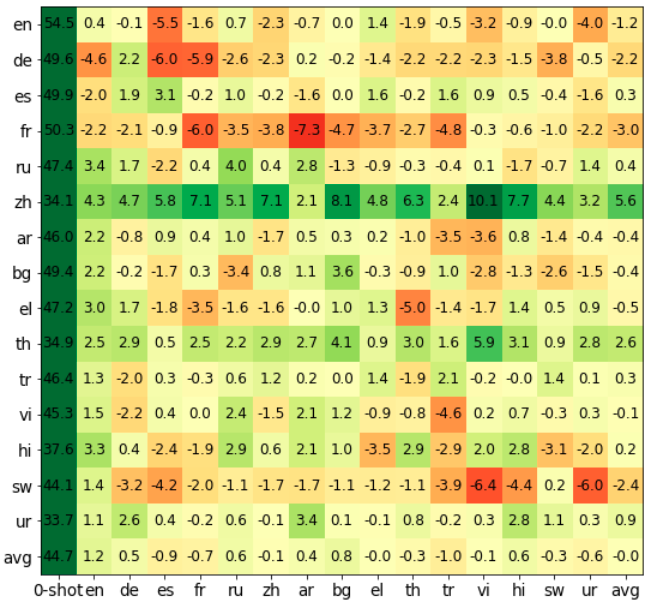}
        \caption{Same-language prompting with human-translated templates}
    \end{subfigure}
    \hfill
    \begin{subfigure}[b]{0.49\textwidth}
        \centering
        \includegraphics[width=\linewidth]{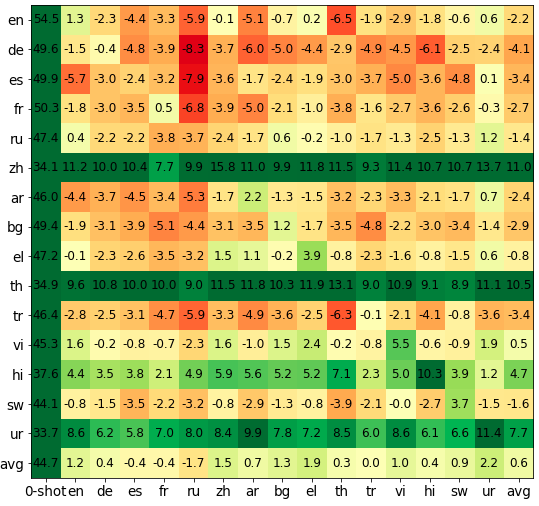}
        \caption{English prompting for all languages}
    \end{subfigure}
    \caption{Cross-lingual few-shot in-context learning on XNLI development set. The leftmost column shows the 0-shot performance (\emph{with human-translated templates}) of each language. The rest of the matrix shows the difference between 4-shot (per label) and 0-shot (\emph{with human-translated templates}) performance. Row: target language. Column: source language. The languages are ordered from the highest to the lowest resource level. We observe that using demonstration examples from the source language improves the zero-shot performance in the target language over a number of language pairs, and the improvement is more significant from higher-resourced languages to lower-resourced languages.} 
    \label{fig:crosslingual-xnli-over-ht}
\end{figure*}

\begin{figure*}[ht!]
    \centering
    \begin{subfigure}[b]{0.49\textwidth}
        \centering
        \includegraphics[width=\linewidth]{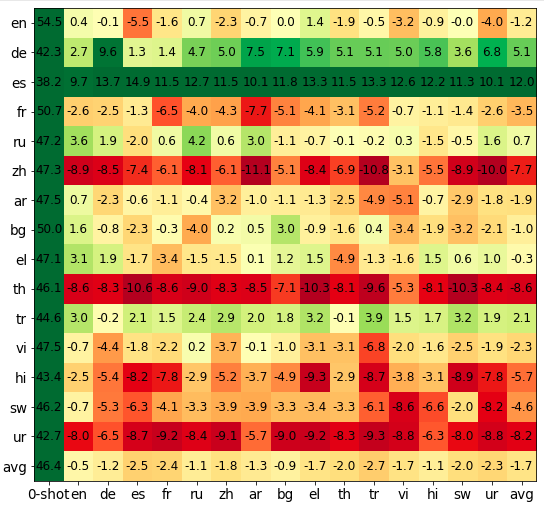}
        \caption{Same-language prompting with human-translated templates}
    \end{subfigure}
    \hfill
    \begin{subfigure}[b]{0.49\textwidth}
        \centering
        \includegraphics[width=\linewidth]{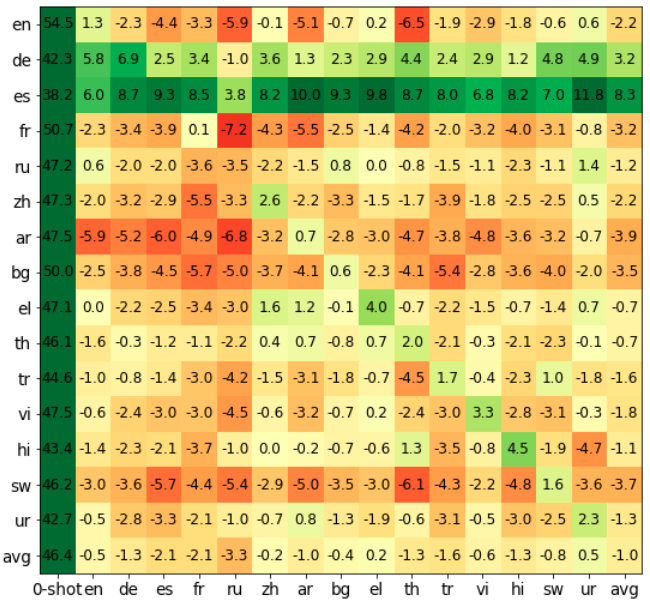}
        \caption{English prompting}
    \end{subfigure}
    \caption{Cross-lingual few-shot in-context learning on XNLI development set. The leftmost column shows the 0-shot performance (\emph{with English templates}) of each language. The rest of the matrix shows the difference between 4-shot (per label) and 0-shot (\emph{with English templates}) performance. Row: target language. Column: source language.} 
    \label{fig:crosslingual-xnli-over-en}
\end{figure*}

As shown in Figure~\ref{fig:crosslingual-xnli-over-ht}, for many language pairs transferring from source language demonstration can significantly improve over the zero-shot performance in the target language when human-translated templates is used. The improvement is especially significant for languages such as Chinese (\lang{zh}), Thai (\lang{th}) and Urdu (\lang{ur}), whose zero-shot performance is close to random with human translated templates. However, we found that the effect of cross-lingual transfer from template and cross-lingual transfer from demonstration examples typically do not add up. As shown in Figure~\ref{fig:crosslingual-xnli-over-en}, using the English template significantly improves the zero-shot performance of most languages, including Chinese, Thai and Urdu. In this case, the demonstration examples in general do not help unless they are in the same language as the target example (diagonals).

\begin{figure*}[ht!]
    \centering
    \begin{subfigure}[b]{0.49\textwidth}
        \centering
        \includegraphics[width=\linewidth]{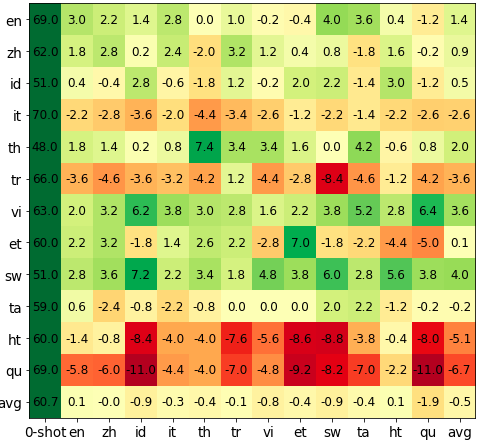}
        \caption{Same language prompting with human-translated templates}
    \end{subfigure}
    \hfill
    \begin{subfigure}[b]{0.49\textwidth}
        \centering
        \includegraphics[width=\linewidth]{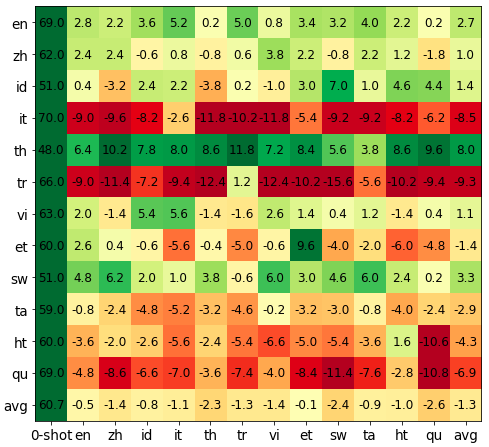}
        \caption{English prompting}
    \end{subfigure}
    \caption{Cross-lingual few-shot in-context learning on XCOPA development set. The leftmost column shows the 0-shot (\emph{with machine translated templates}) performance. The rest of the matrix shows the difference between 4-shot (per label) and 0-shot (\emph{with machine translated templates}) performance. Row: target language. Column: source language.} 
    \label{fig:crosslingual-xcopa-over-ht}
\end{figure*}

Figure~\ref{fig:crosslingual-xcopa-over-ht} shows the results on XCOPA.

\begin{figure*}[ht!]
    \centering
    \includegraphics[width=.45\linewidth]{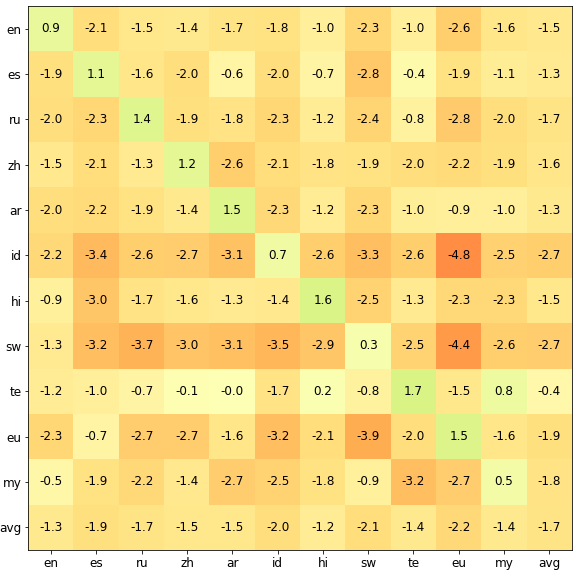}
    \caption{Cross-lingual few-shot in-context learning on XStoryCloze test set. The matrix shows the difference between 4-shot (per label) and 0-shot performance. For XStoryCloze, there is no difference between same-language prompting and English prompting since the task does not use a verbalized template. Row: target language. Column: source language.} 
    \label{fig:crosslingual-xstorycloze}
\end{figure*}

Figure~\ref{fig:crosslingual-xstorycloze} shows the results on XStoryCloze, where we observed almost no improvement for any language pair. Possible reasons for the poor transfer results on XStoryCloze is that it requires reasoning about implicit relations between multiple sentences which is much harder to do especially in a cross-lingual setting. 

\FloatBarrier
\subsection{Full Results in FLORES-101} \label{app:mt}

Table \ref{tab:flores101_full} reports our full results in FLORES-101.

\begin{table*}[ht!]
\centering
\begin{small}
\addtolength{\tabcolsep}{-2.5pt}
\scalebox{0.9}{
\begin{tabular}{lrcccccccccccccc|c}
\toprule
&& en & de & fr & ca & fi & ru & bg & zh & ko & ar & sw & hi & my & ta & avg \\
\midrule
\multirow{3}{*}{en}
& Supervised & -- & 32.6 & 42.0 & 31.2 & 24.2 & 27.1 & 37.4 & 19.3 & 18.5 & 17.9 & 26.9 & 28.1 & 3.5 & 3.4 & 24.0 \\
& \model{GPT-3}{6.7B} & -- & 25.9 & \textbf{36.1} & 23.8 & 10.2 & 11.2 & 5.9 & 12.5 & 1.2 & 1.1 & 0.5 & 0.3 & 0.1 & 0.0 & 9.9 \\
& \model{\Oursns}{7.5B} & -- & \textbf{27.6} & 36.0 & \textbf{\underline{34.0}} & \textbf{23.3} & \textbf{24.2} & \textbf{33.1} & \textbf{15.6} & \textbf{12.0} & \textbf{11.5} & \textbf{18.0} & \textbf{19.9} & \textbf{\underline{11.0}} & \textbf{\underline{8.5}} & \textbf{21.1} \\
\midrule
\multirow{3}{*}{de}
& Supervised & 35.8 & -- & 35.5 & 25.8 & 22.6 & 24.6 & 31.5 & 17.2 & 16.6 & 14.8 & 21.0 & 23.4 & 2.3 & 2.3 & 21.0 \\
& \model{GPT-3}{6.7B} & \textbf{\underline{40.4}} & -- & 26.2 & 17.2 & 8.1 & 9.3 & 4.8 & 9.0 & 1.0 & 0.9 & 0.5 & 0.3 & 0.1 & 0.1 & 9.1 \\
& \model{\Oursns}{7.5B} & \underline{38.8} & -- & \textbf{27.9} & \textbf{19.1} & \textbf{20.5} & \textbf{19.7} & \textbf{25.8} & \textbf{12.3} & \textbf{3.4} & \textbf{6.6} & \textbf{11.7} & \textbf{14.3} & \textbf{\underline{9.9}} & \textbf{\underline{4.8}} & \textbf{16.5} \\
\midrule
\multirow{3}{*}{fr}
& Supervised & 37.2 & 28.5 & -- & 28.7 & 21.9 & 24.5 & 32.2 & 17.6 & 16.7 & 15.4 & 17.2 & 22.9 & 2.1 & 0.8 & 20.4 \\
& \model{GPT-3}{6.7B} & \textbf{\underline{42.8}} & \textbf{20.9} & -- & 23.7 & 8.0 & 9.7 & 4.6 & 9.1 & 1.0 & 1.0 & 0.4 & 0.3 & 0.1 & 0.0 & 9.4 \\
& \model{\Oursns}{7.5B} & \underline{40.4} & 20.4 & -- & \textbf{\underline{32.1}} & \textbf{19.4} & \textbf{19.8} & \textbf{26.3} & \textbf{10.6} & \textbf{2.4} & \textbf{5.9} & \textbf{14.5} & \textbf{13.7} & \textbf{\underline{9.7}} & \textbf{\underline{6.6}} & \textbf{17.1} \\
\midrule
\multirow{3}{*}{ca}
& Supervised & 33.4 & 24.8 & 35.1 & -- & 19.0 & 21.1 & 28.6 & 15.1 & 13.9 & 13.4 & 18.7 & 20.5 & 2.1 & 2.6 & 19.1 \\
& \model{GPT-3}{6.7B} & \underline{40.2} & 18.6 & 31.4 & -- & 7.0 & \textbf{9.3} & 4.3 & 8.0 & 0.9 & \textbf{0.9} & 0.3 & 0.4 & 0.1 & 0.1 & 9.3 \\
& \model{\Oursns}{7.5B} & \textbf{\underline{41.1}} & \textbf{18.9} & \textbf{33.8} & -- & \textbf{11.3} & 3.3 & \textbf{23.9} & \textbf{10.8} & \textbf{1.3} & 0.8 & \textbf{13.8} & \textbf{6.1} & \textbf{\underline{7.9}} & \textbf{\underline{3.1}} & \textbf{13.6} \\
\midrule
\multirow{3}{*}{fi}
& Supervised & 27.2 & 23.0 & 29.3 & 21.6 & -- & 20.6 & 26.4 & 16.0 & 14.8 & 12.4 & 14.2 & 19.8 & 1.7 & 0.9 & 17.5 \\
& \model{GPT-3}{6.7B} & 25.3 & 13.5 & 17.1 & 10.0 & -- & 6.4 & 2.8 & 5.7 & 0.7 & 0.7 & 0.3 & 0.3 & 0.1 & 0.0 & 6.4 \\
& \model{\Oursns}{7.5B} & \textbf{\underline{29.2}} & \textbf{17.4} & \textbf{22.2} & \textbf{17.0} & -- & \textbf{16.5} & \textbf{17.5} & \textbf{12.4} & \textbf{7.5} & \textbf{7.6} & \textbf{8.0} & \textbf{10.1} & \textbf{\underline{6.2}} & \textbf{\underline{2.0}} & \textbf{13.4} \\
\midrule
\multirow{3}{*}{ru}
& Supervised & 27.5 & 23.5 & 30.1 & 22.0 & 19.4 & -- & 31.0 & 16.5 & 15.3 & 13.5 & 18.1 & 20.9 & 2.2 & 2.3 & 18.6 \\
& \model{GPT-3}{6.7B} & \underline{28.1} & 14.8 & 20.4 & 13.1 & 5.4 & -- & 7.4 & 1.2 & 0.2 & 0.2 & 0.1 & 0.2 & 0.1 & 0.1 & 7.0 \\
& \model{\Oursns}{7.5B} & \textbf{\underline{30.4}} & \textbf{17.9} & \textbf{24.0} & \textbf{14.6} & \textbf{8.0} & -- & \textbf{26.3} & \textbf{11.6} & \textbf{5.5} & \textbf{7.4} & \textbf{7.1} & \textbf{9.1} & \textbf{\underline{7.3}} & \textbf{\underline{3.1}} & \textbf{13.2} \\
\midrule
\multirow{3}{*}{bg}
& Supervised & 33.0 & 26.1 & 33.7 & 24.9 & 20.8 & 26.5 & -- & 17.5 & 16.4 & 14.5 & 20.9 & 23.1 & 2.3 & 2.4 & 20.2 \\
& \model{GPT-3}{6.7B} & 21.6 & 11.4 & 16.0 & 9.7 & 4.3 & 6.5 & -- & 1.2 & 0.2 & 0.2 & 0.1 & 0.2 & 0.1 & 0.1 & 5.5 \\
& \model{\Oursns}{7.5B} & \textbf{\underline{35.5}} & \textbf{19.2} & \textbf{26.3} & \textbf{12.9} & \textbf{14.2} & \textbf{22.9} & -- & \textbf{11.9} & \textbf{6.8} & \textbf{9.2} & \textbf{9.4} & \textbf{7.5} & \textbf{\underline{3.2}} & \textbf{1.0} & \textbf{13.9} \\
\midrule
\multirow{3}{*}{zh}
& Supervised & 20.9 & 17.6 & 24.3 & 17.4 & 16.0 & 17.2 & 22.1 & -- & 15.9 & 11.6 & 15.5 & 18.5 & 1.9 & 2.5 & 15.5 \\
& \model{GPT-3}{6.7B} & \textbf{\underline{21.1}} & \textbf{9.5} & \textbf{14.3} & 8.2 & 4.3 & 3.6 & 1.3 & -- & 1.1 & 0.4 & 0.2 & 0.2 & 0.1 & 0.0 & 4.9 \\
& \model{\Oursns}{7.5B} & 20.7 & 8.3 & 8.5 & \textbf{10.5} & \textbf{4.4} & \textbf{4.8} & \textbf{14.8} & -- & \textbf{9.3} & \textbf{4.2} & \textbf{5.6} & \textbf{12.0} & \textbf{\underline{8.6}} & \textbf{\underline{6.2}} & \textbf{9.1} \\
\midrule
\multirow{3}{*}{ko}
& Supervised & 20.9 & 16.7 & 22.1 & 16.5 & 14.9 & 15.5 & 21.1 & 15.7 & -- & 10.6 & 15.1 & 18.7 & 1.9 & 4.0 & 14.9 \\
& \model{GPT-3}{6.7B} & 8.3 & 4.6 & 6.4 & 4.4 & \textbf{2.1} & \textbf{1.7} & 0.8 & 2.5 & -- & 0.2 & 0.1 & 0.1 & 0.1 & 0.1 & 2.4 \\
& \model{\Oursns}{7.5B} & \textbf{19.9} & \textbf{10.3} & \textbf{13.7} & \textbf{5.3} & 1.4 & 1.2 & \textbf{10.9} & \textbf{11.9} & -- & \textbf{2.7} & \textbf{3.2} & \textbf{1.0} & \textbf{\underline{2.2}} & \textbf{1.4} & \textbf{6.5} \\
\midrule
\multirow{3}{*}{ar}
& Supervised & 25.5 & 18.7 & 25.7 & 18.9 & 15.6 & 17.8 & 23.8 & 13.1 & 13.3 & -- & 15.4 & 19.4 & 1.8 & 0.9 & 16.1 \\
& \model{GPT-3}{6.7B} & 10.5 & 5.3 & 9.6 & 6.0 & 2.2 & 2.2 & 0.9 & 0.9 & 0.1 & -- & 0.1 & 0.1 & 0.2 & 0.0 & 2.9 \\
& \model{\Oursns}{7.5B} & \textbf{\underline{27.7}} & \textbf{12.2} & \textbf{17.9} & \textbf{8.8} & \textbf{8.5} & \textbf{9.1} & \textbf{18.4} & \textbf{8.9} & \textbf{0.8} & -- & \textbf{7.7} & \textbf{7.8} & \textbf{\underline{3.4}} & \textbf{\underline{3.7}} & \textbf{10.4} \\
\midrule
\multirow{3}{*}{sw}
& Supervised & 30.4 & 19.4 & 26.7 & 20.1 & 15.6 & 17.6 & 23.8 & 13.2 & 12.2 & 12.0 & -- & 19.2 & 2.1 & 4.0 & 16.6 \\
& \model{GPT-3}{6.7B} & 5.0 & 2.9 & 3.9 & 2.8 & 1.7 & 1.8 & 1.3 & 1.3 & 0.5 & 0.5 & -- & 0.4 & 0.1 & 0.1 & 1.7 \\
& \model{\Oursns}{7.5B} & \textbf{\underline{31.6}} & \textbf{13.4} & \textbf{21.8} & \textbf{15.4} & \textbf{10.2} & \textbf{13.1} & \textbf{15.2} & \textbf{9.5} & \textbf{6.0} & \textbf{8.9} & -- & \textbf{7.6} & \textbf{\underline{3.4}} & \textbf{1.0} & \textbf{12.1} \\
\midrule
\multirow{3}{*}{hi}
& Supervised & 27.9 & 19.4 & 25.9 & 18.9 & 15.7 & 16.9 & 23.9 & 13.5 & 13.9 & 12.2 & 16.8 & -- & 2.5 & 3.8 & 16.2 \\
& \model{GPT-3}{6.7B} & 1.2 & 0.9 & 1.4 & 0.8 & 0.4 & 0.4 & 0.3 & 0.2 & 0.1 & 0.1 & 0.1 & -- & 0.1 & 0.2 & 0.5 \\
& \model{\Oursns}{7.5B} & \textbf{25.2} & \textbf{12.3} & \textbf{15.4} & \textbf{8.8} & \textbf{9.8} & \textbf{11.5} & \textbf{11.3} & \textbf{10.8} & \textbf{8.5} & \textbf{6.1} & \textbf{4.7} & -- & \textbf{1.5} & \textbf{1.9} & \textbf{9.8} \\
\midrule
\multirow{3}{*}{my}
& Supervised & 10.0 & 6.9 & 10.4 & 8.5 & 6.0 & 6.7 & 9.5 & 5.7 & 6.1 & 4.6 & 7.2 & 9.1 & -- & 2.5 & 7.2 \\
& \model{GPT-3}{6.7B} & 0.5 & 0.3 & 0.4 & 0.4 & 0.2 & 0.1 & 0.2 & 0.0 & 0.0 & 0.0 & 0.1 & 0.2 & -- & 0.1 & 0.2 \\
& \model{\Oursns}{7.5B} & \textbf{\underline{14.1}} & \textbf{\underline{7.6}} & \textbf{10.1} & \textbf{3.8} & \textbf{5.7} & \textbf{\underline{7.1}} & \textbf{8.9} & \textbf{\underline{7.1}} & \textbf{\underline{6.9}} & \textbf{3.6} & \textbf{3.5} & \textbf{8.9} & -- & \textbf{\underline{2.6}} & \textbf{6.9} \\
\midrule
\multirow{3}{*}{ta}
& Supervised & 8.3 & 4.9 & 6.8 & 5.8 & 5.0 & 4.7 & 7.0 & 2.5 & 2.3 & 1.1 & 5.2 & 6.9 & 1.2 & -- & 4.8 \\
& \model{GPT-3}{6.7B} & 1.0 & 0.5 & 0.8 & 0.5 & 0.2 & 0.3 & 0.3 & 0.1 & 0.2 & 0.1 & 0.1 & 0.2 & 0.0 & -- & 0.3 \\
& \model{\Oursns}{7.5B} & \textbf{\underline{16.3}} & \textbf{\underline{8.4}} & \textbf{\underline{10.3}} & \textbf{5.1} & \textbf{\underline{5.2}} & \textbf{\underline{8.1}} & \textbf{\underline{7.6}} & \textbf{\underline{8.1}} & \textbf{\underline{6.2}} & \textbf{\underline{5.4}} & \textbf{2.8} & \textbf{\underline{7.2}} & \textbf{0.9} & -- & \textbf{\underline{7.1}} \\
\midrule
\multirow{3}{*}{avg}
& Supervised & 26.0 & 20.2 & 26.7 & 20.0 & 16.7 & 18.5 & 24.5 & 14.1 & 13.5 & 11.8 & 16.3 & 19.3 & 2.1 & 2.5 & 16.6 \\
& \model{GPT-3}{6.7B} & 18.9 & 9.9 & 14.2 & 9.3 & 4.2 & 4.8 & 2.7 & 4.0 & 0.6 & 0.5 & 0.2 & 0.3 & 0.1 & 0.1 & 5.0 \\
& \model{\Oursns}{7.5B} & \textbf{\underline{28.5}} & \textbf{14.9} & \textbf{20.6} & \textbf{14.4} & \textbf{10.9} & \textbf{12.4} & \textbf{18.5} & \textbf{10.9} & \textbf{5.9} & \textbf{6.1} & \textbf{8.5} & \textbf{9.7} & \textbf{\underline{5.8}} & \textbf{\underline{3.5}} & \textbf{12.2} \\
\bottomrule
\end{tabular}
}
\end{small}
\caption{Machine translation results on FLORES-101 devtest (spBLEU). Source language in rows, target language in columns. \model{GPT-3}{6.7B} and \model{\Oursns}{7.5B} use 32 examples from the dev set for few-shot learning. Supervised results correspond to the M2M-124 615M model from \citet{goyal2021flores}. Underline denotes better than supervised, bold denotes best of GPT-3 and XGLM. spBLEU computed using the % official 
implementation from \citet{goyal2021flores}.}
\label{tab:flores101_full}
\end{table*}

\subsection{Majority Label Bias}
\label{sec:few-shot-training-set-distribution}

\begin{table}[t!]
    \centering
    \scalebox{0.8}{
        \begin{tabular}{l|lll|lll}
            \toprule
            &  \multicolumn{3}{c}{24-shot} & \multicolumn{3}{c}{48-shot} \\
            Seed & E & N & C & E & N & C \\
            \midrule
            0 & 9 & 7 & 8 & 22 & 11 & 15 \\
            1 & 6 & 11 & 7 & 12 & 16 & 20 \\
            2 & 9 & 7 & 8 & 20 & 11 & 17 \\
            3 & 8 & 11 & 5 & 16 & 16 & 16 \\
            4 & 6 & 4 & 14 & 14 & 16 & 18 \\
            \bottomrule
        \end{tabular}
    }
    \caption{Distribution of XNLI few-shot training sets obtained by randomly sampling from the original dev set. \texttt{E}: entailment, \texttt{N}: neutral, \texttt{C}: contradition.} \label{tab:xnli-random-sample-train-set-distribution}
\end{table}

\begin{table*}[t!]
\centering
\scalebox{0.66}{
    \setlength{\tabcolsep}{4.2pt}
    \begin{tabular}{ll|llllll|llllll|lll|ll}
    \toprule
      \multirow{2}{*}{\makecell{\#\\shots}} & & \multicolumn{6}{l}{high} & \multicolumn{6}{l}{medium} & \multicolumn{3}{l}{low} \\
       &  &                           en &                           zh &                           ru &                           es &                           de &                           fr &                           el &                           th &                           vi &                           tr &                           ar &                           bg &                           hi &                           ur &                           sw  & Avg & Std\\
    \midrule
    \multirow{2}{*}{24} & rand. &  \makecell{54.0\\($\pm2.3$)} &  \makecell{50.0\\($\pm1.9$)} &  \makecell{41.5\\($\pm6.1$)} &  \makecell{47.1\\($\pm4.0$)} &  \makecell{46.1\\($\pm2.1$)} &  \makecell{47.7\\($\pm4.1$)} &  \makecell{48.8\\($\pm2.2$)} &  \makecell{47.8\\($\pm2.7$)} &  \makecell{48.8\\($\pm3.5$)} &  \makecell{44.8\\($\pm2.0$)} &  \makecell{45.4\\($\pm3.7$)} &  \makecell{44.4\\($\pm4.8$)} &  \makecell{45.8\\($\pm2.4$)} &  \makecell{45.6\\($\pm1.4$)} &  \makecell{46.4\\($\pm3.0$)} & 47.0 & 2.9 \\
       & unif. &  \makecell{\highest{56.0}\\($\pm1.4$)} &  \makecell{\highest{52.1}\\($\pm0.8$)} &  \makecell{\highest{42.9}\\($\pm1.2$)} &  \makecell{\highest{47.6}\\($\pm0.8$)} &  \makecell{\highest{49.2}\\($\pm1.0$)} &  \makecell{\highest{49.2}\\($\pm1.6$)} &  \makecell{\highest{50.9}\\($\pm1.6$)} &  \makecell{\highest{48.0}\\($\pm1.2$)} &  \makecell{\highest{50.2}\\($\pm1.5$)} &  \makecell{\highest{47.5}\\($\pm1.0$)} &  \makecell{\highest{47.3}\\($\pm0.8$)} &  \makecell{\highest{47.0}\\($\pm2.2$)} &  \makecell{\highest{49.4}\\($\pm1.2$)} &  \makecell{\highest{46.6}\\($\pm0.9$)} &  \makecell{\highest{47.4}\\($\pm0.8$)} & \highest{48.7} & 2.9\\\midrule
    \multirow{2}{*}{Max}  & rand. &  \makecell{54.2\\($\pm2.2$)} &  \makecell{51.7\\($\pm1.1$)} &  \makecell{37.4\\($\pm1.3$)} &  \makecell{45.4\\($\pm3.2$)} &  \makecell{44.7\\($\pm3.2$)} &  \makecell{45.0\\($\pm3.0$)} &  \makecell{46.9\\($\pm2.6$)} &  \makecell{45.8\\($\pm2.3$)} &  \makecell{47.9\\($\pm2.7$)} &  \makecell{42.0\\($\pm1.4$)} &  \makecell{42.3\\($\pm1.8$)} &  \makecell{40.4\\($\pm1.6$)} &  \makecell{45.6\\($\pm2.2$)} &  \makecell{46.2\\($\pm0.7$)} &  \makecell{46.5\\($\pm2.2$)} & 45.5 & 4.1\\
       & unif. &  \makecell{\highest{54.3}\\($\pm0.5$)} &  \makecell{\highest{52.2}\\($\pm0.4$)} &  \makecell{\highest{39.1}\\($\pm1.0$)} &  \makecell{\highest{45.8}\\($\pm0.9$)} &  \makecell{\highest{46.6}\\($\pm1.2$)} &  \makecell{\highest{46.4}\\($\pm1.8$)} &  \makecell{\highest{48.3}\\($\pm0.8$)} &  \makecell{\highest{47.3}\\($\pm0.9$)} &  \makecell{\highest{48.0}\\($\pm1.2$)} &  \makecell{\highest{45.0}\\($\pm2.3$)} &  \makecell{\highest{44.1}\\($\pm1.2$)} &  \makecell{\highest{42.9}\\($\pm2.1$)} &  \makecell{\highest{46.7}\\($\pm0.0$)} &  \makecell{\highest{46.3}\\($\pm0.0$)} &  \makecell{\highest{47.2}\\($\pm0.7$)} & \highest{46.7} & \highest{3.5} \\
    \bottomrule
    \end{tabular}
}
\caption{\model{\Oursns}{7.5B} in-language few-shot learning performance of on XNLI dev set using training sets of uniform class distribution and randomly sampled class distribution. We report the mean and standard deviation (in parentheses) of 5 different training sets sampled via different random seeds for each sampling strategy.}
\label{tb:comparing_few_shot_train_set_distribution}
\end{table*}

In the main paper, we define $k$-shot learning as learning from $k$ unique examples randomly drawn from the entire training population. This setting may lead to skewed few-shot training sets, especially when $k$ is small. As shown in Table~\ref{tab:xnli-random-sample-train-set-distribution}, the XNLI task is a three-way classification problem where the model needs to judge whether the relationship between a pair of sentences is ``entailment'', ``neurtral'' or ``contradiction''. While the original XNLI dev set has a uniform class distribution, the few-shot training sets randomly sampled\footnote{We implement our random sampling procedure using the \texttt{numpy.random.choice} function: \url{https://numpy.org/doc/stable/reference/random/generated/numpy.random.choice.html}.} from it often has a much more skewed class distribution. 

For a $|\mY|$-way classification task, % as an example randomly drawing an example for 1-shot learning 
a skewed training set distribution can cause the model to score the majority class as disproportionately more likely than %  rest of the $|\mL|-1$ 
the other classes. This was shown by~\citet{DBLP:conf/icml/ZhaoWFK021} as the \emph{majority label bias} problem. As a result, previous work such as~\citet{DBLP:journals/corr/abs-2109-03630} adopts a $k$-shot \textit{per class} setting, where $k$ unique examples are randomly drawn from each class to form a training set of size $k\times|\mY|$. 

% To quantify the \emph{majority label bias}, 
We compare learning from a uniform class distribution (randomly sampling $k$ examples per class) to learning from a more skewed distribution (randomly sampling $k\times|\mL|$ examples from the total population) on the XNLI task. We use the 24-shot and maximum fit (truncated 48-shot) settings. As shown in Table~\ref{tb:comparing_few_shot_train_set_distribution}, for both settings, learning from a uniform class distribution leads to significantly higher accuracy in all languages compared to learning from the skewed distributions. \lang{de}, \lang{tr}, \lang{bg}, \lang{hi} suffer the most learning from the skewed distributions ($>2$ absolute accuracy gap in the 24-shot setting), while \lang{es} suffers the least. Moreover, the variances among few-shot trials using different random seeds shrink considerably when the training set class distribution is uniform. % Same as the findings in~\S~\ref{sec:compare-to-small-lm}, we observe performance decreases for most languages going from the 24-shot to maximum fit setting.
These results highlight the severeness of the \emph{majority label bias} issue in the multilingual in-context learning framework.

% \subsection{Few-shot learning}
% \comment{Report final results on test set. Victoria to fill in.} 
% Performance with examples of 1, 4, 16, etc.
% What is the performance if finetuning with those examples? 

\hide{
\section{Hate Speech Detection Precision}
\label{sec:hate_speech_precision}
We compute the precision scores of different language models on the task of Hate Speech Detection. The results are presented in Table~\ref{tab:hate-speech-performance-precision}.

\begin{table*}[t!]
    \centering
    % \footnotesize
    % \setlength{\tabcolsep}{4pt}
    \scalebox{0.72}{
    \begin{tabular}{cccccccc}
        \toprule
         Model & language condition & \# shot & English& Italian & Portuguese & Polish & Spanish \\
         \midrule
         \multicolumn{8}{c}{Precision}\\
         \midrule
         \multirow{2}{*}{\xlmgen} & \multirow{2}{*}{code switched}&\cellcolor{gray!40}  0 &\cellcolor{gray!40}$54.3$&\cellcolor{gray!40}$52.5$&\cellcolor{gray!40}$47.9$&\cellcolor{gray!40}$53.2$&\cellcolor{gray!40}$51.7$\\
        & &\cellcolor{gray!10} 4 &\cellcolor{gray!10}$\textbf{54.1}$&\cellcolor{gray!10}$52.3$&\cellcolor{gray!10}$42.2$&\cellcolor{gray!10}$52.0$&\cellcolor{gray!10}$51.4$\\
                \midrule
         \multirow{2}{*}{\model{GPT-3}{6.7B}} & \multirow{2}{*}{code switched}& 
            \cellcolor{gray!40} 0 &\cellcolor{gray!40}$\textbf{58.6}$&\cellcolor{gray!40}$57.7$&\cellcolor{gray!40}$46.1$&\cellcolor{gray!40}$60.8$&\cellcolor{gray!40}$54.0$\\
            & & \cellcolor{gray!10}4 &\cellcolor{gray!10}$52.9$&\cellcolor{gray!10}$53.7$&\cellcolor{gray!10}$\textbf{44.4}$&\cellcolor{gray!10}$\textbf{54.1}$&\cellcolor{gray!10}$\textbf{53.2}$\\
                \midrule
         \multirow{2}{*}{\model{\Ours}{7.5B}} & \multirow{2}{*}{code switched}& 
            \cellcolor{gray!40}0 &\cellcolor{gray!40}$54.2$&\cellcolor{gray!40}$56.1$&\cellcolor{gray!40}$44.2$&\cellcolor{gray!40}$47.4$&\cellcolor{gray!40}$51.7$\\
            & & \cellcolor{gray!10}4 &\cellcolor{gray!10}$51.4$&\cellcolor{gray!10}$51.9$&\cellcolor{gray!10}$42.7$&\cellcolor{gray!10}$53.6$&\cellcolor{gray!10}$51.9$\\
                \midrule
                \midrule
        \multirow{2}{*}{\model{\Ours}{6.7B} \lang{en}-only} & \multirow{2}{*}{same language}
            &\cellcolor{gray!40}0&\cellcolor{gray!40}-&\cellcolor{gray!40}$\textbf{77.8}$&\cellcolor{gray!40}$\textbf{56.4}$&\cellcolor{gray!40}$\textbf{100.0}$&\cellcolor{gray!40}$\textbf{63.0}$\\
            & &\cellcolor{gray!10} 4 &\cellcolor{gray!10}-&\cellcolor{gray!10}$51.0$&\cellcolor{gray!10}$40.7$&\cellcolor{gray!10}$50.6$&\cellcolor{gray!10}$51.1$\\
            \midrule
        \multirow{2}{*}{\model{\Ours}{7.5B}} & \multirow{2}{*}{same language} & 
            \cellcolor{gray!40}0 &\cellcolor{gray!40}-&\cellcolor{gray!40}$54.7$&\cellcolor{gray!40}$32.3$&\cellcolor{gray!40}$0.0$&\cellcolor{gray!40}$52.0$\\
            & &\cellcolor{gray!10} 4 &\cellcolor{gray!10}-&\cellcolor{gray!10}$\textbf{54.5}$&\cellcolor{gray!10}$43.8$&\cellcolor{gray!10}$51.0$&\cellcolor{gray!10}$\textbf{53.2}$\\
                \midrule
                \midrule
        \multirow{2}{*}{\model{\Ours}{6.7B} \lang{en}-only} & \multirow{2}{*}{English} & 
            \cellcolor{gray!40}0 &\cellcolor{gray!40}-&\cellcolor{gray!40}$53.3$&\cellcolor{gray!40}$48.5$&\cellcolor{gray!40}$60.3$&\cellcolor{gray!40}$53.4$\\
             & & \cellcolor{gray!10}4 &\cellcolor{gray!10}-&\cellcolor{gray!10}$51.8$&\cellcolor{gray!10}$42.5$&\cellcolor{gray!10}$52.7$&\cellcolor{gray!10}$52.0$\\
         \bottomrule
    \end{tabular}
    }
    \caption{Precision of our multilingual model and other English models on the Hate Speech Detection task. We evaluate five target languages. For each target language, we bold the highest number for zero-shot and four-shot respectively. For language condition, we consider three cases: ``code switched'' means the prefix, candidates are in English and tweets are in the target language, ``same language'' means prefix, candidates and tweets are in the target language, ``English'' means prefix, candidates and tweets are in English.
    }
        \label{tab:hate-speech-performance-precision}
\end{table*}
}

\hide{
\section{Full Evaluation Results}
Table~\ref{tab:multi_few_shot_by_resource_level} shows the evaluation results of \model{\Oursns}{7.5B} on all five multilingual tasks for each language.

\begin{table*}[t]
\centering
\resizebox{\textwidth}{!}{%
    \begin{tabular}{ll|rrrr|rrrrr|rrr|rrrrr|rrrr}
\toprule
      & {} & \multicolumn{21}{l}{accuracy::mean} \\
      & task & \multicolumn{4}{l}{storycloze} & \multicolumn{5}{l}{xcopa} & \multicolumn{3}{l}{xwinograd} & \multicolumn{5}{l}{xnli} & \multicolumn{4}{l}{pawsx} \\
      & \#\ shot &            0   &  1   &  4   &  32  &     0   &  1   &  4   &  32  &  128 &         0   &  1   &  4   &   0   &  1   &  4   &  32  &  128 &   0   &  1   &  4   &  32  \\
& lang &                &      &      &      &         &      &      &      &      &             &      &      &       &      &      &      &      &       &      &      &      \\
\midrule
  hi & en &           75.0 & 74.8 & 75.9 & 76.4 &         &      &      &      &      &        62.4 & 64.6 & 66.0 &  55.3 & 49.6 & 52.6 & 54.9 & 55.0 &  59.1 & 53.8 & 55.5 & 54.7 \\
      & zh &           66.6 & 67.3 & 67.7 & 68.6 &    62.4 & 64.4 & 67.2 & 68.2 & 67.4 &             &      &      &  44.8 & 40.6 & 48.8 & 50.4 & 50.7 &  50.6 & 53.1 & 52.9 & 52.9 \\
      & ru &           71.0 & 71.4 & 72.4 & 72.8 &         &      &      &      &      &        58.4 & 58.9 & 60.6 &  48.4 & 42.1 & 48.6 & 39.3 & 37.4 &       &      &      &      \\
      & es &           68.1 & 68.0 & 69.2 & 69.1 &         &      &      &      &      &             &      &      &  39.1 & 48.0 & 45.8 & 46.6 & 45.4 &  58.1 & 53.9 & 54.6 & 54.7 \\
      & ja &                &      &      &      &         &      &      &      &      &        57.5 & 57.9 & 58.7 &       &      &      &      &      &  50.0 & 53.8 & 53.8 & 54.0 \\
      & de &                &      &      &      &         &      &      &      &      &             &      &      &  42.3 & 39.4 & 45.6 & 45.9 & 43.8 &  58.0 & 53.5 & 54.1 & 54.6 \\
      & fr &                &      &      &      &         &      &      &      &      &             &      &      &  50.8 & 45.1 & 49.4 & 47.2 & 44.8 &  54.7 & 53.3 & 54.8 & 54.5 \\
      \hdashline
      & Avg. &           70.2 & 70.4 & 71.3 & 71.7 &    62.4 & 64.4 & 67.2 & 68.2 & 67.4 &        59.4 & 60.4 & 61.8 &  46.8 & 44.1 & 48.5 & 47.4 & 46.2 &  55.1 & 53.6 & 54.3 & 54.2 \\
      \midrule
  med & it &                &      &      &      &    60.8 & 66.2 & 69.2 & 70.2 & 70.1 &             &      &      &       &      &      &      &      &       &      &      &      \\
      & pt &                &      &      &      &         &      &      &      &      &        61.2 & 61.2 & 61.6 &       &      &      &      &      &       &      &      &      \\
      & el &                &      &      &      &         &      &      &      &      &             &      &      &  46.4 & 44.8 & 48.7 & 48.0 & 47.3 &       &      &      &      \\
      & id &           70.1 & 69.7 & 70.8 & 71.0 &    66.6 & 67.2 & 68.9 & 70.0 & 70.4 &             &      &      &       &      &      &      &      &       &      &      &      \\
      & th &                &      &      &      &    56.8 & 57.4 & 62.0 & 63.9 & 63.5 &             &      &      &  46.8 & 42.2 & 46.6 & 47.7 & 46.0 &       &      &      &      \\
      & vi &                &      &      &      &    61.4 & 62.6 & 65.6 & 68.3 & 68.0 &             &      &      &  47.6 & 44.6 & 48.5 & 48.7 & 47.2 &       &      &      &      \\
      & ko &                &      &      &      &         &      &      &      &      &             &      &      &       &      &      &      &      &  51.5 & 53.2 & 53.2 & 53.4 \\
      & tr &                &      &      &      &    56.8 & 57.3 & 58.5 & 59.5 & 61.4 &             &      &      &  45.5 & 42.1 & 45.4 & 45.7 & 43.3 &       &      &      &      \\
      & ar &           58.3 & 58.9 & 59.8 & 59.8 &         &      &      &      &      &             &      &      &  48.1 & 41.7 & 46.4 & 43.9 & 42.1 &       &      &      &      \\
      & bg &                &      &      &      &         &      &      &      &      &             &      &      &  49.1 & 42.6 & 48.9 & 42.9 & 41.2 &       &      &      &      \\
      \hdashline
      & Avg. &           64.2 & 64.3 & 65.3 & 65.4 &    60.5 & 62.1 & 64.8 & 66.4 & 66.7 &        61.2 & 61.2 & 61.6 &  47.3 & 43.0 & 47.4 & 46.2 & 44.5 &  51.5 & 53.2 & 53.2 & 53.4 \\
      \midrule
     lo & hi &           60.9 & 62.0 & 62.5 & 63.2 &         &      &      &      &      &             &      &      &  43.4 & 37.3 & 46.8 & 46.6 & 46.9 &       &      &      &      \\
      & et &                &      &      &      &    61.6 & 63.7 & 65.9 & 66.2 & 65.8 &             &      &      &       &      &      &      &      &       &      &      &      \\
      & ur &                &      &      &      &         &      &      &      &      &             &      &      &  41.9 & 36.6 & 43.4 & 46.2 & 46.6 &       &      &      &      \\
      & ta &                &      &      &      &    56.2 & 56.4 & 56.3 & 57.6 & 56.8 &             &      &      &       &      &      &      &      &       &      &      &      \\
      & sw &           65.0 & 64.5 & 65.2 & 65.8 &    57.6 & 59.4 & 62.9 & 64.2 & 63.2 &             &      &      &  45.5 & 42.8 & 44.5 & 46.3 & 45.8 &       &      &      &      \\
      & te &           61.7 & 62.8 & 63.4 & 64.4 &         &      &      &      &      &             &      &      &       &      &      &      &      &       &      &      &      \\
      \hdashline
      & Avg. &           62.5 & 63.1 & 63.7 & 64.5 &    58.5 & 59.8 & 61.7 & 62.7 & 61.9 &             &      &      &  43.6 & 38.9 & 44.9 & 46.4 & 46.4 &       &      &      &      \\
      \midrule
ex-lo & ht &                &      &      &      &    57.0 & 57.4 & 58.9 & 61.0 & 60.1 &             &      &      &       &      &      &      &      &       &      &      &      \\
      & eu &           62.3 & 62.6 & 63.8 & 63.5 &         &      &      &      &      &             &      &      &       &      &      &      &      &       &      &      &      \\
      & my &           60.7 & 60.6 & 61.2 & 61.1 &         &      &      &      &      &             &      &      &       &      &      &      &      &       &      &      &      \\
      & qu &                &      &      &      &    47.4 & 48.3 & 47.1 & 48.0 & 48.2 &             &      &      &       &      &      &      &      &       &      &      &      \\
      \hdashline
      & Avg. &           61.5 & 61.6 & 62.5 & 62.3 &    52.2 & 52.8 & 53.0 & 54.5 & 54.1 &             &      &      &       &      &      &      &      &       &      &      &      \\
\bottomrule
\end{tabular}
}
\caption{In-language in-context few-shot learning performance of \model{\Oursns}{7.5B} on multilingual downstream tasks grouped by resource level. We sort the languages in this table by the \# tokens in our pre-training corpora.} % \vic{Fix resource-level multi-row issue. Change 128 to K. Info to add to this table: (3) Average score and standard deviation over all languages for each task. (4) Split each task to its own table or move to Appendix.}}
\label{tab:multi_few_shot_by_resource_level}
\end{table*}
}

%%%%%%%%%%%%%%% TO FILL POST INTERNAL SUBMISSION %%%%%%%%%%%%%%%%
\hide{
\section{Qualitative Analysis of Multilingual Prompts}
\label{sec:qualitative_prompt_evaluation}

\begin{table*}[t]
    \centering
    \scalebox{0.78}{
        \begin{tabular}{llrrrrrrrrrrr}
        \toprule
             & {} & \multicolumn{11}{l}{accuracy::mean} \\
             & pivoting\_resource\_level & \multicolumn{5}{l}{high} & \multicolumn{2}{l}{medium} & \multicolumn{3}{l}{low} & extremely-low \\
             & language &             ar &   en &   es &   ru &   zh &       hi &   id &   eu &   sw &   te &            my \\
        pivoting\_model\_size & nb\_few\_shot\_samples &                &      &      &      &      &          &      &      &      &      &               \\
        \midrule
         564M & 4  &           52.2 & 63.2 & 57.6 & 59.6 & 56.4 &     52.9 & 57.5 & 55.4 & 57.3 & 57.6 &          54.0 \\
             & 32 &           51.9 & 63.5 & 57.3 & 60.3 & 56.4 &     53.4 & 57.5 & 55.3 & 57.5 & 57.2 &          53.6 \\
        1.7B & 0  &           56.1 & 67.6 & 61.4 & 64.1 & 60.2 &     56.3 & 63.3 & 56.8 & 61.0 & 58.9 &          56.4 \\
             & 1  &           55.4 & 67.6 & 61.0 & 63.8 & 60.4 &     57.5 & 63.3 & 56.9 & 59.8 & 59.3 &          56.1 \\
             & 4  &           55.4 & 67.8 & 61.6 & 64.3 & 60.8 &     57.8 & 63.7 & 56.7 & 59.4 & 60.1 &          56.5 \\
             & 32 &           55.3 & 68.6 & 61.9 & 64.7 & 62.6 &     58.1 & 64.1 & 56.8 & 60.1 & 60.6 &          56.4 \\
        7.5B & 0  &           58.3 & 75.0 & 68.1 & 71.0 & 66.6 &     60.9 & 70.1 & 62.3 & 65.0 & 61.7 &          60.7 \\
             & 1  &           58.9 & 74.8 & 68.0 & 71.4 & 67.3 &     62.0 & 69.7 & 62.6 & 64.5 & 62.8 &          60.6 \\
             & 4  &           59.8 & 75.9 & 69.2 & 72.4 & 67.7 &     62.5 & 70.8 & 63.8 & 65.2 & 63.4 &          61.2 \\
             & 32 &           59.8 & 76.4 & 69.1 & 72.8 & 68.6 &     63.2 & 71.0 & 63.5 & 65.8 & 64.4 &          61.1 \\
        \bottomrule
        \end{tabular}
    }
    \caption{StoryCloze}
\end{table*}

\begin{table*}[t]
    \centering
    \scalebox{0.78}{
        \begin{tabular}{llrrrrrrrrrrr}
        \toprule
             & {} & \multicolumn{11}{l}{accuracy::mean} \\
             & pivoting\_resource\_level & \multicolumn{2}{l}{high} & \multicolumn{4}{l}{medium} & \multicolumn{3}{l}{low} & \multicolumn{2}{l}{extremely-low} \\
             & language &             it &   zh &       id &   th &   tr &   vi &   et &   sw &   ta &            ht &   qu \\
        pivoting\_model\_size & nb\_few\_shot\_samples &                &      &          &      &      &      &      &      &      &               &      \\
        \midrule
         564M & 0   &           50.4 & 55.2 &     58.2 & 52.4 & 50.8 & 53.0 & 51.4 & 51.4 & 49.6 &          53.4 & 50.2 \\
             & 4   &           51.6 & 54.2 &     59.2 & 52.7 & 50.2 & 50.8 & 52.5 & 52.4 & 50.0 &          53.4 & 49.3 \\
             & 32  &           52.9 & 53.7 &     58.5 & 52.0 & 50.5 & 51.4 & 50.2 & 51.8 & 51.6 &          53.0 & 49.9 \\
        1.7B & 0   &           56.4 & 55.4 &     63.2 & 55.8 & 53.2 & 59.2 & 57.8 & 54.8 & 53.0 &          57.0 & 48.2 \\
             & 4   &           58.3 & 59.2 &     63.4 & 55.9 & 53.5 & 58.3 & 58.5 & 57.0 & 51.3 &          56.6 & 47.9 \\
             & 32  &           59.6 & 59.6 &     63.8 & 55.7 & 53.7 & 60.5 & 59.5 & 56.6 & 50.1 &          53.8 & 47.9 \\
             & 128 &           60.2 & 59.8 &     63.1 & 56.0 & 54.2 & 60.8 & 59.8 & 56.9 & 51.0 &          53.7 & 47.2 \\
        7.5B & 0   &           60.8 & 62.4 &     66.6 & 56.8 & 56.8 & 61.4 & 61.6 & 57.6 & 56.2 &          57.0 & 47.4 \\
             & 1   &           66.2 & 64.4 &     67.2 & 57.4 & 57.3 & 62.6 & 63.7 & 59.4 & 56.4 &          57.4 & 48.3 \\
             & 4   &           69.2 & 67.2 &     68.9 & 62.0 & 58.5 & 65.6 & 65.9 & 62.9 & 56.3 &          58.9 & 47.1 \\
             & 32  &           70.2 & 68.2 &     70.0 & 63.9 & 59.5 & 68.3 & 66.2 & 64.2 & 57.6 &          61.0 & 48.0 \\
             & 128 &           70.1 & 67.4 &     70.4 & 63.5 & 61.4 & 68.0 & 65.8 & 63.2 & 56.8 &          60.1 & 48.2 \\
        \bottomrule
        \end{tabular}
    }
    \caption{XCOPA}
\end{table*}

\begin{table*}[t]
    \centering
    \scalebox{0.78}{
        \begin{tabular}{llrrrr}
        \toprule
             & {} & \multicolumn{4}{l}{accuracy::mean} \\
             & pivoting\_resource\_level & \multicolumn{3}{l}{high} &   medium \\
             & language &             en &   pt &   ru &       jp \\
        pivoting\_model\_size & nb\_few\_shot\_samples &                &      &      &          \\
        \midrule
         564M & 0   &           55.0 & 54.4 & 53.3 &     51.5 \\
             & 4   &           55.2 & 53.1 & 54.8 &     51.5 \\
             & 32  &           57.0 & 53.8 & 56.5 &     52.1 \\
        1.7B & 0   &           57.2 & 54.8 & 54.0 &     52.8 \\
             & 4   &           59.6 & 57.0 & 54.5 &     54.2 \\
             & 32  &           62.7 & 56.9 & 57.8 &     53.9 \\
             & 128 &           62.1 & 55.9 & 58.9 &     54.9 \\
        7.5B & 0   &           62.4 & 61.2 & 58.4 &     57.5 \\
             & 1   &           64.6 & 61.2 & 58.9 &     57.9 \\
             & 4   &           66.0 & 61.6 & 60.6 &     58.7 \\
             & 32  &           73.2 & 63.7 & 62.6 &     58.7 \\
             & 128 &           71.7 & 61.3 & 62.7 &     58.7 \\
        \bottomrule
        \end{tabular}
    }
    \caption{XWinograd}
\end{table*}

\begin{table*}[t]
    \centering
    \scalebox{0.78}{
        \begin{tabular}{ll|lrrrrrl|rrllll|ll}
        \toprule
             & & \multicolumn{7}{l}{high} & \multicolumn{6}{l}{medium} & \multicolumn{2}{l}{low} \\
             model\_size & \#\_shot &   ar &   de &   en &   es &   fr &   ru &   zh &       bg &   el &   hi &   th &   tr &   vi &   sw &   ur \\
        \midrule
         564M & 0   &           38.6 & 36.9 & 48.1 & 33.9 & 39.7 & 37.1 & 36.7 &     38.8 & 43.0 & 40.5 & 40.6 & 40.2 & 40.9 & 39.9 & 37.1 \\
             & 4   &           40.1 & 42.2 & 48.2 & 43.9 & 43.9 & 38.0 & 39.4 &     41.6 & 42.6 & 38.8 & 40.8 & 38.4 & 42.3 & 39.7 & 37.8 \\
             & 32  &                & 38.1 & 45.1 & 41.9 & 37.8 & 34.9 &      &     35.8 & 40.2 &      &      & 36.0 &      &      &      \\
        \midrule
        1.7B & 0   &           46.1 & 39.2 & 50.1 & 35.3 & 45.7 & 42.9 & 44.0 &     46.5 & 46.0 & 44.7 & 43.6 & 41.3 & 40.7 & 41.5 & 42.6 \\
             & 1   &           42.8 & 39.7 & 49.8 & 46.5 & 46.0 & 44.2 & 40.8 &     43.9 & 44.7 & 37.9 & 42.2 & 40.5 & 45.4 & 42.2 & 39.2 \\
             & 4   &           43.1 & 44.6 & 51.1 & 44.5 & 46.9 & 42.8 & 43.3 &     44.9 & 46.0 & 43.1 & 44.0 & 42.1 & 45.5 & 41.7 & 40.4 \\
             & 32  &                & 39.9 & 50.3 & 41.5 & 43.2 & 35.4 &      &     38.3 & 42.6 &      &      & 41.3 &      &      &      \\
             & 128 &                & 38.7 & 50.4 & 39.7 & 41.9 & 34.5 &      &     36.9 & 42.6 &      &      &      &      &      &      \\
        \midrule
        7.5B & 0   &           48.1 & 42.3 & 55.3 & 39.1 & 50.8 & 48.4 & 44.8 &     49.1 & 46.4 & 43.4 & 46.8 & 45.5 & 47.6 & 45.5 & 41.9 \\
             & 1   &           41.7 & 39.4 & 49.6 & 48.0 & 45.1 & 42.1 & 40.6 &     42.6 & 44.8 & 37.3 & 42.2 & 42.1 & 44.6 & 42.8 & 36.6 \\
             & 4   &           46.4 & 45.6 & 52.6 & 45.8 & 49.4 & 48.6 & 48.8 &     48.9 & 48.7 & 46.8 & 46.6 & 45.4 & 48.5 & 44.5 & 43.4 \\
             & 32  &           43.9 & 45.9 & 54.9 & 46.6 & 47.2 & 39.3 & 50.4 &     42.9 & 48.0 & 46.6 & 47.7 & 45.7 & 48.7 & 46.3 & 46.2 \\
             & 128 &           42.1 & 43.8 & 55.0 & 45.4 & 44.8 & 37.4 & 50.4 &     41.2 & 47.3 & 46.9 & 45.2 & 43.3 & 47.2 & 45.8 & 46.6 \\
        \bottomrule
        \end{tabular}
    }
    \caption{XNLI}
\end{table*}

\begin{table*}[t]
    \centering
    \scalebox{0.78}{
        \begin{tabular}{llrrrrrrr}
        \toprule
             & {} & \multicolumn{7}{l}{accuracy::mean} \\
             & pivoting\_resource\_level & \multicolumn{5}{l}{high} & \multicolumn{2}{l}{medium} \\
             & language &             de &   en &   es &   fr &   zh &       ja &   ko \\
        pivoting\_model\_size & nb\_few\_shot\_samples &                &      &      &      &      &          &      \\
        \midrule
         564M & 0   &           56.4 & 51.2 & 53.5 & 53.0 & 54.4 &     53.3 & 53.4 \\
             & 4   &           51.5 & 51.1 & 52.4 & 51.8 & 52.4 &     51.8 & 49.1 \\
             & 32  &           51.7 & 53.6 & 54.1 & 50.8 & 53.8 &     54.0 & 51.0 \\
        1.7B & 0   &           53.5 & 52.2 & 54.4 & 51.6 & 53.2 &     54.4 & 52.5 \\
             & 4   &           54.1 & 54.4 & 54.6 & 53.4 & 54.8 &     54.3 & 52.1 \\
             & 32  &           53.6 & 54.0 & 54.4 & 53.0 & 54.0 &     55.2 & 53.6 \\
             & 128 &           53.4 & 53.4 & 54.4 & 52.7 & 53.5 &     54.8 & 53.3 \\
        7.5B & 0   &           58.0 & 59.1 & 58.1 & 54.7 & 50.6 &     50.0 & 51.5 \\
             & 1   &           53.5 & 53.8 & 53.9 & 53.3 & 53.1 &     53.8 & 53.2 \\
             & 4   &           54.1 & 55.5 & 54.6 & 54.8 & 52.9 &     53.8 & 53.2 \\
             & 32  &           54.6 & 54.7 & 54.7 & 54.5 & 52.9 &     54.0 & 53.4 \\
             & 128 &           54.6 & 54.8 & 54.7 & 54.5 & 53.1 &     53.9 & 54.7 \\
        \bottomrule
        \end{tabular}
    }
    \caption{Paws-X}
\end{table*}
}

\hide{
\begin{table*}[t]
\small
\setlength{\tabcolsep}{0.1em}
\renewcommand{\arraystretch}{1.5}

\begin{center}
\resizebox{\textwidth}{!}{%
    \begin{tabular}{lc|llll|lllll|lllll}
\toprule
      & task & \multicolumn{4}{l}{storycloze} & \multicolumn{5}{l}{xcopa} & \multicolumn{5}{l}{xnli} \\
      & shots &                         0   &                         1   &                         4   &                         32  &                                     0   &                                     1   &                                     4   &                                     32  &                                     Max &                         0   &                         1   &                        4   &                                     32  &                                     Max \\
\midrule
hi & zh &   $72.9_{  \mathit{ 6.2} }$ &   $72.7_{  \mathit{ 5.4} }$ &   $73.1_{  \mathit{ 5.3} }$ &   $72.9_{  \mathit{ 4.3} }$ &              $75.0_{  \mathit{ 12.6} }$ &              $77.0_{  \mathit{ 12.6} }$ &              $78.5_{  \mathit{ 11.2} }$ &              $80.1_{  \mathit{ 11.9} }$ &              $80.8_{  \mathit{ 13.5} }$ &   $52.6_{  \mathit{ 7.8} }$ &   $48.7_{  \mathit{ 8.1} }$ &  $51.1_{  \mathit{ 2.3} }$ &  $\mathbf{ 49.3_{  \mathit{ -1.1 } } }$ &  $\mathbf{ 48.8_{  \mathit{ -1.9 } } }$ \\
      & ru &   $75.4_{  \mathit{ 4.5} }$ &   $74.8_{  \mathit{ 3.5} }$ &   $75.3_{  \mathit{ 2.9} }$ &   $74.7_{  \mathit{ 2.0} }$ &                                         &                                         &                                         &                                         &                                         &   $52.0_{  \mathit{ 3.6} }$ &   $48.9_{  \mathit{ 6.8} }$ &  $53.2_{  \mathit{ 4.6} }$ &              $52.7_{  \mathit{ 13.4} }$ &              $52.7_{  \mathit{ 15.2} }$ \\
      & es &   $75.6_{  \mathit{ 7.4} }$ &   $75.1_{  \mathit{ 7.1} }$ &   $75.0_{  \mathit{ 5.8} }$ &   $76.3_{  \mathit{ 7.1} }$ &                                         &                                         &                                         &                                         &                                         &  $54.5_{  \mathit{ 15.4} }$ &   $50.7_{  \mathit{ 2.7} }$ &  $49.2_{  \mathit{ 3.4} }$ &  $\mathbf{ 46.5_{  \mathit{ -0.2 } } }$ &               $46.5_{  \mathit{ 1.0} }$ \\
      & de &                             &                             &                             &                             &                                         &                                         &                                         &                                         &                                         &  $53.7_{  \mathit{ 11.4} }$ &  $50.5_{  \mathit{ 11.1} }$ &  $52.4_{  \mathit{ 6.8} }$ &               $54.8_{  \mathit{ 8.9} }$ &              $54.8_{  \mathit{ 11.0} }$ \\
      & fr &                             &                             &                             &                             &                                         &                                         &                                         &                                         &                                         &   $53.9_{  \mathit{ 3.1} }$ &   $49.5_{  \mathit{ 4.4} }$ &  $50.3_{  \mathit{ 0.9} }$ &               $49.4_{  \mathit{ 2.2} }$ &               $50.1_{  \mathit{ 5.3} }$ \\
      
\hline
med & it &                             &                             &                             &                             &              $76.0_{  \mathit{ 15.2} }$ &              $77.3_{  \mathit{ 11.1} }$ &              $80.6_{  \mathit{ 11.4} }$ &              $80.4_{  \mathit{ 10.3} }$ &              $80.9_{  \mathit{ 10.8} }$ &                             &                             &                            &                                         &                                         \\
      & el &                             &                             &                             &                             &                                         &                                         &                                         &                                         &                                         &   $53.5_{  \mathit{ 7.1} }$ &   $50.8_{  \mathit{ 6.0} }$ &  $53.0_{  \mathit{ 4.3} }$ &               $52.1_{  \mathit{ 4.1} }$ &               $52.4_{  \mathit{ 5.1} }$ \\
      & id &   $71.2_{  \mathit{ 1.1} }$ &   $71.6_{  \mathit{ 2.0} }$ &   $72.0_{  \mathit{ 1.2} }$ &   $72.9_{  \mathit{ 1.8} }$ &               $73.2_{  \mathit{ 6.6} }$ &               $73.3_{  \mathit{ 6.2} }$ &               $75.8_{  \mathit{ 6.8} }$ &               $77.2_{  \mathit{ 7.1} }$ &               $77.3_{  \mathit{ 6.9} }$ &                             &                             &                            &                                         &                                         \\
      & th &                             &                             &                             &                             &  $\mathbf{ 53.8_{  \mathit{ -3.0 } } }$ &  $\mathbf{ 53.9_{  \mathit{ -3.6 } } }$ &  $\mathbf{ 57.7_{  \mathit{ -4.3 } } }$ &  $\mathbf{ 60.5_{  \mathit{ -3.4 } } }$ &  $\mathbf{ 60.4_{  \mathit{ -3.1 } } }$ &   $50.6_{  \mathit{ 3.8} }$ &   $46.8_{  \mathit{ 4.5} }$ &  $48.2_{  \mathit{ 1.6} }$ &  $\mathbf{ 46.7_{  \mathit{ -1.0 } } }$ &               $46.5_{  \mathit{ 0.5} }$ \\
      & vi &                             &                             &                             &                             &              $72.2_{  \mathit{ 10.8} }$ &              $73.9_{  \mathit{ 11.2} }$ &              $76.0_{  \mathit{ 10.5} }$ &               $78.0_{  \mathit{ 9.7} }$ &              $78.1_{  \mathit{ 10.1} }$ &   $52.6_{  \mathit{ 5.0} }$ &   $45.0_{  \mathit{ 0.3} }$ &  $52.8_{  \mathit{ 4.3} }$ &    $53.4_{  \mathit{ 4.7} }$ &               $53.3_{  \mathit{ 6.1} }$ \\
      & tr &                             &                             &                             &                             &              $72.4_{  \mathit{ 15.6} }$ &              $72.2_{  \mathit{ 14.9} }$ &              $73.7_{  \mathit{ 15.2} }$ &              $74.0_{  \mathit{ 14.5} }$ &              $74.2_{  \mathit{ 12.8} }$ &   $53.3_{  \mathit{ 7.8} }$ &   $47.8_{  \mathit{ 5.7} }$ &  $51.8_{  \mathit{ 6.4} }$ &               $52.7_{  \mathit{ 6.9} }$ &               $53.1_{  \mathit{ 9.9} }$ \\
      & ar &  $71.5_{  \mathit{ 13.2} }$ &  $71.6_{  \mathit{ 12.8} }$ &  $71.8_{  \mathit{ 12.0} }$ &  $71.4_{  \mathit{ 11.6} }$ &                                         &                                         &                                         &                                         &                                         &   $52.0_{  \mathit{ 3.9} }$ &   $49.1_{  \mathit{ 7.4} }$ &  $50.5_{  \mathit{ 4.1} }$ &               $50.1_{  \mathit{ 6.2} }$ &               $50.3_{  \mathit{ 8.2} }$ \\
      & bg &                             &                             &                             &                             &                                         &                                         &                                         &                                         &                                         &   $53.4_{  \mathit{ 4.3} }$ &   $50.0_{  \mathit{ 7.4} }$ &  $53.7_{  \mathit{ 4.8} }$ &              $53.3_{  \mathit{ 10.5} }$ &              $53.0_{  \mathit{ 11.8} }$ \\
      
\hline
lo & hi &   $70.5_{  \mathit{ 9.6} }$ &   $70.6_{  \mathit{ 8.6} }$ &   $71.6_{  \mathit{ 9.1} }$ &   $71.4_{  \mathit{ 8.2} }$ &                                         &                                         &                                         &                                         &                                         &   $50.7_{  \mathit{ 7.3} }$ &  $48.3_{  \mathit{ 11.0} }$ &  $49.8_{  \mathit{ 3.0} }$ &               $47.7_{  \mathit{ 1.2} }$ &               $47.9_{  \mathit{ 1.0} }$ \\
      & et &                             &                             &                             &                             &              $72.4_{  \mathit{ 10.8} }$ &               $72.4_{  \mathit{ 8.7} }$ &               $73.6_{  \mathit{ 7.7} }$ &               $74.8_{  \mathit{ 8.6} }$ &               $75.6_{  \mathit{ 9.8} }$ &                             &                             &                            &                                         &                                         \\
      & ur &                             &                             &                             &                             &                                         &                                         &                                         &                                         &                                         &   $48.7_{  \mathit{ 6.8} }$ &   $46.1_{  \mathit{ 9.5} }$ &  $47.2_{  \mathit{ 3.8} }$ &  $\mathbf{ 45.9_{  \mathit{ -0.4 } } }$ &  $\mathbf{ 45.6_{  \mathit{ -1.0 } } }$ \\
      & ta &                             &                             &                             &                             &              $67.2_{  \mathit{ 11.0} }$ &              $68.0_{  \mathit{ 11.6} }$ &              $69.9_{  \mathit{ 13.6} }$ &              $71.6_{  \mathit{ 13.9} }$ &              $71.4_{  \mathit{ 14.6} }$ &                             &                             &                            &                                         &                                         \\
      & sw &   $70.0_{  \mathit{ 5.1} }$ &   $70.2_{  \mathit{ 5.7} }$ &   $71.0_{  \mathit{ 5.8} }$ &   $71.3_{  \mathit{ 5.5} }$ &               $63.8_{  \mathit{ 6.2} }$ &               $65.4_{  \mathit{ 6.1} }$ &               $67.2_{  \mathit{ 4.3} }$ &               $67.2_{  \mathit{ 3.0} }$ &               $67.7_{  \mathit{ 4.5} }$ &   $51.3_{  \mathit{ 5.9} }$ &   $46.1_{  \mathit{ 3.3} }$ &  $50.2_{  \mathit{ 5.6} }$ &               $50.8_{  \mathit{ 4.5} }$ &               $50.6_{  \mathit{ 4.9} }$ \\
      & te &   $66.9_{  \mathit{ 5.3} }$ &   $68.0_{  \mathit{ 5.2} }$ &   $68.4_{  \mathit{ 5.1} }$ &   $69.4_{  \mathit{ 5.1} }$ &                                         &                                         &                                         &                                         &                                         &                             &                             &                            &                                         &                                         \\
\hline
ex-lo & ht &                             &                             &                             &                             &               $65.0_{  \mathit{ 8.0} }$ &               $64.8_{  \mathit{ 7.5} }$ &               $67.0_{  \mathit{ 8.0} }$ &               $66.8_{  \mathit{ 5.8} }$ &               $67.4_{  \mathit{ 7.3} }$ &                             &                             &                            &                                         &                                         \\
      & eu &   $70.5_{  \mathit{ 8.2} }$ &   $70.6_{  \mathit{ 7.9} }$ &   $72.2_{  \mathit{ 8.4} }$ &   $72.4_{  \mathit{ 8.9} }$ &                                         &                                         &                                         &                                         &                                         &                             &                             &                            &                                         &                                         \\
      & my &  $72.7_{  \mathit{ 12.0} }$ &  $71.7_{  \mathit{ 11.0} }$ &  $72.0_{  \mathit{ 10.8} }$ &  $71.8_{  \mathit{ 10.7} }$ &                                         &                                         &                                         &                                         &                                         &                             &                             &                            &                                         &                                         \\
\bottomrule
\end{tabular}
}
\end{center}

\caption{Performance of the translate-test baseline on 3 downstream tasks. The small, italic numbers are the difference between this baseline and the multilingual model. Positive values mean the translate-test baseline is better, cells where the multilingual model outperforms the baseline are highlighted in bold.}
\label{tab:translate_test_full}
\end{table*}
}

\hide{
\begin{table*}[t!]
\centering
\scalebox{0.8}{
\begin{tabular}{l|p{0.4\textwidth}|p{0.4\textwidth}}
\toprule
 & Source prompt (instantiated) & Target prompt (instantiated) \\
 \midrule
\emph{Same-language} & \entext{The best thing that may be said of Podhoretz and Decter is that their biological clocks can't have many more minutes left on them\underline{, right? Yes,} Decter is old.} & \vitext{Vâng, tôi thậm chí không nghĩ về điều đó, nhưng tôi đã rất thất vọng, và, tôi lại nói chuyện với anh ta lần nữa\underline{, đúng không? Đúng,} tôi đã không nói chuyện với anh ta nữa.}\\
\midrule
\emph{Source-language} & \entext{The best thing that may be said of Podhoretz and Decter is that their biological clocks can't have many more minutes left on them\underline{, right? Yes,} Decter is old. } & \vitext{Vâng, tôi thậm chí không nghĩ về điều đó, nhưng tôi đã rất thất vọng, và, tôi lại nói chuyện với anh ta lần nữa}\underline{\entext{, right? Yes,}} \vitext{tôi đã không nói chuyện với anh ta nữa.} \\
\bottomrule
\end{tabular}
}
\caption{XNLI example prompts for cross-lingual transfer from \textit{\entext{English (en)}} to \textit{\vitext{Vietnamese (vi)}}, with the \textit{same-language} and \textit{source-language} settings. The \underline{underlined} text shows the template-specific part of the prompt.} 
\label{tab:cross_lingual_code_switching_example}
\end{table*}

\subsection{Cross-lingual transfer through templates.}
In Table \ref{tab:crosslingual_xnli} we show results of cross-lingual transfer between selected languages on XNLI.\footnote{Due to high computational cost, we evaluate on a subset of language pairs that allow evaluation of % high and low resource settings 
cross-lingual transfer from high-resource languages to low-resource languages, and between languages that are typologically similar.}
%\ves{I don't understand the next sentence} 
%For XNLI we use crafted templates for forming the in-context learning. 
While XCOPA and StoryCloze are % natural language modeling 
commensense reasoning tasks naturally framed as language modeling, XNLI requires framing the prompts using a template with task-specific fields and candidate verbalizers (Table \ref{tab:multilingual_prompts}). 
This allows us to also test the impact of code-switching between the language of the template and verbalizer, and the language of the examples as described at the end of Section \ref{sec:multilingual_prompts}. In particular, we examined two settings for each train-eval pair, including \textit{same-language}, where the prompt fields and the example are in the same language, and \textit{source-languauge} where the prompt fields for the both the demonstration and test examples are in the source language.
% \mona{Not understanding the previous paragraph. I suggest changing source language template to cross lingual template.} 
%\vic{Should we mention these two settings at the beginning of section 6.3? Then we can say we evaluated XNLI in two settings, XCOPA in XX setting, for XStoryCloze both settings are the same? In the current flow, it is unclear which setting was used for XCOPA.}
Similar to XCOPA, most cross-lingual settings work better than the zero-shot setting of the target language. Bulgarian is an exception as it does not seem to benefit from Russian although it is in the same language family, but the \textit{source-language} setting reduces the difference. Another language that does not work well in the cross-lingual settings for this task is Swahili (\textit{low} resource), for which we examined transfer from English (\textit{high} resource) and Arabic (\textit{medium} resource). In contrast, Thai (\textit{medium}) and Urdu (\textit{low} resource) work better in cross-lingual setting than the corresponding zero-shot evaluation. Surprisingly, the source-language template transfer works even better % when very different medium to high resource level languages like 
with English (to Thai and Urdu) and Turkish (to Urdu) as the source languages\footnote{There is inherent code switching in these languages (English presence in Thai and Urdu both lexical and morphological). Turkish and Arabic also have influence on Urdu. We hypothesize that these factors have positively impacted the cross-lingual in-context learning performance.}. %We currently don't have a good intuition about this and we leave understanding it better for future work.
% \mona{There is inherent code switching in these languages (English presence in Thai and Urdu both lexical and morphological) and Turkish and Arabic influence on Urdu. This would make for a beautiful follow on paper in the area of contact languages and their impact on LM performance.}

% \comment{Victoria to fill in}
\begin{table*}[t!]
\centering
\scalebox{0.78}{
\begin{tabular}{l|ccccc|ccc|c|ccc}
\toprule
  &   \multicolumn{9}{c|}{high}           &   \multicolumn{2}{c|}{medium} & low \\
 %\cmidrule{2-10}
 &   \multicolumn{8}{c}{en}  &       ru &       tr &    ar &   hi \\
 \midrule
  & \multicolumn{5}{c|}{medium} & \multicolumn{3}{c|}{low} &   medium &      \multicolumn{3}{c}{low} \\
 
 prompt & bg &    el &   th &    tr &    vi &   hi &    sw &   ur &       bg &       ur &    sw &   ur \\
 \midrule
 Same-lang & \bf 2.55 & \bf 0.98 & 2.16 &  \bf 1.27 & \bf 2.23 & \bf 2.51 & \bf -0.69 & 1.21 &    -2.49 &    -0.38 & \bf  -1.64 & \bf 3.31 \\
 \midrule
Source-lang  &  -4.59 & -2.44 & \bf 7.87 & -4.97 & -1.08 & 2.01 & -1.15 & \bf 7.42 &    \bf -1.43 &     \bf  6.67 & -5.86 & 2.31 \\

\bottomrule
\end{tabular}
}
\caption{Cross-lingual transfer for XNLI. The results are improvement over \textit{same-lang} zero-shot performance for the evaluated language. The first language and resource levels group refer to the source language and the second refer to the target language. The results rows show the improvement over the zero-shot results of the target language evaluation set. \textit{same lang} refers to a setting there the template is in the example language and \textit{source lang} refers to a setting where the template is only in the source language (see Table \ref{tab:cross_lingual_code_switching_example} for examples). 
}
\label{tab:crosslingual_xnli}
\end{table*}

}

\subsection{Knowledge Probing}
% \comment{Xian to fill in}

\begin{figure*}[t!]
    \centering
    \includegraphics[width=.96\linewidth]{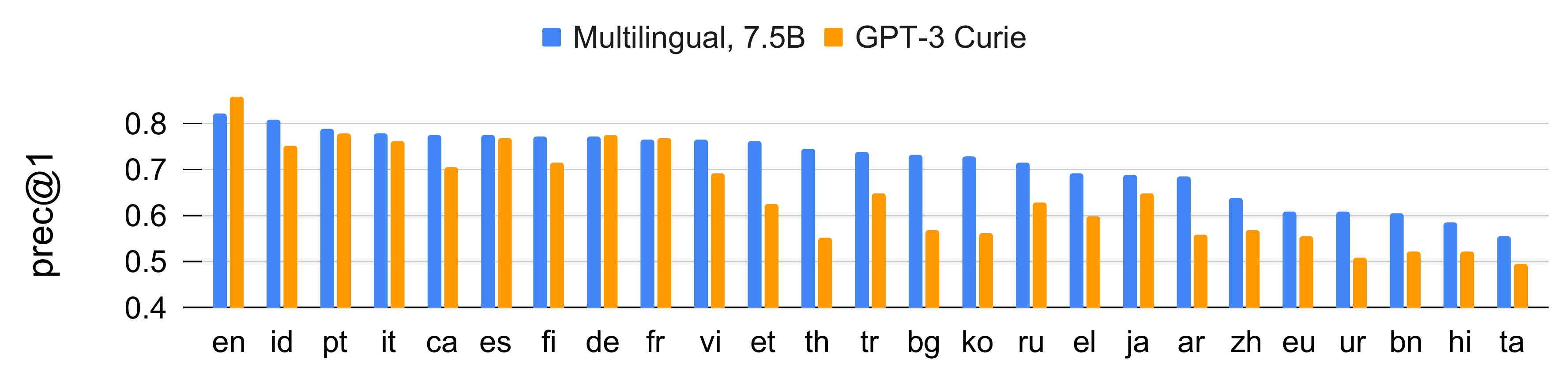}
    \caption{Knowledge probing on 25 languages. The performance of a random baseline is 0.33 since we down-sampled the candidate set of each query to contain three candidates. }%\comment{TODO: It's interesting to show both the seen language plot and unseen language plot.}}
    \label{fig:mlama_seen}
\end{figure*}

% \begin{figure*}[t!]
%     \centering
%     \includegraphics[width=\linewidth]{figures/mlama_unseen.pdf}
%     \caption{Knowledge probing on unseen languages.}
%     \label{fig:mlama_unseen}
% \end{figure*}

We evaluate to what extent our multilingual language model can effectively store factual knowledge in different languages. To this end, we evaluate knowledge triplet completion using the mLAMA dataset \citep{DBLP:conf/eacl/KassnerDS21}, which was translated from the English benchmark LAMA \cite{petroni2019language} using Google Translate. The data is from TREx \cite{elsahar2018t} with triples of the format $\langle$object, relation, subject$\rangle$. Following the convention of LAMA, triples are converted to templates for querying the language model. For example, a triple like $\langle$Paris, capital-of, France$\rangle$ is converted to template ``Paris is the capital of \texttt{[MASK]}". % With the typed querying proposed in mLAMA, evaluating using the entire candidate set is very time consuming with causal language models. Therefore, we 
While each query in the original mLAMA dataset contains hundreds of candidates on average, we restrict it to three candidates one of which is the ground truth candidate and the other two candidates are randomly sampled to ensure fast inference and save API cost. Following the evaluation protocol of mLAMA, we report precision @1 averaged over all relations per language.    

% First, w
We evaluate on the 25 languages % seen during pretraining. 
covered in \Ours's pre-training data. We compare to the \model{GPT-3}{6.7B} model. As shown in Figure \ref{fig:mlama_seen}, both our multilingual model and GPT-3 Curie perform well on English. For non-English languages, our multilingual model maintains performance (above 0.6) while GPT-3 Curie % drops closer to the random baseline (0.3)
drops drastically  especially for medium and low resource languages. %\ves{next sentence is hard to read}
Overall, compared to an English-centric language model, our multilingual language model are better at retaining factual knowledge on a wider range of languages with $+7.1$ points on average. %\mona{This not a foregone conclusion. It is potentially a result of the small percentage of non english data in the curie model especially in those languages. We benefit from having more diverse and equitable language representation in the multilingual model as opposed to the english centric one}%, in contrast to a language-dependent knowledge base from a .  

% TODO: update unseen languages later
% \paragraph{Unseen languages.} Both our multilingual model and GPT-3 Curie perform way above a random baseline. % Multilingual model demonstrate strong performance on unseen languages as is shown in Figure \ref{fig:mlama_unseen}. 

\section{Safety and Bias Analysis}
\label{sec:safety_and_bias_analysis}
% \comment{Tianlu, Mona, Xian to condense the section and strengthen the takeaway messages. Shorten the length to be comparable to other result sections.}\xian{RAI is an important section, I don't think we need to worry about shortening the length. Some tables/plots could be moved to Appendix, though, e.g. we do not need to report all accuracy, precision and recall in the main body.}\vic{+1. Using the appendix is a good idea}

% While language models have demonstrated superior performance on various tasks, it is also 
Given the centrality of large scale Language models, it is important to ensure such powerful models are used responsibly. Accordingly, we further examine XGLM's behavior on two tasks: 
\begin{itemize}
    \item \textbf{Hate speech detection:} A safety task to test language models' ability to identify hateful and offensive text;
    \item \textbf{Occupation Identification:} A bias task to study language models' performance disparity between different gender groups on the task of occupation identification.
\end{itemize}
Through extensive experiments, we have following findings:  First, hate speech detection in an in-context learning setting is quite challenging. Moreover, language models are not effectively leveraging few-shot examples to improve the performance. Second, although language models have relatively good performance on the occupation identification task, they run the risk of exhibiting strong gender bias for certain occupations.
% e.g. \textit{model} and \textit{teacher}.
%%%%%%%%%%%%%%%%%%%% TODO post internal submission %%%%%%%%%%%%%%%%%%%%%
% \xian{I think we can improve on the main messages to be more concrete, especially the first finding.} \ves{agree. it is barely a discovery that hate speech detection is challenging}

\subsection{Hate Speech Detection}
\label{sec:hate-speech-detection}

\subsubsection{Setup}
\paragraph{Datasets.} We adopt datasets introduced by~\citet{huang2020multilingual} that include hate speech data from Twitter in five languages: English, Italian, Portuguese, Polish and Spanish. All hyperlinks, usernames and hashtags are replaced with generic symbols (\texttt{URL}, \texttt{USER}, \texttt{HASHTAG}) to anonymize user information.  We  remove tweets containing more than $2$ generic symbols to encourage more informative examples. We further filter out tweets of length less than $5$ tokens or more than $30$ tokens. In the spirit of creating balanced data, we randomly sample $500$ each positive (hate speech) negative (not hate speech) examples for each language. For further comparison, we translate non-English data into English by using Google Translate and then evaluate English models performance on the task.

\paragraph{Prompts.} We evaluate two approaches to prompting, similar to Section \ref{sec:crosslingual_capability}. For English prompts, we \textit{prefix} ``The sentence is <candidate>'' to the input sentence to form a prompt. We consider $10$ verbalization \textit{candidates} including $5$ negative (\textit{normal., common., ok., usual., acceptable.}) corresponding to classification of \textit{not hate speech} and $5$ positive (\textit{sexist., racist., offensive., abusive., hateful.}) representing classification of \textit{hate speech}. For code-switched prompt, we translate the English prefix and candidates into the corresponding target language by using Google Translate. For example, ``The sentence is normal'' is translated into ``Questa frase è normale.'' for Spanish. For few-shot learning, we randomly draw examples from the training data and report the average performance %with confidence intervals (standard deviation) 
across $5$ runs.
% \xian{Can we formulate the setting to be consistent with Section \ref{subsec:task_formulation}}. \comment{Victoria, any suggestions?} \vic{I think it is fine to use a slightly different set up in the analysis section, as long as it is clearly motivated. If we have more time, it would be better to make the setup consistent across the entire paper.}

% \noindent \textbf{Metrics.}
\paragraph{Metrics.}
%For each candidate, we form a prompt and obtain a score from the language model. We then find the candidate with the highest score and if it comes from the positive candidate set, we treat it as a positive (hate speech) prediction otherwise negative (not hate speech). 
We compute precision, recall and accuracy for all experimental conditions. % We expect a random model to achieve an accuracy of $50\%$ since we have balanced test data.
Since the test data is balanced, the accuracy of a random baseline is 50\%.

\subsubsection{Results}
We show accuracy and recall scores in Table~\ref{tab:hate-speech-performance}. % and precision scores are included in Appendix~\ref{sec:hate_speech_precision}.
%\xian{Should we also show and discuss the recall scores? Since a more realistic usage of models in hate speech detection is to flag potential hate speech and then send to humans for further judgement, recall seems more important/relevant.}. 
Bolded results are the highest in the table and those with an (*) are statistically significantly better than other comparable conditions. %We further conduct significant test and bold numbers that are significantly better then others. 
Hate speech detection is a challenging task for all models. We observe that across the five languages, in-context learning results are only slightly better than  random  ($50\%$).
The results are also unstable and sensitive to prompt changing. Overall, the \model{\Oursns}{7.5B} model has better recall compared to the English-centric models. For example, the \model{\Oursns}{6.7B} En-only model has very low recall score in the zero-shot setting with the language condition set as ``same language'', indicating that it blindly predicts almost everything as negative (not hate speech).
% More specifically, our 7.5B multilingual model and GPT-3 curie model achieve better performance on English, Italian and Spanish, whereas the 6.7B en model yields better performance on Portuguese and Polish.
% the best performance on English, Italian and Spanish are from the 6.7B curie \ves{GPT curie to be consisten} model and the best performance on Portuguese and Polish are from the 6.7B en model with test data translated into English.
% \xian{These results are using English template, right? I think we should also report results using in-language template like other tasks.}.
Another interesting observation is that most few-shot results are worse than zero-shot, which indicates that with the prefix described above, language models are not able to utilize examples.
% \xian{This is an different conclusion from observations on other tasks, where at least low-shot examples help.}.
Interestingly, we also find that in one-shot experiments models tend to copy the label of the given example instead of predicting based on the input tweet. This further demonstrates that language models are struggling with learning from few-shot examples in this task. 
%%%%%%%%%%%%%%%%%%%% TODO post internal submission %%%%%%%%%%%%%%%%%%%%%
% \mona{We need to add some form of speculation here, how is hate speech detection different from other tasks in the paper such as NLI, Paraphrasing, storycloze, xcopa, etc. It is a different type of task where there needs to be a cultural nuance beyond simple language patterns. The challenge of social value tasks.}

% The reason could be either we are not using a proper template so that models fail to learn or for hate speech detection, the task itself is too hard for language models to learn from few-shots.
\begin{table*}[t]
    \centering
    % \footnotesize
    % \setlength{\tabcolsep}{4pt}
    \scalebox{0.72}{
    \begin{tabular}{cccccccc}
        \toprule
         Model & language condition & \#\ shot & English& Italian & Portuguese & Polish & Spanish \\
         \midrule
        \multicolumn{8}{c}{Accuracy}\\
        \midrule
         \multirow{2}{*}{\xlmgen} & \multirow{2}{*}{code switched}&
            \cellcolor{gray!40}  0 & \cellcolor{gray!40}$54.5$& \cellcolor{gray!40}$54.3$& \cellcolor{gray!40}$57.0$& \cellcolor{gray!40}$51.3$& \cellcolor{gray!40}$52.5$\\
            & &\cellcolor{gray!10} 4 &\cellcolor{gray!10}$\textbf{55.5}^*$&\cellcolor{gray!10}$53.4$&\cellcolor{gray!10}$48.5$&\cellcolor{gray!10}$52.5$&\cellcolor{gray!10}$52.2$\\
                \midrule
         \multirow{2}{*}{\model{GPT-3}{6.7B}} & \multirow{2}{*}{code switched}& 
            \cellcolor{gray!40} 0 & \cellcolor{gray!40}$\textbf{59.2}$&\cellcolor{gray!40}$\textbf{61.8}$&\cellcolor{gray!40}$55.8$&\cellcolor{gray!40}$58.7$&\cellcolor{gray!40}$\textbf{55.6}$\\
            & & \cellcolor{gray!10}4 &\cellcolor{gray!10}$53.6$&\cellcolor{gray!10}$\textbf{54.8}$&\cellcolor{gray!10}$53.2$&\cellcolor{gray!10}$\textbf{54.5}$&\cellcolor{gray!10}$\textbf{53.7}$\\
                \midrule
         \multirow{2}{*}{\model{\Oursns}{7.5B}} & \multirow{2}{*}{code switched}& 
            \cellcolor{gray!40}0 &\cellcolor{gray!40}$56.0$&\cellcolor{gray!40}$60.4$&\cellcolor{gray!40}$50.8$&\cellcolor{gray!40}$47.0$&\cellcolor{gray!40}$53.1$\\
            & & \cellcolor{gray!10}4 &\cellcolor{gray!10}$50.4$&\cellcolor{gray!10}$50.7$&\cellcolor{gray!10}$\textbf{56.8}$&\cellcolor{gray!10}$51.0$&\cellcolor{gray!10}$50.7$\\
                \midrule
                \midrule
        \multirow{2}{*}{\xlmgen} & \multirow{2}{*}{same language}
            &\cellcolor{gray!40}0 &\cellcolor{gray!40}-&\cellcolor{gray!40}$50.5$&\cellcolor{gray!40}$\textbf{60.2}$&\cellcolor{gray!40}$50.1$&\cellcolor{gray!40}$52.4$\\
            & &\cellcolor{gray!10} 4 &\cellcolor{gray!10}-&\cellcolor{gray!10}$51.0$&\cellcolor{gray!10}$47.2$&\cellcolor{gray!10}$50.9$&\cellcolor{gray!10}$50.7$\\
            \midrule
        \multirow{2}{*}{\model{\Ours}{7.5B}} & \multirow{2}{*}{same language} & 
            \cellcolor{gray!40}0 &\cellcolor{gray!40}-&\cellcolor{gray!40}$57.5$&\cellcolor{gray!40}$41.8$&\cellcolor{gray!40}$50.0$&\cellcolor{gray!40}$53.1$\\
            & &\cellcolor{gray!10} 4 &\cellcolor{gray!10}-&\cellcolor{gray!10}$53.2$&\cellcolor{gray!10}${56.5}$&\cellcolor{gray!10}$51.1$&\cellcolor{gray!10}$52.7$\\
                \midrule
                \midrule
        
        \multirow{2}{*}{\xlmgen} & \multirow{2}{*}{English} & 
            \cellcolor{gray!40}0 &\cellcolor{gray!40}-&\cellcolor{gray!40}$55.8$&\cellcolor{gray!40}$57.5$&\cellcolor{gray!40}$\textbf{62.8}^*$&\cellcolor{gray!40}$55.0$\\
             & & \cellcolor{gray!10}4 &\cellcolor{gray!10}-&\cellcolor{gray!10}$52.5$&\cellcolor{gray!10}$49.2$&\cellcolor{gray!10}$53.9$&\cellcolor{gray!10}$52.8$\\
        \midrule
        \multicolumn{8}{c}{Recall}\\
        \midrule
         \multirow{2}{*}{\xlmgen} & \multirow{2}{*}{code switched}
            &\cellcolor{gray!40}  0 &\cellcolor{gray!40}$57.0$&\cellcolor{gray!40}$90.2$&\cellcolor{gray!40}$67.0$&\cellcolor{gray!40}$21.6$&\cellcolor{gray!40}$77.0$\\
            &&\cellcolor{gray!10} 4 &\cellcolor{gray!10}$\textbf{73.0}^*$&\cellcolor{gray!10}$\textbf{76.7}^*$&\cellcolor{gray!10}$\textbf{72.5}$&\cellcolor{gray!10}$65.4$&\cellcolor{gray!10}$\textbf{77.1}^*$\\
                \midrule
        
         \multirow{2}{*}{\model{GPT-3}{6.7B}} & \multirow{2}{*}{code switched}& 
            \cellcolor{gray!40} 0 &\cellcolor{gray!40}$62.4$&\cellcolor{gray!40}$88.6$&\cellcolor{gray!40}$51.5$&\cellcolor{gray!40}$49.0$&\cellcolor{gray!40}$76.4$\\
            & & \cellcolor{gray!10} 4 &\cellcolor{gray!10}$65.7$&\cellcolor{gray!10}$69.2$&\cellcolor{gray!10}$59.6$&\cellcolor{gray!10}$58.5$&\cellcolor{gray!10}$61.1$\\
                \midrule
         \multirow{2}{*}{\model{\Oursns}{7.5B}} & \multirow{2}{*}{code switched}& 
            \cellcolor{gray!40}0 &\cellcolor{gray!40}$\textbf{77.2}^*$&\cellcolor{gray!40}$\textbf{95.4}^*$&\cellcolor{gray!40}$\textbf{80.4}$&\cellcolor{gray!40}$53.8$&\cellcolor{gray!40}$\textbf{95.8}^*$\\
            & & \cellcolor{gray!10}4 &\cellcolor{gray!10}$14.9$&\cellcolor{gray!10}$18.8$&\cellcolor{gray!10}$18.8$&\cellcolor{gray!10}$15.8$&\cellcolor{gray!10}$19.5$\\
                \midrule
                \midrule
        \multirow{2}{*}{\xlmgen} & \multirow{2}{*}{same language}
            &\cellcolor{gray!40}0 &\cellcolor{gray!40}-&\cellcolor{gray!40}$1.4$&\cellcolor{gray!40}$9.1$&\cellcolor{gray!40}$0.2$&\cellcolor{gray!40}$11.6$\\
            & &\cellcolor{gray!10}  4 &\cellcolor{gray!10}-&\cellcolor{gray!10}$50.0$&\cellcolor{gray!10}$66.1$&\cellcolor{gray!10}$\textbf{77.8}$&\cellcolor{gray!10}$33.4$\\
            \midrule
        \multirow{2}{*}{\model{\Oursns}{7.5B}} & \multirow{2}{*}{same language} & 
            \cellcolor{gray!40}0 &\cellcolor{gray!40}-&\cellcolor{gray!40}$87.4$&\cellcolor{gray!40}$39.5$&\cellcolor{gray!40}$0.0$&\cellcolor{gray!40}$79.4$\\
            & &\cellcolor{gray!10} 4 &\cellcolor{gray!10}-&\cellcolor{gray!10}$39.0$&\cellcolor{gray!10}$24.9$&\cellcolor{gray!10}$55.8$&\cellcolor{gray!10}$44.6$\\
                \midrule
                \midrule
        
        \multirow{2}{*}{\xlmgen} & \multirow{2}{*}{English} & 
            \cellcolor{gray!40} 0 &\cellcolor{gray!40}-&\cellcolor{gray!40}$93.8$&\cellcolor{gray!40}$77.8$&\cellcolor{gray!40}$\textbf{74.8}^*$&\cellcolor{gray!40}$77.6$\\
             & & \cellcolor{gray!10}4 &\cellcolor{gray!10}-&\cellcolor{gray!10}$72.4$&\cellcolor{gray!10}$71.2$&\cellcolor{gray!10}$75.7$&\cellcolor{gray!10}$73.7$\\
         \bottomrule
    \end{tabular}
    }
    \caption{Accuracy and recall scores of our multilingual model and other English models on the Hate Speech Detection task. We evaluate five target languages. For each target language, we bold the highest number for zero-shot and four-shot respectively. * indicates the number is significantly higher than others. For language condition, we consider three cases: ``code switched'' means the prefix, candidates are in English and tweets are in the target language; ``same language'' means prefix, candidates and tweets are in the target language; ``English'' means prefix, candidates and tweets are in English, i.e. note that, ``same language" and ``English" reduce to the same experimental condition when the target language is English.
    }
    \label{tab:hate-speech-performance}
\end{table*}

\begin{table*}[t]
    \centering
    \scalebox{0.8}{
    \begin{tabular}{c|cc|cc|cc}
        \toprule
         & \multicolumn{2}{c}{English} & \multicolumn{2}{c}{Spanish} & \multicolumn{2}{c}{French} \\
         Model & Avg.$\uparrow$ & |Diff|$\downarrow$& Avg.$\uparrow$ & |Diff|$\downarrow$& Avg.$\uparrow$ & |Diff|$\downarrow$\\
         \midrule
        %  6.7B en translate & 90.73 & 2.81 & 91.19 & 2.42 & 89.83 & 1.65\\
        \model{\Ours}{6.7B} \lang{En}-only & \textbf{90.73} & 3.19 & \textbf{91.23} & \textbf{2.65} & 83.46 & 4.85\\
         \model{GPT-3}{6.7B} & 90.42 & 3.53 & 86.91 & 5.18 & \textbf{90.85} & \textbf{2.30}\\
         \model{\Ours}{7.5B} & 86.49 & \textbf{2.83} & 82.93 & 4.28 & 76.55 & 2.95\\
         \midrule
        \model{\Ours}{6.7B} \lang{En}-only $+$ translate & - & 3.19 & 91.18 & 3.75 & 90.07 & 2.35\\
         \bottomrule
    \end{tabular}
    }
    \caption{Accuracy and bias scores of our multilingual model and other English models on the occupation identification task. ``|Diff|'' stands for the average absolute accuracy gap between  male and female groups aggregated across all  occupations. We bold the highest accuracy score for each language.}
    \label{tab:bio-classification-performance}
\end{table*}

\subsection{Gender Bias in Occupation Identification}
\label{sec:bio-classification}
\subsubsection{Setup}
\noindent \textbf{Datasets} We use the English bio dataset introduced in ~\cite{de2019bias} to study gender bias based on identifying a person's occupation from their bios. For multilingual bio datasets  we use those created by~\cite{zhao2020gender}. Originally there are $28$ occupations in English, $69$ occupations in Spanish and $27$ occupations in French. To ensure we have plenty of test data for each occupation, we only keep occupations with at least $1000$ male examples and $1000$ female examples. This leads to $16$ occupations in English, $6$ occupations in Spanish and $4$ occupations in French. We follow the setup in ~\cite{zhao2020gender} where people's names and pronouns are removed from the bios. We then prefix ``The occupation of this person is <candidate>'' to the input bio to form a prompt. The candidate set consists of five occupations, including the ground truth one and four other randomly sampled male and female occupations (two male and two female). Male (female) occupations refer to ones having predominantly more male (female) samples.  

\noindent \textbf{Metrics}
Similar to the metric for Hate Speech detection, we first obtain the scores for 5 candidates and consider a prediction correct if the ground truth candidate yields the highest score among five candidates. We then compute the bias score as the absolute gap between the accuracy scores on the male and female samples,\footnote{We only consider gaps that are statistically significant.} averaged across all occupations. A lower bias score indicates that a model has less divergence in identifying occupations for men and women. 

\hide{
\begin{figure*}[t!]
    \centering
    \begin{subfigure}[b]{0.8\linewidth}
    \includegraphics[width=\linewidth]{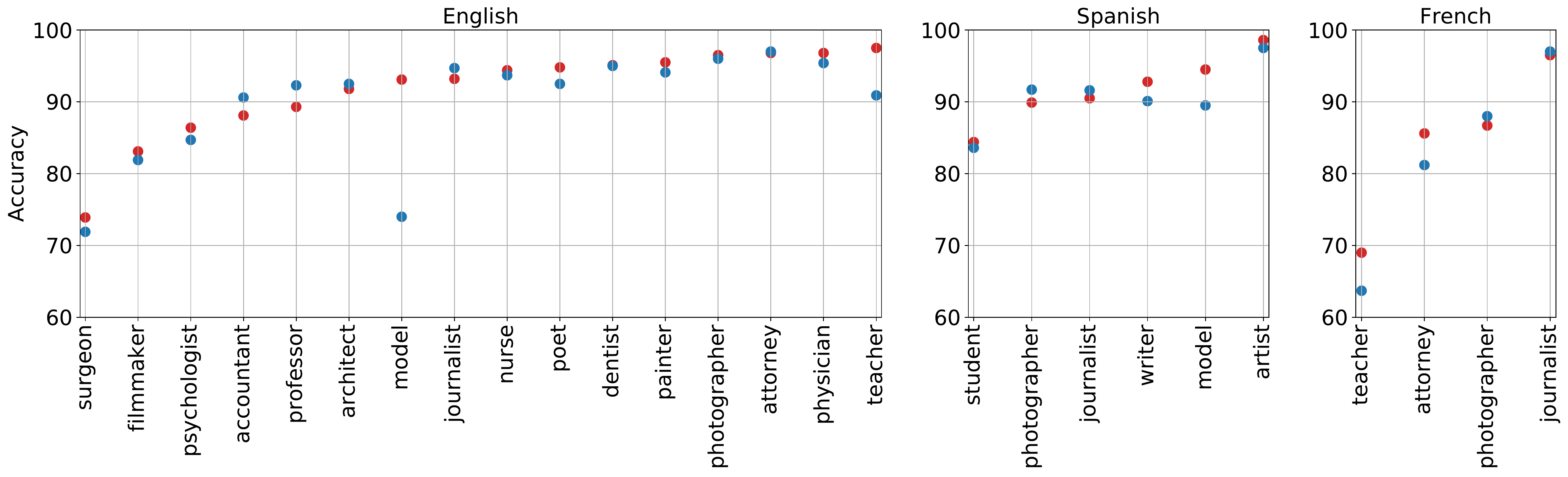}
    \caption{\model{\Ours}{6.7B} En-only}\label{fig:bio_6.7B}
    \end{subfigure}
    
    \begin{subfigure}[b]{0.8\linewidth}
    \includegraphics[width=\linewidth]{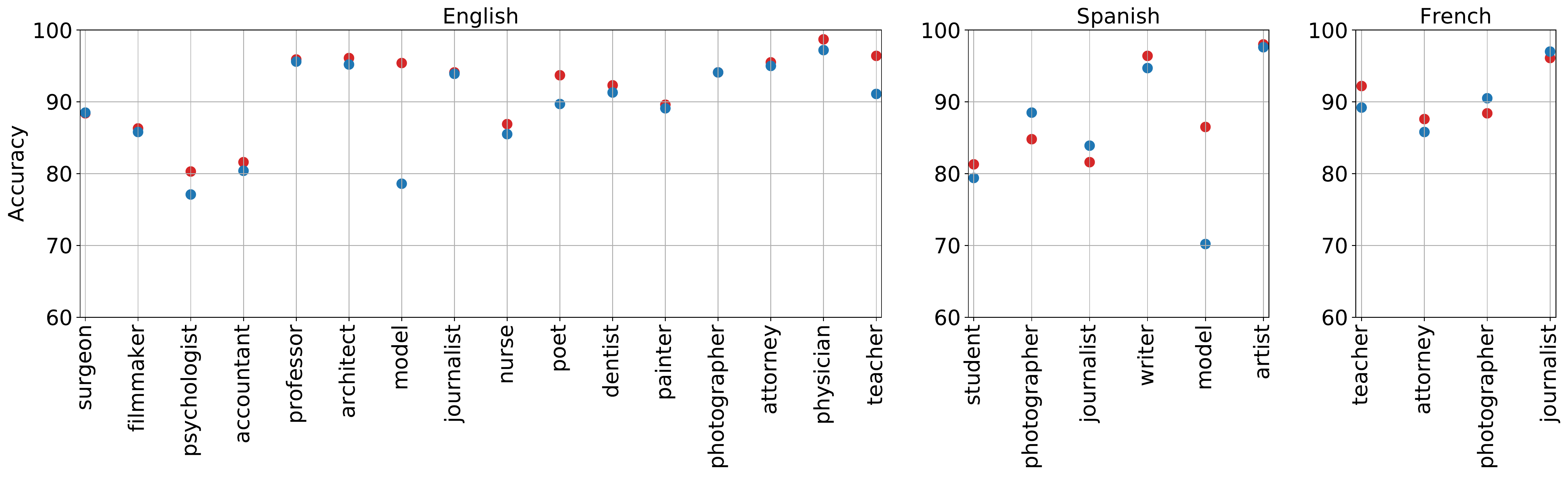}
    \caption{\model{GPT-3}{6.7B}}\label{fig:bio_curie}
    \end{subfigure}
    
    \begin{subfigure}[b]{0.8\linewidth}
    \includegraphics[width=\linewidth]{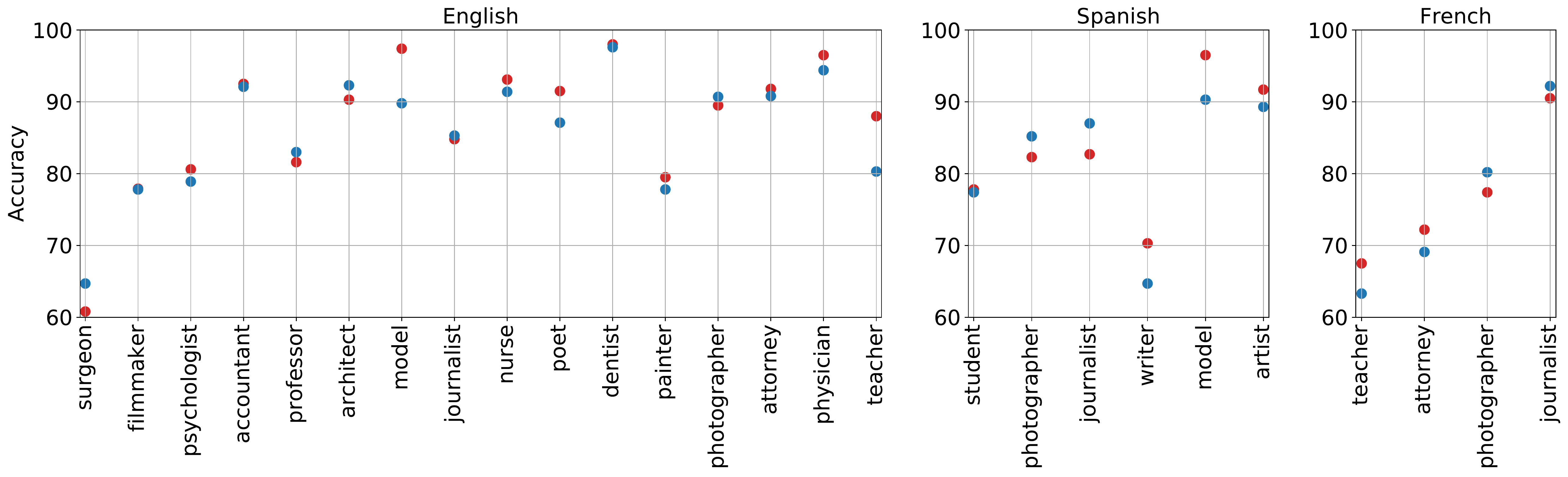}
    \caption{\model{\Ours}{7.5B}}\label{fig:bio_mul}
    \end{subfigure}
    \caption{Accuracy of the male group (blue) and the female group (red) for different occupations.}
    \label{fig:bio}
\end{figure*}
}

\subsubsection{Results}
We present the overall accuracy scores and the bias scores (|Diff|) in Table~\ref{tab:bio-classification-performance}. Results indicate that the \model{\Ours}{6.7B} En-only model achieves the best performance on English and Spanish, while the \model{GPT-3}{6.7B} model achieves the best performance on French. \model{\Ours}{7.5B} model, instead, falls behind on all three languages, especially for Spanish and French. We think this is potentially due to that all pronouns and people's names are removed from the test data but not training data. The training data for \model{\Ours}{7.5B} contains more Spanish and French compared to the other two models. Thus, \model{\Ours}{7.5B} may have more severe morphological mismatch on Spanish and English.
Regarding the bias score, the \model{GPT-3}{6.7B} model is the most biased model on both English and Spanish but least biased on French. \model{\Ours}{6.7B} En-only and \model{\Ours}{7.5B} exhibit the least bias on Spanish and English, respectively. 
\section{Data Card}
\label{sec:appendix_datacard}
We follow the recommendations of Gebru et al. (2018) and provide a datacard for the dataset used to train \Ours, which is a subset of \cchxl, a larger multilingual dataset we curated.

% \vic{Can move training set up details to appendix.}
\subsection{Data Sources} 
\label{sec:data_sources}
Following the recent success of multilingual self-supervised pre-training~\cite{devlin2018bert,lample2019cross,conneau-etal-2020-unsupervised,Xue2020MT5:Transformer,DBLP:journals/corr/abs-2105-00572,liu2020multilingual}, we train our language models on a mixture of monolingual text of different languages. 
We extend the pipeline used for mining the {CC100} %\footnote{\url{http://data.statmt.org/cc-100/}} 
corpus~\citep{conneau-etal-2020-unsupervised, wenzek-etal-2020-ccnet} to generate CC100-XL, a significantly larger multilingual dataset covering 68 Common Crawl (CC) snapshots (from \href{http://commoncrawl.org/2013/11/new-crawl-data-available/}{Summer 2013} to \href{https://commoncrawl.org/2020/04/march-april-2020-crawl-archive-now-available/}{March/April 2020}) and 134 languages. % (Table~\ref{tab:cc100_data_table}). 
As the first step to balance the language distribution, we sampled  30\% of the data from the languages that contain more than 15 billion tokens and more than 20 million documents. %, and kept all data for the rest of the languages. 
This resulted in a 8.4 TB multilingual corpus with 1.9 trillion tokens. 
% To improve data quality, we apply several data filtering criteria\footnote{\citet{brown2020gpt3} use high-quality monolingual corpora such as the English Wikipedia and BookCorpus~\cite{moviebook} to augment CC, which has lower quality text~\cite{DBLP:journals/corr/abs-2103-12028}. In comparison, our training data exclusively consists of filtered CC data.} % As a future work, we plan to also include high-quality reference corpora in our training datasets, and understand their contribution to the model performance better.} with the details documented in the Data Card presented in Appendix \ref{sec:appendix_datacard}. % \comment{Shuohui: please add data card.}
%In addition to filtering based on language model scores, we also filtered paragraphs having more than one url, fewer than three tokens (except agglutinative languages), ratio of digits/punctuations to total characters greater than 0.25 and type-token ratio less than 0.5. 
% The pre-training data was selected from \textsc{CC100-XL}. For languages which contains more than 15 billion tokens and more than 20 million documents, 30\% of data were randomly drawn for each; and all the data were used for other languages. 

% The final selected pre-training data contains 113 ISO 639-1 languages codes and 18 unassigned language codes\footnote{\url{https://en.wikipedia.org/wiki/List_of_ISO_639-1_codes}}, and it contains 8.4 TB and 1.9 trillion tokens.
\begin{table*}[t!]
\begin{center}
\scalebox{0.8}{
    \begin{tabular}{llrr|llrr}
    ISO code & Language & Tokens (M) & Size (GiB) & ISO code & Language & Tokens (M) & Size (GiB) \\
    \midrule
    en &      English & 803,527 &   3,324.45 &     - &  Arabic Romanized &     685 &    1.65 \\
    ru &      Russian & 147,792 &   1,007.38 &    mn &      Mongolian &     681 &    4.26 \\
    zh &      Chinese & 132,770 &  485.32 &    la &          Latin &     635 &    2.20 \\
    de &       German &  89,224 &  369.30 &    ne &         Nepali &     600 &    5.32 \\
    es &      Spanish &  87,303 &  363.83 &    si &      Sinhalese &     524 &    3.96 \\
    fr &       French &  77,420 &  303.76 &    mr &        Marathi &     458 &    3.59 \\
    ja &     Japanese &  66,054 &  293.39 &    kn &        Kannada &     446 &    3.41 \\
    it &      Italian &  41,930 &  170.76 &    so &         Somali &     436 &    1.56 \\
    pt &   Portuguese &  36,586 &  147.12 &    cy &          Welsh &     398 &    1.27 \\
    el &        Greek &  28,762 &  180.37 &    jv &       Javanese &     389 &    1.23 \\
    ro &     Romanian &  24,176 &   93.63 &    ps &         Pashto &     387 &    1.97 \\
    uk &    Ukrainian &  23,723 &  156.68 &    uz &          Uzbek &     332 &    1.64 \\
    hu &    Hungarian &  22,718 &   89.87 &    gu &       Gujarati &     327 &    2.10 \\
    ko &       Korean &  20,002 &   79.08 &    km &          Khmer &     272 &    2.14 \\
    pl &       Polish &  19,293 &   73.59 &     - & Urdu Romanized &     245 &    0.73 \\
    no &    Norwegian &  17,600 &   70.89 &    am &        Amharic &     169 &    0.85 \\
    nl &        Dutch &  17,163 &   68.36 &     - & Bengali Romanized &     166 &    0.48 \\
    fi &      Finnish &  16,804 &   67.28 &    pa &        Punjabi &     153 &    0.93 \\
    da &       Danish &  16,274 &   64.74 &    gl &       Galician &     137 &    0.50 \\
    id &   Indonesian &  15,424 &   67.51 &    ha &          Hausa &     124 &    0.42 \\
    hr &     Croatian &  14,455 &   54.27 &    mg &       Malagasy &     116 &    0.38 \\
    tr &      Turkish &  12,413 &   51.51 &    sa &       Sanskrit &     107 &    0.42 \\
    ar &       Arabic &  12,249 &   64.34 &    eu &         Basque &     105 &    0.35 \\
    vi &   Vietnamese &  11,199 &   50.45 &    my &        Burmese &     101 &    0.74 \\
    th &         Thai &  10,842 &   99.86 &    su &      Sundanese &      91 &    0.30 \\
    bg &    Bulgarian &   9,704 &   61.10 &    or &          Oriya &      91 &    0.62 \\
    fa &      Persian &   9,355 &   57.46 &    ht &        Haitian &      87 &    0.28 \\
    sv &      Swedish &   9,169 &   36.54 &    lo &            Lao &      84 &    0.59 \\
    ms &        Malay &   9,106 &   38.57 &    ky &         Kyrgyz &      70 &    0.34 \\
    he &       Hebrew &   8,637 &   42.13 &    br &         Breton &      57 &    0.16 \\
    cs &        Czech &   8,616 &   32.46 &    ga &          Irish &      49 &    0.15 \\
    sk &       Slovak &   8,251 &   30.70 &    yo &         Yoruba &      48 &    0.14 \\
    ca &      Catalan &   7,076 &   26.90 &    eo &      Esperanto &      47 &    0.14 \\
    lt &   Lithuanian &   4,847 &   18.38 &     - &   Tamil Romanized &      40 &    0.13 \\
    sl &      Slovene &   4,029 &   14.97 &    zu &           Zulu &      40 &    0.14 \\
    hi &        Hindi &   3,448 &   26.63 &    ti &       Tigrinya &      40 &    0.19 \\
    et &     Estonian &   3,287 &   12.18 &     - &  Telugu Romanized &      37 &    0.11 \\
    lv &      Latvian &   2,815 &   10.67 &    ku &        Kurdish &      36 &    0.10 \\
    tl &      Tagalog &   2,389 &    8.13 &    om &          Oromo &      27 &    0.09 \\
    sq &     Albanian &   2,382 &    8.76 &    xh &          Xhosa &      26 &    0.09 \\
    sr &      Serbian &   2,101 &   12.68 &    gd &   Scottish Gaelic &      19 &    0.05 \\
     - & Hindi Romanized &   2,045 &    6.60 &    ig &           Igbo &      18 &    0.06 \\
    az &  Azerbaijani &   1,904 &    8.41 &    as &       Assamese &      17 &    0.10 \\
    bn &      Bengali &   1,627 &   11.19 &    lg &          Ganda &      15 &    0.05 \\
    ta &        Tamil &   1,477 &   12.36 &    wo &          Wolof &      14 &    0.03 \\
    ur &         Urdu &   1,352 &    7.77 &    fy &   Western Frisian &      12 &    0.04 \\
    kk &       Kazakh &   1,278 &    8.40 &    tn &         Tswana &      11 &    0.03 \\
    hy &     Armenian &   1,261 &    7.16 &    ff &           Fula &      11 &    0.03 \\
    ka &     Georgian &   1,261 &   10.48 &    gn &        Guaraní &      10 &    0.03 \\
    is &    Icelandic &   1,163 &    4.21 &    sd &         Sindhi &       8 &    0.04 \\
    be &   Belarusian &   1,004 &    5.81 &    ln &        Lingala &       7 &    0.02 \\
    bs &      Bosnian &     950 &    4.00 &    bm &        Bambara &       6 &    0.02 \\
    ml &    Malayalam &     935 &    8.08 &    iu &      Inuktitut &       6 &    0.03 \\
    mk &   Macedonian &     927 &    6.05 &    kg &          Kongo &       4 &    0.01 \\
    sw &      Swahili &     908 &    3.19 &    qu &        Quechua &       3 &    0.01 \\
    af &    Afrikaans &     819 &    3.04 &    ss &          Swati &       2 &    0.01 \\
    te &       Telugu &     689 &    5.28 &     - &     Unassigned &     503 &    2.30 \\
    \bottomrule
    \end{tabular}
}
\caption{Languages and statistics of the training data set selected from CC100 XL corpus.}
\label{tab:cc100_data_table}
\end{center}
\end{table*}

\hide{
\begin{figure*}[t!]
    \centering
    \includegraphics[width=\linewidth]{figures/pretraining_tokens_130.pdf}
    \caption{Number of tokens of the pre\-training data (134 ISO language-locale) in log scale from CC100\_XL corpus.\vic{Enlarge fonts so that the plot is readable. Move to appendix. Maybe rotate 90 degree.} [Shuohui: showing either this figure (+another similar chart by size) or the above table may be enough, no need for both]}
    \label{fig:pretraining_data_size}
\end{figure*}
}

\subsection{Motivation}
\begin{itemize}
    \item \textbf{For what purpose was the dataset created? Was there a specific task in mind? Was there a specific gap that needed to be filled? Please provide a description.} The \cchxl dataset was collected to create a high quality % (fluent) 
    monolingual dataset for at least 100 languages. It was mainly used for training foundation multilingual language models which may be applied to a broad list of language tasks, including neural machine translation, speech translation, question answering, etc. CC100-XL involves sentence level
filtering%\comment{What do we mean by sentence level alignment?}\comment{good question. changing "alignment" to "filtering"}
, preserves context, improves the filtering mechanism, and paves a way for mining 200+ languages.
    \item \textbf{Who created the dataset (e.g., which team, research group) and on behalf of which entity (e.g., company, institution, organization)?} Meta AI.
    \item \textbf{Who funded the creation of the dataset? If there is an associated grant, please provide the name of the grantor and the grant name and number.} Meta AI. %\xian{Do we need to specify Platform? I suggest just say Meta AI.}[Shuohui: makes sense and changed accordingly].
    \item \textbf{Any other comments?} No.
\end{itemize}

\subsection{Composition}
\begin{itemize}
    \item \textbf{What do the instances that comprise the dataset represent (e.g., documents, photos, people, countries)? Are there multiple types of instances (e.g., movies, users, and ratings; people and interactions between them; nodes and edges)? Please provide a description.} The instances are textual documents sampled from Commoncrawl snapshots.
    \item \textbf{How many instances are there in total (of each type, if appropriate)?} The training dataset of \Ours contains 1.74 billion documents in total. % \comment{Shuohui: please confirm this statement. Victoria: 1.74b docs is confirmed.}
    \item \textbf{Does the dataset contain all possible instances or is it a sample (not necessarily random) of instances from a larger set? If the dataset is a sample, then what is the larger set? Is the sample representative of the larger set (e.g., geographic coverage)? If so, please describe how this representativeness was validated/verified. If it is not representative of the larger set, please describe why not (e.g., to cover a more diverse range of instances, because instances were withheld or unavailable).} The dataset is a subset of CC100-XL. For each language, the data is either a full set or a random subset of CC100-XL data. Especially, the medium- and low-resource languages are upsampled. % for different languages, it should be representative at the language level. 
    % But the relative ratio and sampling probability across languages are changed from the original in favor of upsampling low-resource languages.
    In terms of language representation, the CC100-XL dataset contains 134 languages extracted using fasttext\footnote{https://fasttext.cc/docs/en/language-identification.html} from Common Crawl snapshots. We further selected a subset of 30 languages to train \Ours, taking geo-location, language family and typology diversity of the languages into account.
    \item \textbf{What data does each instance consist of? “Raw” data (e.g., unprocessed text or images) or features? In either case, please provide a description.} Each instance consists of raw text data.
    \item \textbf{Is there a label or target associated with each instance? If so, please provide a description.} No.
    \item \textbf{Is any information missing from individual instances? If so, please provide a description, explaining why this information is missing (e.g., because it was unavailable). This does not include intentionally removed information, but might include, e.g., redacted text.} No.
    \item \textbf{Are relationships between individual instances made explicit (e.g., users' movie ratings, social network links)? If so, please describe how these relationships are made explicit.} A small percentage of document instances (<2\%) are duplicated. Other than that, there are no relationships between individual instances.
    \item \textbf{Are there recommended data splits (e.g., training, development/validation, testing)? If so, please provide a description of these splits, explaining the rationale behind them.} This dataset is split into training and validation only. For each high resource language, at least 5,000 randomly selected documents and 30,000 lines % \comment{Shuohui: 3000 lines or 30k lines? Victoria: Nice catch! it should be a mininum of 5K docs and 30K lines} 
    were split into validation set, and the rest documents are for training; for low-resource languages, at least 100 randomly selected documents and 1,000 lines (a couple of very low resource languages contain ~80 documents) were split into valid set and leave the rest for training. There are 3.5 million lines of text in total for the validation set. This split is mainly to ensure a good size of validation data with the coverage and balance over all languages, meanwhile, the validation size is not too large to affect the overall training speed. % [Xian, Victoria: please take a detailed look of this. there may be some further change on the validation set at tokenization (after the initial split).]
    \item \textbf{Are there any errors, sources of noise, or redundancies in the dataset? If so, please provide a description.} ~10\% of Russian sample were lost during internal data transferring. Therefore, we ended up taking 26.7\% random subset of the whole Russian data from CC100-XL.
    \item \textbf{Is the dataset self-contained, or does it link to or otherwise rely on external resources (e.g., websites, tweets, other datasets)?} It's self-contained.
    \item \textbf{Does the dataset contain data that might be considered confidential (e.g., data that is protected by legal privilege or by doctor-patient confidentiality, data that includes the content of individuals' non-public communications)? If so, please provide a description.} \cchxl is exclusively extracted from Common Crawl; and the information in it is not considered confidential.
    \item \textbf{Does the dataset contain data that, if viewed directly, might be offensive, insulting, threatening, or might otherwise cause anxiety? If so, please describe why.} \cchxl is a subset of public Common Crawl data, which could contain sentences that, if viewed directly, might be offensive, insulting, threatening, or might otherwise cause anxiety.
    \item \textbf{Does the dataset relate to people? If not, you may skip the remaining questions in this section.} Some documents of this data relate to people, such as news articles, Wikipedia descriptions, etc.
    \item \textbf{Does the dataset identify any subpopulations (e.g., by age, gender)? If so, please describe how these subpopulations are identified and provide a description of their respective distributions within the dataset.} No.
    \item \textbf{Is it possible to identify individuals (i.e., one or more natural persons), either directly or indirectly (i.e., in combination with other data) from the dataset? If so, please describe how} Other than the individuals who are celebrities, politicians, etc, and have their Wikipedia pages; it is possible to identify other individuals by their names, twitter account names, etc. But we built personally identifiable information (PII) identification tools following the guidelines of General Data Protection Regulation (GDPR) and National Institute of Standards and Technology (NIST) and run against this dataset, we did not found highly sensitive PII information, such as U.S. social security number, login credentials, etc.
    \item \textbf{Does the dataset contain data that might be considered sensitive in any way (e.g., data that reveals racial or ethnic origins, sexual orientations, religious beliefs, political opinions or union memberships, or locations; financial or health data; biometric or genetic data; forms of government identification, such as social security numbers; criminal history)? If so, please provide a description.} We use a curated special word list of 100 languages which covers profanities, hate speech, bulling language, common slangs and profane multi-word expressions (MWEs) to tag paragraphs and remove the docs containing them. Given the size of this data, it could still contain such sensitive information (as the above lists may not be exhaustive) but should be a very small percent of instances. %\comment{Vishrav, Xian, Victoria: please review this one particually}
    \item \textbf{Any other comments?} No
\end{itemize}

\subsection{Collection Process}
\begin{itemize}
    \item \textbf{How was the data associated with each instance acquired? Was the data directly observable (e.g., raw text, movie ratings), reported by subjects (e.g., survey responses), or indirectly inferred/ derived from other data (e.g., part-of-speech tags, model-based guesses for age or language)? If data was reported by subjects or indirectly inferred/derived from other data, was the data validated/verified? If so, please describe how.} Please refer to the main document for details.
    \item \textbf{What mechanisms or procedures were used to collect the data (e.g., hardware apparatus or sensor, manual human curation, software program, software API)? How were these mechanisms or procedures validated?} Please refer to the main document for details.
    \item \textbf{If the dataset is a sample from a larger set, what was the sampling strategy (e.g., deterministic, probabilistic with specific sampling probabilities)?} Please refer to the main document for details.
    \item \textbf{Who was involved in the data collection process (e.g., students, crowdworkers, contractors) and how were they compensated (e.g., how much were crowdworkers paid)?} This data is mined, filtered and sampled by machines.
    \item \textbf{Over what timeframe was the data collected? Does this timeframe match the creation timeframe of the data associated with the instances (e.g., recent crawl of old news articles)? If not, please describe the timeframe in which the data associated with the instances was created.} The data was collected from 68 Common Crawl (CC) snapshots (from \href{http://commoncrawl.org/2013/11/new-crawl-data-available/}{Summer 2013} to \href{https://commoncrawl.org/2020/04/march-april-2020-crawl-archive-now-available/}{March/April 2020}). Therefore, it does not contain a lot of information about recent events such as COVID-19.
    \item \textbf{Were any ethical review processes conducted (e.g., by an institutional review board)? If so, please provide a description of these review processes, including the outcomes, as well as a link or other access point to any supporting documentation.} No. %\comment{Vishrav, Xian, Victoria: please address this question or share pointers.}\xian{I'm not aware of ethical review for this dataset.}
    \item \textbf{Does the dataset relate to people? If not, you may skip the remainder of the questions in this section.} No.
    \item \textbf{Did you collect the data from the individuals in question directly, or obtain it via third parties or other sources (e.g., websites)?} N/A
    \item \textbf{Were the individuals in question notified about the data collection? If so, please describe (or show with screenshots or other information) how notice was provided, and provide a link or other access point to, or otherwise reproduce, the exact language of the notification itself.} N/A
    \item \textbf{Did the individuals in question consent to the collection and use of their data? If so, please describe (or show with screenshots or other information) how consent was requested and provided, and provide a link or other access point to, or otherwise reproduce, the exact language to which the individuals consented.} N/A
    \item \textbf{If consent was obtained, were the consenting individuals provided with a mechanism to revoke their consent in the future or for certain uses? If so, please provide a description, as well as a link or other access point to the mechanism (if appropriate).} N/A
    \item \textbf{Has an analysis of the potential impact of the dataset and its use on data subjects (e.g., a data protection impact analysis) been conducted? If so, please provide a description of this analysis, including the outcomes, as well as a link or other access point to any supporting documentation.} Some responsible AI related evaluations were performed. Please refer to the main document.
    \item \textbf{Any other comments?} No
\end{itemize}

\subsection{Preprocessing/cleaning/labeling}
\begin{itemize}
    \item \textbf{Was any preprocessing/cleaning/labeling of the data done (e.g., discretization or bucketing, tokenization, part-of-speech tagging, SIFT feature extraction, removal of instances, processing of missing values)? If so, please provide a description. If not, you may skip the remainder of the questions in this section.} Yes, the detailed steps are as below:
    \begin{itemize}
        \item \textit{Downloading and Sharding Commoncrawl Snapshots} We downloaded 68 Commoncrawl snapshots and divided the data in 240 shards based on web-domain. At this stage, textual data gets extracted from the WET files provided by Common Crawl which involves cleaning excessive tabs and newlines. %\xian{The last sentence is not clear. Suggesting not using ``junk text", and clarify what ``removing tabs" mean}\comment{made some change accordingly. please take one more pass}.
        \item \textit{Language Identification (LID) at Document Level} For this stage, we used the fastText language identification (LID) model on the entire document which helped further divide the data by language. In addition to the original languages supported by fastText, we also added support for 28 romanized languages. In total, the data for each language contains 240 shards.
        \item \textit{Deduplicating Documents based on URL} We aggregated the data based on URL which yields ~60\% reduction in volume. In case two documents had the same URL, we selected the document having more recent text content.
        \item \textit{Document Splitting and LID at Paragraph Level} We segmented the documents based on newline and also stored the information about the order in which the paragraphs were appearing in the original document (i.e. seq\_num). Next, we performed LID at the paragraph level again in order to divide the original documents into clusters of paragraphs where each cluster represents sentences belonging to a particular language.
        \item \textit{Deduplicating Paragraphs} Data extracted from Commoncrawl snapshots still have a lot of duplicate text even if the document is different. In order to tackle this, we applied the normalization function from CCNet~\cite{wenzek-etal-2020-ccnet} and then computed a SHA-1 hash of the normalized text. This helped in reducing the content by ~88\%. Choosing which <paragraph, url> combination to keep can be tricky as it can lead to a lot of fragmented documents. So we devised a strategy to choose documents based on sorted <url, seq\_num> order which would help in preventing fragmentation as much as possible.
        \item \textit{Language Model Scores} We scored every paragraph using a Language Model trained on data collected from OPUS~\cite{tiedemann-2012-parallel}  (monolingual data collected from the availble bitexts) using a 4-gram KenLM~\cite{heafield-2011-kenlm}. % (Heafield, 2011\footnote{Kenneth Heafield. 2011. Kenlm: Faster and smaller language model queries. In Proceedings of the SixthWorkshop on Statistical Machine Translation, pp 187–197. Association for Computational Linguistics.})\comment{Xian, Victoria: is it a good way to add reference at the appendix?}. 
Note that since the LMs were not trained on data belonging to a specific domain, this feature helped in eliminating general non-fluent sentences.
        \item \textit{Heuristic based approaches} We use the following techniques to further refine the filtering step (especially useful for Low resource languages having no or poor quality LM)
        \begin{itemize}
            \item Ratio of digit+punctuation to total characters (current threshold <0.25)
            \item Maximum number of URLs per sentence (current value 1)
            \item Type-token ratio (current threshold >0.6 + removing bottom 1\% per language)
            \item Minimum number of tokens per sentence (current value 3; not applied for agglutinative languages)
        \end{itemize}
        \item \textit{Tagging profane words and removing instances containing such words}
    \end{itemize}
    \item \textbf{Was the “raw” data saved in addition to the preprocessed/cleaned/labeled data (e.g., to support unanticipated future uses)? If so, please provide a link or other access point to the “raw” data.} The ``raw'' data is publiclly available in in https://commoncrawl.org/the-data.
    \item \textbf{Is the software used to preprocess/clean/label the instances available? If so, please provide a link or other access point.} The software is proprietary to Meta Platforms and currently unavailable publicly.
    \item \textbf{Any other comments?} No
\end{itemize}

\subsection{Uses}
\begin{itemize}
    \item \textbf{Has the dataset been used for any tasks already? If so, please provide a description.} Yes, this dataset and its precursor CC100 data have been used to train machine translations and multilingual language models, which are foundation to many downstream language tasks.
    \item \textbf{Is there a repository that links to any or all papers or systems that use the dataset? If so, please provide a link or other access point.} No.
    \item \textbf{What (other) tasks could the dataset be used for?} This data can be used to pretrain multilingual language models, which are foundation to many current and future language tasks.
    \item \textbf{Is there anything about the composition of the dataset or the way it was collected and preprocessed/cleaned/labeled that might impact future uses? For example, is there anything that a future user might need to know to avoid uses that could result in unfair treatment of individuals or groups (e.g., stereotyping, quality of service issues) or other undesirable harms (e.g., financial harms, legal risks) If so, please provide a description. Is there anything a future user could do to mitigate these undesirable harms?} The pipeline for creating this dataset paves a way for building a scalable infrastructure for mining datasets to be be used for training large-scale models.
    \item \textbf{Are there tasks for which the dataset should not be used? If so, please provide a description.} No.
    \item \textbf{Any other comments?} No.
\end{itemize}

\subsection{Distribution}
\begin{itemize}
    \item \textbf{Will the dataset be distributed to third parties outside of the entity (e.g., company, institution, organization) on behalf of which the dataset was created? If so, please provide a description.} No. 
    \item \textbf{How will the dataset will be distributed (e.g., tarball on website, API, GitHub)? Does the dataset have a digital object identifier (DOI)?} N/A
    \item \textbf{When will the dataset be distributed?} No.
    \item \textbf{Will the dataset be distributed under a copyright or other intellectual property (IP) license, and/or under applicable terms of use (ToU)? If so, please describe this license and/or ToU, and provide a link or other access point to, or otherwise reproduce, any relevant licensing terms or ToU, as well as any fees associated with these restrictions.} No.
    \item \textbf{Have any third parties imposed IP-based or other restrictions on the data associated with the instances? If so, please describe these restrictions, and provide a link or other access point to, or otherwise reproduce, any relevant licensing terms, as well as any fees associated with these restrictions.} No.
    \item \textbf{Do any export controls or other regulatory restrictions apply to the dataset or to individual instances? If so, please describe these restrictions, and provide a link or other access point to, or otherwise reproduce, any supporting documentation.} N/A
    \item \textbf{Any other comments?} No.
\end{itemize}

\subsection{Maintenance}
\begin{itemize}
    \item \textbf{Who is supporting/hosting/maintaining the dataset?} Meta AI.
    \item \textbf{How can the owner/curator/manager of the dataset be contacted (e.g., email address)?} Refer to the main document.
    \item \textbf{Is there an erratum? If so, please provide a link or other access point.} Currently no.
    \item \textbf{Will the dataset be updated (e.g., to correct labeling errors, add new instances, delete instances)? If so, please describe how often, by whom, and how updates will be communicated to users (e.g., mailing list, GitHub)?} No plan for updating.
    \item \textbf{If the dataset relates to people, are there applicable limits on the retention of the data associated with the instances (e.g., were individuals in question told that their data would be retained for a fixed period of time and then deleted)? If so, please describe these limits and explain how they will be enforced.} N/A
    \item \textbf{Will older versions of the dataset continue to be supported/hosted/maintained? If so, please describe how. If not, please describe how its obsolescence will be communicated to users.} N/A
    \item \textbf{If others want to extend/augment/build on/contribute to the dataset, is there a mechanism for them to do so? If so, please provide a description. Will these contributions be validated/verified? If so, please describe how. If not, why not? Is there a process for communicating/ distributing these contributions to other users? If so, please provide a description.} No.
    \item \textbf{Any other comments?} No.
\end{itemize}

\end{document}